\documentclass[final]{cvpr}

\long\def\ignorethis#1{}

\newcommand{\img}[1]{\mathbf{I}_{\text{#1}}}
\newcommand{\paren}[1]{\left( #1 \right)}

\newcommand{\normtwo}[1]{\left\lVert #1 \right\rVert_2^2}
\newcommand{\normone}[1]{\left\lVert #1 \right\rVert_1}
\newcommand{\normzero}[1]{\left\lVert #1 \right\rVert_0}
\newcommand{\normfrob}[1]{\left\lVert #1 \right\rVert_F}

\newcommand{\transpose}{^\mathsf{T}}
\newcommand{\round}[1]{\lfloor #1 \rceil}

\newcommand{\Paragraph}[1]{\noindent{\textbf{#1}}}

\newcommand{\fakeref}[1]{\textcolor{blue}{#1}}

\newcommand{\ignore}[1]{}   

\newcommand{\tabcspace}{\vspace{-2.5mm}}
\newcommand{\tabspace}{\vspace{-4.5mm}}
\newcommand{\figcspace}{\vspace{-2.5mm}}
\newcommand{\figspace}{\vspace{-4.5mm}}
\newcommand{\best}[1]{\textbf{#1}}
\newcommand{\secondbest}[1]{\underline{#1}}

\usepackage{paralist}
\usepackage{times}
\usepackage{epsfig}
\usepackage[font=normal]{subfig}
\usepackage{graphicx}
\usepackage{graphbox}
\usepackage{animate}
\usepackage{tabularx}
\usepackage{amsmath}
\usepackage{nicefrac}
\usepackage{booktabs}
\usepackage{amssymb}
\usepackage{pifont}

\usepackage{multirow}
\usepackage{makecell}

\usepackage[pagebackref=false,colorlinks,bookmarks=false,citecolor=blue,linkcolor=blue]{hyperref}

\def\cvprPaperID{951} 
\def\confYear{CVPR 2021}

\newcommand{\cv}[1]{{\fontfamily{qcr}\selectfont #1}}
\begin{document}

\title{SRWarp: Generalized Image Super-Resolution under Arbitrary Transformation}

\author{Sanghyun Son \hspace{15mm} Kyoung Mu Lee \\
ASRI, Department of ECE, Seoul National University, Seoul, Korea \\
{\tt\small \{thstkdgus35, kyoungmu\}@snu.ac.kr}
}

\maketitle

\begin{abstract}
    Deep CNNs have achieved significant successes in image processing and its applications, including single image super-resolution (SR).
    However, conventional methods still resort to some predetermined integer scaling factors, e.g., $\times 2$ or $\times 4$.
    Thus, they are difficult to be applied when arbitrary target resolutions are required.
    Recent approaches extend the scope to real-valued upsampling factors, even with varying aspect ratios to handle the limitation.
    In this paper, we propose the SRWarp framework to further generalize the SR tasks toward an arbitrary image transformation.
    We interpret the traditional image warping task, specifically when the input is enlarged, as a spatially-varying SR problem.
    We also propose several novel formulations, including the adaptive warping layer and multiscale blending, to reconstruct visually favorable results in the transformation process.
    Compared with previous methods, we do not constrain the SR model on a regular grid but allow numerous possible deformations for flexible and diverse image editing.
    Extensive experiments and ablation studies justify the necessity and demonstrate the advantage of the proposed SRWarp method under various transformations.
\end{abstract}

\section{Introduction}
As one of the fundamental vision problems, image super-resolution (SR) aims to reconstruct a high-resolution (HR) image from a given low-resolution (LR) input.
The SR methods are widely used in several applications such as perceptual image enhancement~\cite{sr_srgan, sr_esrgan}, editing~\cite{sr_explorable, sr_srflow}, and digital zooming~\cite{wronski2019handheld}, due to its practical importance.
Similar to the other vision-related tasks, recent convolutional neural networks~(CNNs) have achieved promising SR performance with large-scale datasets~\cite{data_div2k, sr_edsr}, efficient structures~\cite{sr_rcan, sr_rdn, sr_dbpn}, and novel optimization techniques~\cite{sr_edsr, sr_esrgan}.
Recent state-of-the-art methods can reconstruct sharp edges and crisp textures with fine details up to $\times 4$ or $\times 8$ scaling factors on various types of input data including real-world images~\cite{sr_zllz, sr_realworld, sr_cdc}, videos~\cite{vsr_edvr, vsr_zooming_slowmo, vsr_starnet, vsr_tdan, vsr_mucan}, hyperspectral~\cite{Fu_2019_CVPR, Zhang_2020_CVPR, yao2020cross}, and light field arrays~\cite{Zhang_2019_CVPR, sr_crossnetpp, wang2020spatial}.

From the perspective of image editing applications, the SR algorithm supports users to effectively increase the number of pixels in the image and reconstruct high-frequency details when its HR counterpart is unavailable.
Such manipulation may include simple resizing with some predefined scaling factors or synthesizing images of arbitrary target resolutions.
However, directly applying existing SR methods in these situations is difficult because the models are usually designed to cope with some fixed integer scales~\cite{sr_srgan, sr_rcan, sr_esrgan}.
Recently, few methods have extended the scope of the SR to upsample a given image by arbitrary scaling factors~\cite{sr_meta} and aspect ratios~\cite{sr_arb}.
These novel approaches provide more flexibility and versatility to existing SR applications for their practical usage.

\begin{figure}[t]
    \vspace{-3mm}
    \centering
    \subfloat[Input]{\includegraphics[width=0.392\linewidth]{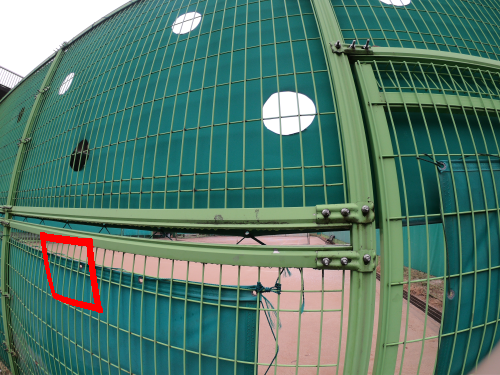}}
    \hfill
    \subfloat[OpenCV]{\includegraphics[width=0.294\linewidth]{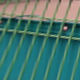}}
    \hfill
    \subfloat[\textbf{SRWarp}]{\includegraphics[width=0.294\linewidth]{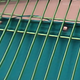}}
    \\
    \figcspace
    \caption{
        \textbf{Real-world lens distortion correction using the proposed SRWarp.}
        The image is captured by the GoPro HERO6 handheld camera.
        Our SRWarp implements SR with locally-varying scale factors, which can be used to transform an input image to the desired geometry.
    }
    \label{fig:teaser}
    \figspace
\end{figure}

Nevertheless, existing SR models are not fully optimized for general image editing tasks due to their intrinsic formulations.
Previous approaches are also designed to take and reconstruct such rectangular frames because digital images are defined on a rectangular grid.
On the contrary, images may undergo various deformations in practice to effectively adjust their contents within the context.
For instance, homographic transformation~\cite{zhang2020content, le2020deep} aligns images from different views, and cameras incorporate various correction algorithms to remove distortions from the lens~\cite{swaminathan2000nonmetric, xue2019learning}.
One of the shortcomings in the conventional warping methods is that interpolation-based algorithms tend to generate blurry results when a local region of the image is stretched.
Hence, an appropriate enhancement algorithm is required to preserve sharp edges and detailed textures, as the SR methods for image upscaling.
However, such applications may require images to be processed on irregular grids, which cannot be handled by na\"ive CNNs for regular-shaped data.

To prevent blurry warping results, state-of-the-art SR models may be introduced prior to image transformation.
By doing so, supersampled pixels alleviate blurriness and artifacts from simple interpolation.
However, Hu~\etal~\cite{sr_meta} and Wang~\etal~\cite{sr_arb} have demonstrated that the solution is suboptimal in the arbitrary-scale SR task.
Furthermore, a generalized warping algorithm should deal with more complex and even spatially-varying deformations, which are not straightforward to be considered in the previous approaches.
%
%
Therefore, an appropriate solution is required to effectively combine the SR and warping methods in a single pipeline.

In this study, we interpret the general image transformation task as a spatially-varying SR problem.
For such purpose, we construct an end-to-end learnable framework, that is, SRWarp.
Different from the previous SR methods, our method is designed to handle the warping problem in two specific aspects.
First, we introduce an adaptive warping layer (AWL) to dynamically predict appropriate resampling kernels for different local distortions.
Second, our multiscale blending strategy combines features of various resolutions based on their contents and local deformations to utilize richer information from a given image.
With powerful backbones~\cite{sr_edsr, sr_esrgan}, the proposed SRWarp can successfully reconstruct image structures that can be missed from conventional warping methods, as shown in \figref{fig:teaser}.
Our contributions can be organized in threefolds as follows:
\begin{compactitem}[$\bullet$]
    \item The novel SRWarp model generalizes the concept of SR under arbitrary transformations and formulates a framework to learn image transformation.
    \item Extensive analysis shows that our adaptive warping layer and multiscale blending contribute to improving the proposed SRWarp method.
    \item Compared with existing methods, our SRWarp model reconstruct high-quality details and edges in transformed images, quantitatively and qualitatively.
\end{compactitem}
\section{Related Work}
\Paragraph{Conventional deep SR.}
After Dong~\etal~\cite{sr_srcnn} has successfully applied CNNs to the SR task, numerous approaches have been studied toward better reconstruction.
VDSR~\cite{sr_vdsr} is one of the most influential works which introduces a novel residual learning strategy to enable faster training and very deep SR network architecture.
ESPCN~\cite{sr_espcn} constructs an efficient pixel-shuffling layer to implement a learnable upsampling module.
LapSRN~\cite{sr_lapsrn} architecture efficiently handles the multiscale SR task using a laplacian upscaling pyramid.
Ledig~\etal~\cite{sr_srgan} have adopted the residual block from high-level image classification task~\cite{net_residual} to implement SRResNet and SRGAN models.
With increasing computational resources, state-of-the-art methods such as EDSR~\cite{sr_edsr} have focused on larger and more complex network structures, producing high quality images.
%
Recently, several advances in neural network designs such as attention~\cite{sr_rcan, sr_san, sr_han}, back-projection~\cite{sr_dbpn}, and dense connections~\cite{sr_rdn, ir_rdn, sr_dbpn, sr_esrgan} have made it possible to reconstruct high-quality images very efficiently.

\Paragraph{SR for arbitrary resolution.}
%
Most conventional SR methods~\cite{sr_srcnn, sr_vdsr} have relied on na\"ive interpolation to enlarge a given LR image before Shi~\etal~\cite{sr_espcn} have introduced the pixel-shuffling layer for learnable upscaling.
For example, VDSR~\cite{sr_vdsr} upscales the LR image to their target resolution and then applies the SR model to refine local details and textures so that the method can serve as an arbitrary-scale SR framework.
However, a significant drawback is that extensive computations are required proportional to the output size.
Therefore, subsequent methods have been specialized for some fixed integer scales, e.g., $\times 2$ or $\times 4$, which are commonly used in various applications.

Recently, Hu~\etal~\cite{sr_meta} have proposed the meta-upscale module to replace scale-specific upsampling layers in previous approaches.
Meta-SR~\cite{sr_meta} is designed to utilize dynamic filters to deal with real-valued upscaling factors. 
Subsequently, Wang~\etal~\cite{sr_arb} introduce scale-aware features and upsampling modules to reconstruct images of arbitrary target resolutions.
The previous methods mainly consider the SR task along horizontal or vertical axes.
However, our differentiable warping module in SRWarp allows images to be transformed into any shape.

\Paragraph{Irregular spatial sampling in CNNs.}
Pixels in a digital image are uniformly placed on a 2D rectangle.
However, objects may appear in arbitrary shapes and orientations in the image, making it challenging to handle them with a simple convolution.
To overcome the limitation, the spatial transformer networks~\cite{jaderberg2015spatial} estimate appropriate warping parameters to compensate for possible deformations in the input image.
Rather than transform the image, deformable convolutions~\cite{dai2017deformable} and active convolutions~\cite{jeon2017active} predict input-dependent kernel offsets and modulators~\cite{zhu2019deformable} to perform irregular spatial sampling.
Furthermore, the deformable kernel~\cite{gao2019deformable} approach resamples filter weights to adjust the effective receptive field adaptively.
Recent state-of-the-art image restoration models, especially with temporal data, introduce the irregular sampling strategy for accurate alignment~\cite{vsr_tdan, vsr_edvr, vsr_zooming_slowmo}.
However, our approach is the first novel attempt to interpret image warping as an SR problem.
\section{Method}
We introduce our generalized SR framework, namely SRWarp, in detail.
$\img{LR} \in \mathbb{R}^{H \times W}$, $\img{HR} \in \mathbb{R}^{H' \times W'}$, and $\img{SR} \in \mathbb{R}^{H' \times W'}$ represent source LR, ground-truth HR, and target super-resolved images, respectively.
$H \times W$ and $H' \times W'$ correspond to image resolutions, and we omit RGB color channels for simplicity.
Different from the conventional SR, the target resolution $H' \times W'$ varies depending on the given transformation.
We define its resolution using a bounding box of the image because the warping may produce irregular output shapes rather than rectangular.
For more detailed descriptions and analysis of this section, please refer to our supplementary material.

\subsection{Super-Resolution Under Homography}
\label{ssec:generalize}
Given a $3 \times 3$ projective homography matrix $M$ and a point $p = \paren{ x, y, 1 }\transpose$ on the source image, we calculate the target homogeneous coordinate $p' = w' \paren{ x', y', 1 }\transpose$ as $Mp = p'$ or $f_M \paren{ x, y } = \paren{ x', y' }$, where $f_M$ is a corresponding functional representation.
In the backward warping, $p = M^{-1} p'$ is calculated instead for each output pixel $p'$ to remove cavities.
%
If we simply scale the image along $x$ and $y$ axes, then the matrix $M$ is defined as follows:
\begin{equation}
    M_{s_x s_y} =
    \begin{pmatrix}
        s_x & 0 & 0.5 \paren{ s_x - 1 } \\
        0 & s_y & 0.5 \paren{ s_y - 1 } \\
        0 & 0 & 1
    \end{pmatrix},
    \label{eq:scale_matrix}
\end{equation}
where $s_x$ and $s_y$ correspond to scale factors along the axes, and the translation components, i.e., $0.5 \paren{s_\ast - 1}$, compensate subpixel shift to ensure an accurate alignment~\cite{sr_arb}.
Most early SR methods are designed to deal with the case where $s_x = s_y$ represents predefined integers~\cite{sr_espcn, sr_srgan, sr_dbpn, sr_esrgan}.
Recent approaches have relaxed such constraint by allowing arbitrary real numbers~\cite{sr_meta, sr_arb}.
However, numerous possible forms of $M$ remain unexplored.

\begin{figure}
    \centering
    \includegraphics[width=\linewidth]{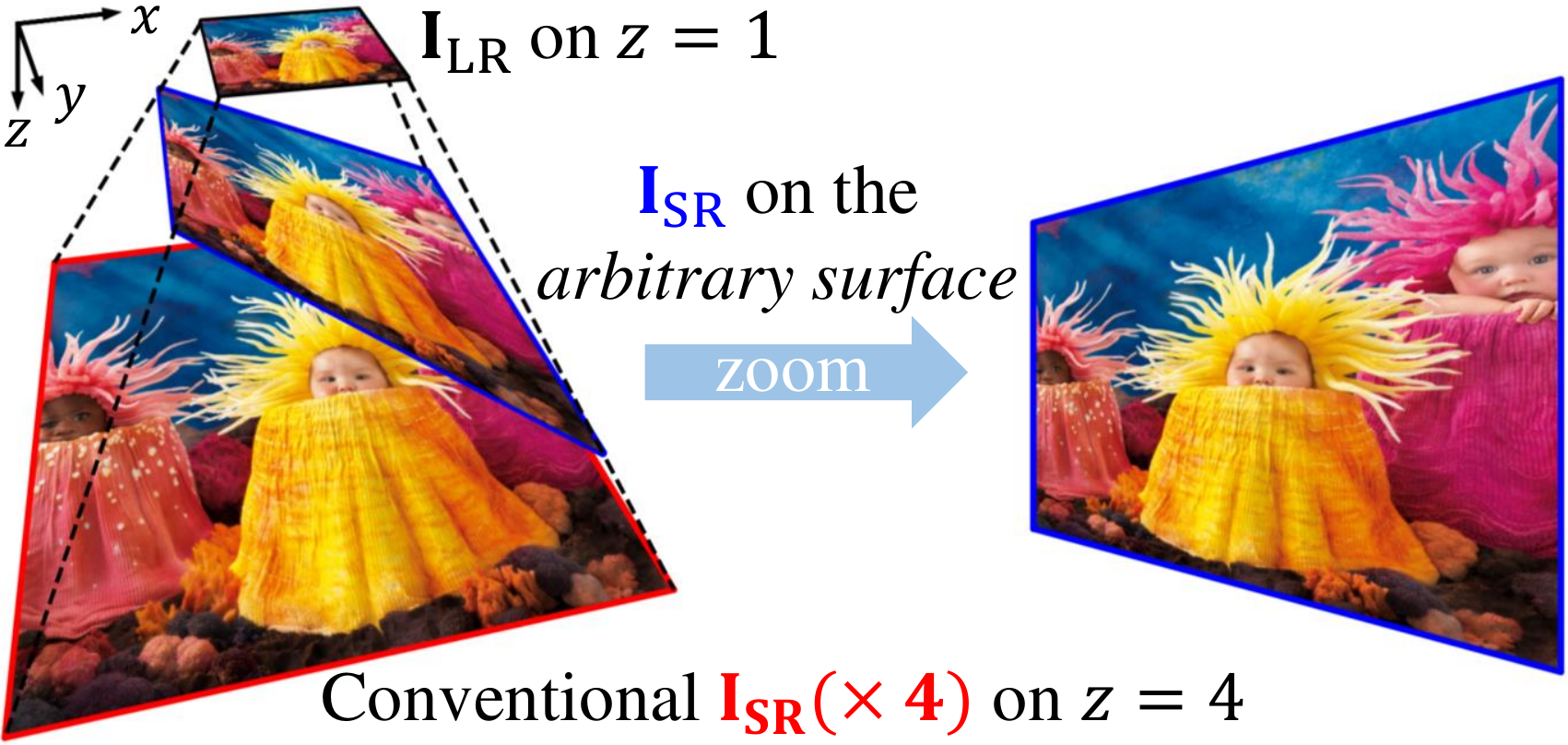}
    \\
    \figcspace
    \caption{
        \textbf{Concept of the generalized SR.}
        Dotted lines represent the upscale image pyramid in $xyz$-coordinate.
        %
        While $\times s$ super-resolved images from conventional methods~(\textcolor{red}{\textbf{red}}) exist on the plane $z = s$ only, our results~(\textcolor{blue}{\textbf{blue}}) exists on any arbitrary cutting surfaces of the pyramid.
        We note that the aspect ratio of the pyramid can be varied as well.
    }
    \label{fig:pyramid}
    \figspace
\end{figure}

\figref{fig:pyramid} presents a concept of the conventional SR methods and our SRWarp method from the perspective of the image scale pyramid.
For convenience, we assume that an LR image $\img{LR}$ is placed on the plane $z = 1$.
Then, obtaining the $\times s$ SR result $\img{SR}$ is equivalent to slicing the pyramid by the plane $z = s$, where all points on the SR image have the same $z$-coordinate.
Previous methods aim to learn the image representations that are parallel to the given LR input.
However, slicing the image pyramid with an arbitrary plane, or even general surfaces in the space, is also possible.
Therefore, We propose to redefine the warping problem as a generalized SR task with spatially-varying scales and even aspect ratios because the pixels in the resulting image can have different $z$ values depending on their positions.

\subsection{Adaptive Warping Layer}
\label{ssec:elliptical}
Image warping consists of two primitive operations, namely, mapping and resampling.
%
%
The mapping initially determines the spatial relationship between input and output images.
For a target position $p' = \paren{ x', y' }$, the corresponding source pixel is located as $p = \paren{ x, y } = f_M^{-1} \paren{ x', y' }$.
We omit the homogeneous representation for simplicity.
While pixels in digital images are placed on integer coordinates only, $x$ and $y$ may have arbitrary real values depending on the function $f^{-1}_M$.
%
%
%
Therefore, an appropriate resampling is required to obtain a plausible pixel value as follows:
%
%
\begin{equation}
    \mathbf{W} \paren{ x', y' } = \sum_{i, j = a}^{ b } { \mathbf{k} \paren{ x', y', i, j } \mathbf{F} \paren{ \round{x} + i, \round{y} + j } },
    \label{eq:kernel}
\end{equation}
where $\round{ \cdot }$ is a rounding operator, $\mathbf{F} \in \mathbb{R}^{H \times W}$ is an input, $\mathbf{W} \in \mathbb{R}^{H' \times W'}$ is an output, and $\mathbf{k}$ is a point-wise interpolation kernel, respectively.
$a$ and $b$ are boundary indices of the $k \times k$ window, where $k = b - a + 1$.
For example, we set $a = -1$ and $b = 1$ for standard $3 \times 3$ kernels.

\begin{figure}[t]
    \centering
    \vspace{-3mm}
    \definecolor{mygreen}{RGB}{0, 176, 80}
    \definecolor{myblue}{RGB}{0, 112, 192}
    \subfloat[Regular resampling\label{fig:example_area_regular}]{\includegraphics[width=0.495\linewidth]{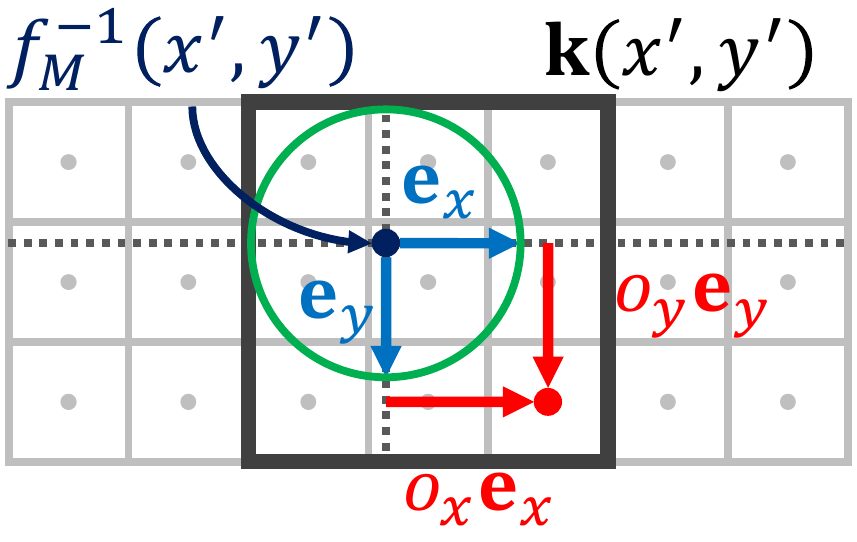}}
    \subfloat[Adaptive resampling\label{fig:example_area_irregular}]{\includegraphics[width=0.495\linewidth]{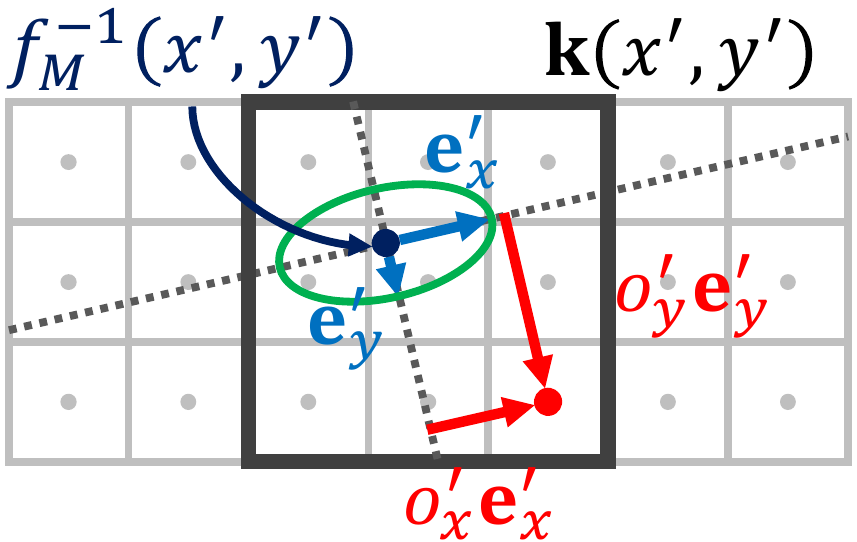}}
    \\
    \figcspace
    \caption{
        \textbf{Example of the adaptive grid.}
        Each point represents a pixel on the source domain $\mathbf{F}$.
        (a) On a regular grid, the resampling bases $\mathbf{e}_x$ and $\mathbf{e}_y$ in \textcolor{myblue}{\textbf{blue}} are orthonormal and aligned with the source image.
        (b) We adopt irregular bases $\mathbf{e}'_x$ and $\mathbf{e}'_y$ with varying lengths and orientations for each target position $\paren{ x', y' }$.
        The relative offset vector $\paren{ o_x, o_y }$ of an example point in \textcolor{red}{\textbf{red}} is mapped to $\paren{ o'_x, o'_y }$ following the change of basis.
        The \textcolor{mygreen}{\textbf{green}} ellipse illustrates how a unit circle on the target image is projected to the source domain.
    }
    \label{fig:example_area}
    \figspace
\end{figure}

Conventional resampling algorithms introduce a fixed sampling coordinate and kernel function to calculate the weight $\mathbf{k}$, regardless of the transformation $M$.
For example, a widely-used bicubic warping initially calculates a relative offset $\paren{ o_x, o_y }$ of each point in the $k \times k$ window with respect to $\paren{x, y}$ as shown in \figref{fig:example_area_regular} and constructs $\mathbf{k}$ using a cubic spline.
However, due to the diversity of possible transformations, such formulation may not be optimal in several aspects.
First, it is difficult to consider the transformed geometry where the target image is not defined on a rectangular grid.
%
%
Second, the fixed kernel function limits generalizability, while recent SR models prefer learnable upsampling~\cite{sr_espcn, sr_meta, sr_arb} rather than the predetermined one~\cite{sr_vdsr}.
To handle these issues, we propose an adaptive warping layer (AWL) so that the resampling kernel $\mathbf{k}$ can be trained to consider local deformations.

To determine an appropriate sampling coordinate for each target position $\paren{ x', y' }$, we linearize the backward mapping $M^{-1}$ at the point with the Jacobian $J \paren{ x', y' } = \begin{pmatrix} \mathbf{u}\transpose & \mathbf{v}\transpose \end{pmatrix}$.
Specifically, we calculate $\mathbf{u}$ and $\mathbf{v}$ as follows:
\begin{equation}
    \begin{split}
        \mathbf{u} &= \frac{f^{-1}_{M} \paren{ x' + \epsilon, y'} - f_{M}^{-1} \paren{ x' - \epsilon, y'}}{2 \epsilon}, \\
        \mathbf{v} &= \frac{f^{-1}_{M} \paren{ x', y' + \epsilon} - f^{-1}_{M} \paren{ x', y' - \epsilon}}{2 \epsilon}, \\
    \end{split}
    \label{eq:approximation}
\end{equation}
where $f^{-1}_{M} = f_{M^{-1}}$ and $\epsilon = 0.5$.
We project a unit circle centered around $\paren{ x', y' }$ on the target domain to an ellipse~\cite{ewa} on the source image, using the local approximation.
Then, we calculate two principal axes $\mathbf{e'}_x$ and $\mathbf{e'}_y$ of the ellipse.
As shown in \figref{fig:example_area_irregular}, the relative offset vector $\mathbf{o} = \paren{ o_x, o_y }$ is represented as $\mathbf{o'} = \paren{ o'_x, o'_y }$ under the new locally adaptive coordinate system.
%
%
In the resampling process, the actual contribution of each point is calculated with respect to the distance from the origin.
Therefore, we utilize the adaptive coordinate to adjust the point considering local distortions.
The original offset vectors are rescaled as $\frac{\lVert \mathbf{o'} \rVert }{\lVert \mathbf{o} \rVert} \mathbf{o}$ and used to calculate the kernel $\mathbf{k}$.
%

Subsequently, we introduce a kernel estimator $\mathcal{K}$ to estimate adaptive resampling weights $\mathbf{k}$.
Similar to the conventional interpolation functions, it takes $k^2$ offset vectors to determine the contributions of each point $\mathbf{F} \paren{ x, y }$ in the window $\mathbf{k} \paren{ x', y' }$.
However, we adopt a series of fully-connected layers~\cite{sr_meta, sr_arb} to learn the function rather than using a predetermined one.
The learnable network allows considering local deformations and generates appropriate dynamic filters~\cite{jia2016dynamic, vsr_duf} for a given transformation.
We construct the proposed AWL $\mathcal{W}$ by combining the adaptive resampling grid and kernel prediction layer $\mathcal{K}$, as follows:
\begin{equation}
    \mathbf{W} = \mathcal{W} \paren{ \mathbf{F}, f_{M} }.
    \label{eq:awl}
\end{equation}

\subsection{Multiscale Blending}
\label{ssec:ms_warping}
\figref{fig:pyramid} illustrates that images under the generalized SR task suffer distortions with spatially-varying scaling factors.
Therefore, multiscale representations can play an essential role in reconstructing high-quality images.
To effectively utilize the property, we further introduce a blending method for the proposed SRWarp framework.

\begin{figure*}[t]
    \centering
    \includegraphics[width=\linewidth]{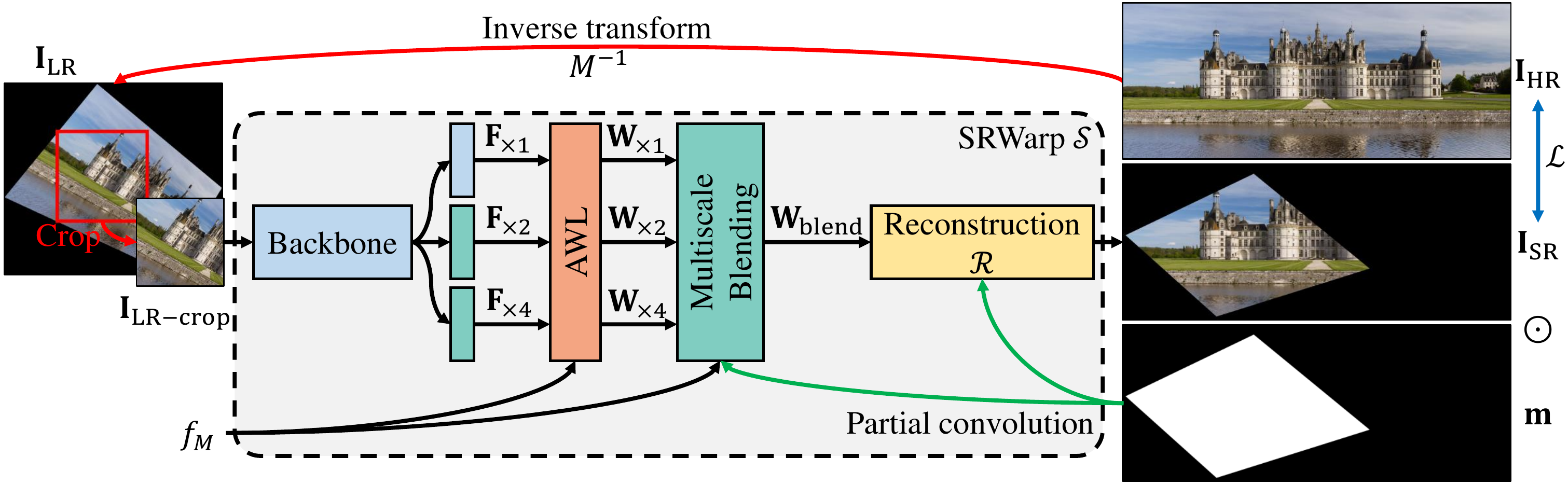}
    \\
    \figcspace
    \caption{
        \textbf{Overall organization of the proposed SRWarp model.}
        More detailed architectures are described in our supplementary material.
        Black regions outside the warped image $\img{SR}$ represent void pixels that are ignored.
    }
    \label{fig:architecture}
    \figspace
\end{figure*}

\Paragraph{Multiscale feature extractor.}
We define a scale-specific feature extractor $\mathcal{F}_{\times s}$ with a fixed integer scaling factor of $s$, which adopts the state-of-the-art SR architectures~\cite{sr_edsr, sr_esrgan}.
Given an LR image $\img{LR}$, the module extracts the scale-specific feature $\mathbf{F}_{\times s} \in \mathbb{R}^{C \times sH \times sW}$, where $C$ denotes the number of output channels.
While it is possible to separate the network for each scaling factor, we adopt a shared feature extractor with multiple upsampling layers~\cite{sr_edsr} in practice for several reasons.
For instance, previous approaches have demonstrated that multiscale representations can be jointly learned~\cite{sr_vdsr, sr_lapsrn, lai2018fast, sr_arb} within a single model.
Also, using the shared backbone network is computationally efficient compared to applying multiple different models to extract spatial features.
From the state-of-the-art SR architecture, we replace the last upsampling module with $\times 1$, $\times 2$, and $\times 4$ feature extractor to implement our multiscale backbone as shown in \figref{fig:architecture}.

\Paragraph{Multiscale warping and blending.}
For each scale-specific feature $\mathbf{F}_{\times s}$, we construct the corresponding transformation as $M M_{s^{-1} s^{-1}}$ by using \eqref{eq:scale_matrix}.
As a result, features of different resolutions can be mapped to a fixed spatial dimension, i.e., $H' \times W'$.
We use a term $\mathbf{W}_{\times s} \in \mathbb{R}^{C \times H' \times W'}$ to represent the warped features as follows:
\begin{equation}
    \mathbf{W}_{\times s} = \mathcal{W} \paren{ \mathcal{F}_{\times s} \paren{ \img{LR} }, MM_{s^{-1} s^{-1}} }.
    \label{eq:ms_warping}
\end{equation}
Then, the output SR image $\img{SR}$ can be reconstructed from a set of the multiscale warped features $\left\{ \mathbf{W}_{\times s} | s = s_0, s_1, \cdots \right\}$.
However, a simple combination, e.g., averaging or concatenation, of those features may not reflect the spatially-varying property of the generalized SR problem.
Therefore, we introduce a multiscale blending module to combine information from different resolutions effectively.
To determine appropriate scales for each local region, image contents play a critical role.
For example, low-frequency components are preferred in the warping process to prevent aliasing and undesirable artifacts for plain regions.
On the contrary, high-frequency details are considered to represent edges and textures accurately.
Therefore, we use learnable scale-specific and global content feature extractors $\mathcal{C}_{\times s}$ and $\mathcal{C}$ as follows:
\begin{equation}
    \mathbf{C} = \mathcal{C} \paren{ \mathcal{C}_{\times s_0} \paren{ \mathbf{W}_{\times s_0} }, \mathcal{C}_{\times s_1} \paren{ \mathbf{W}_{\times s_1} }, \cdots },
    \label{eq:sampling_contents}
\end{equation}
where the global content feature $\mathbf{C} \in \mathbb{R}^{C \times H' \times W'}$ is represented by scale-specific representations $\mathcal{C}_{\times s_i} \paren{ \mathbf{W}_{\times s_i}}$.

Since our SRWarp method handles spatially-varying distortions, appropriate feature scales may also depend on the local deformation.
The proposed model may benefit from the degree of transformation around the pixel to determine the contributions of each multiscale representation.
Therefore, we acquire the scale feature $\mathbf{S} \in \mathbb{R}^{H' \times W'}$ as follows:
\begin{equation}
    \mathbf{S} \paren{ x', y' } = -\log{ \lvert \det \paren{ J \paren{ x', y' } } \rvert }.
    \label{eq:sampling_scale}
\end{equation}
Physically, the determinant of the Jacobian describes a local magnification factor of the transformation.
When adopting backward mapping, we consider the reciprocal of the Jacobian determinant and normalize it by taking the natural log.

Our multiscale blending module then applies $1 \times 1$ convolutions to the concatenated content and scale features $\mathbf{C}$ and $\mathbf{S}$.
By doing so, appropriate blending weights $w_{\times s}$ are determined for each output position $\paren{ x', y' }$.
The blended features $\mathbf{W}_\text{blend}$ can be represented as follows:
\begin{equation}
    \mathbf{W}_\text{blend} = \sum_s {w_{\times s} \odot \mathbf{W}_{\times s}},
    \label{eq:blending}
\end{equation}
where $\odot$ is an element-wise multiplication.

\Paragraph{Partial convolution.}
Image warping can produce void pixels on the target coordinate when the point is mapped to outside the source image.
Such regions may negatively affect the model performance because conventional CNNs consider all pixels equally.
To efficiently deal with the problem, we define a 2D binary mask $\mathbf{m}$ as follows:
\begin{equation}
    \mathbf{m} \paren{ x', y' } =
    \begin{cases}
        0, & \text{if}\ \paren{ x, y } \text{is outside of}\ \mathbf{F}_{\times{1}}, \\
        1, & \text{otherwise},
    \end{cases}
    \label{eq:mask}
\end{equation}
where $f_M \paren{ x, y } = \paren{ x', y' }$.
We calculate the mask $\mathbf{m}$ from the $\times 1$ feature $\mathbf{F}_{\times 1}$ and share it across scales to maintain consistency between different resolutions.
Then, we adopt the partial convolution~\cite{liu2018partialinpainting, liu2018partialpadding} for our content feature extractors $\mathcal{C}$ and $\mathcal{C}_{\times s}$, using the mask to ignore void pixels.

\subsection{SRWarp}
\label{ssec:srwarp}
%
Finally, we introduce a reconstruction module $\mathcal{R}$ with five residual blocks~\cite{sr_srgan, sr_edsr}.
%
%
We combine the SR backbone, AWL, blending, and reconstruction modules to construct the SRWarp model $\mathcal{S}$ as shown in \figref{fig:architecture}.
For stable training, the residual connection~\cite{sr_vdsr} is incorporated as $\img{SR} = \mathcal{R} \paren{ \mathbf{W}_\text{blend} } + \img{bic}$, where $\img{bic}$ is a warped image using bicubic interpolation and $\img{SR}$ is the final output.
Given a set of training input and target pairs $\paren{ \img{LR}^n, \img{HR}^n }$, we minimize an average $L_1$ loss~\cite{sr_lapsrn, sr_edsr} $\mathcal{L}$ between the reconstructed and ground-truth~(GT) images as follows:
\begin{equation}
    \mathcal{L} = \frac{1}{N} \sum_{n = 1}^{N} \frac{ 1 }{ \normzero{ \mathbf{m}} } \normone{ \mathbf{m} \odot \paren{ \mathcal{S} \paren{ \img{LR}^n, f_M } - \img{HR}^n } },
    \label{eq:objective}
\end{equation}
where $N = 4$ is the number of samples in a mini-batch, $n$ is a sample index, 0-norm $\normzero{ \cdot }$ represents the number of nonzero values, and $\mathcal{S} \paren{ \img{LR}^{n}, f_M } + \img{bic}^n = \img{SR}^{n} $, respectively.
The transform function $f_M$ is shared in a single mini-batch for efficient calculation.
The binary mask $\mathbf{m}$ in \eqref{eq:mask} prevents backward gradients from being propagated from void pixels.
The proposed SRWarp model can be trained in an end-to-end manner with the ADAM~\cite{others_adam} optimizer.
\section{Experiments}
\label{sec:exp}
We adopt two different SR networks as a backbone of the multiscale feature extractor for the proposed SRWarp model.
The modified MDSR~\cite{sr_edsr} architecture serves as a smaller baseline, whereas RRDB~\cite{sr_esrgan} with customized multiscale branches (MRDB) provides a larger backbone for improved performance.
We describe more detailed training arguments in our supplementary material.
%
%
PyTorch codes with an efficient CUDA implementation and dataset will be publicly available from the following repository: \href{https://github.com/sanghyun-son/srwarp}{https://github.com/sanghyun-son/srwarp}.

\subsection{Dataset and Metric}
\label{sec:data_metric}
\Paragraph{Dataset.}
In conventional image SR methods, acquiring real-world LR and HR image pairs is very challenging due to several practical issues, such as outdoor scene dynamics and subpixel misalignments~\cite{sr_realworld, sr_camera, sr_zllz}.
Similarly, collecting high-quality image pairs with corresponding transformation matrices in the wild for our generalized SR task is also difficult.
Therefore, we propose the DIV2K-Warping (DIV2KW) dataset by synthesizing LR samples from the existing DIV2K~\cite{data_div2k} dataset to train our SRWarp model in a supervised manner.
We first assign 500, 100, and 100 random warping parameters $\left\{ M_{i} \right\}$ for training, validation, and test, respectively.
Each matrix is designed to include random upscaling, sheering, rotation, and projection because we mainly aim to enlarge the given image.
We describe more details in our supplementary material.

During the learning phase, we randomly sample square HR patches from 800 images in the DIV2K training dataset and one warping matrix $M_{i}^{-1}$ to construct a ground-truth batch $\img{HR}$.
Then, we warp the batch with $M_{i}^{-1}$ to obtain corresponding LR inputs $\img{LR}$.
For efficiency, the largest valid square from the transformed region is cropped for the input $\img{LR-crop}$.
With the transformation matrix $M_{i}$ and LR patches $\img{LR-crop}$, we optimize our warping model to reconstruct the original image $\img{HR}$ as described in \ref{ssec:srwarp}.
Figure~\ref{fig:architecture} illustrates the actual training pipeline regarding our SRWarp model.
We use 100 images from the DIV2K valid dataset with different transformation parameters following the same pipeline to evaluate our method.

\Paragraph{Metric.}
We adopt a traditional PSNR metric on RGB color space to evaluate the quality of warped images.
However, we only consider valid pixels in a $H' \times W'$ grid similar to our training objective in \eqref{eq:objective} because they have irregular shapes rather than standard rectangles.
The modified PSNR with a binary mask $\mathbf{m}$ (mPSNR) is described as follows:
\begin{equation}
    \begin{split}
        \text{mPSNR(dB)} = 10 \log_{10}{ 
            \frac{
                \normzero{ \mathbf{m} }
            }{
                \normtwo{ \mathbf{m} \odot \paren{ \img{SR} - \img{HR} } }
            }
        },
    \end{split}
    \label{eq:mpsnr}
\end{equation}
where images $\mathbf{I}_\ast$ are normalized between 0 and 1.

\subsection{Ablation Study}
\label{ssec:ablation}
We extensively validate possible combinations of the proposed modules in \tabref{tab:baseline} because they are orthogonal to each other.
The modified baseline MDSR~\cite{sr_edsr} structure is used as a backbone by default for lightweight evaluations.
We refer to the model with a single-scale SR backbone~\cite{sr_edsr, sr_esrgan} and standard warping layer at the end as a baseline.
Sequences of A, M, and R represent corresponding configurations, e.g., A-R for the one which shows 32.19dB mPSNR in \tabref{tab:baseline}.
Our SRWarp model is represented as A-M-R and achieves 32.29dB in \tabref{tab:baseline}.
More details are described in our supplementary material.

\Paragraph{Adaptive warping layer.}
\tabref{tab:baseline} demonstrates that AWL introduces consistent performance gains by providing spatially-adaptive resampling kernels.
\tabref{tab:ablation_warping} extensively compares possible implementations of the AWL in the proposed SRWarp method.
We replace the regular resampling grid from M-R in \tabref{tab:baseline} to the spatially-varying representations (Adaptive in \tabref{tab:ablation_warping}) as shown in \figref{fig:example_area_irregular}.
However, because the resampling weights $\mathbf{k}$ are not learnable, the formulation does not bring an advantage even when the spatially-varying property is considered.
%
Introducing a kernel estimator to the regular grid (Layer in \tabref{tab:ablation_blending}) yields +0.06dB of mPSNR gain over the M-R method.
The performance is further improved to 32.29dB (A-M-R in \tabref{tab:baseline}) by combining the spatially-varying coordinates and trainable module.
We note that the adaptive resampling grid at each output position does not require any additional parameters and can be calculated efficiently.

\begin{table}[t]
    \centering
    \begin{tabularx}{\linewidth}{>{\centering\arraybackslash}X >{\centering\arraybackslash}X >{\centering\arraybackslash}X c c}
        \toprule
        A & M & R & B & mPSNR$^\uparrow$(dB) on $\text{DIV2KW}_\text{Valid.}$ \\
        \midrule
        $-$ & $-$ & $-$ & \multirow{4}{*}{EDSR} & 31.36 (+0.00) \\
        \ding{51} & $-$ & $-$ & & 32.06 (+0.70) \\
        $-$ & $-$ & \ding{51} & & 32.08 (+0.72) \\
        \ding{51} & $-$ & \ding{51} & & 32.19 (+0.83) \\
        \midrule
        $-$ & \ding{51} & $-$ & \multirow{4}{*}{MDSR} & 32.19 (+0.83) \\
        \ding{51} & \ding{51} & $-$ & & 32.21 (+0.85) \\
        $-$ & \ding{51} & \ding{51} & & 32.19 (+0.83) \\
        \ding{51} & \ding{51} & \ding{51} & & \textbf{32.29 (+0.93)} \\
        \midrule
        $-$ & $-$ & $-$ & RRDB & 31.64 (+0.28) \\
        \ding{51} & \ding{51} & \ding{51} & MRDB & \textbf{32.56 (+1.20)} \\
        \bottomrule
    \end{tabularx}
    \\
    \tabcspace
    \caption{
        \textbf{Contributions of each module in our SRWarp method.}
        A, M, R, and B denote the adaptive warping layer (AWL), multiscale blending, reconstruction module, and backbone architecture, respectively.
        Numbers in parentheses indicate performance gains over the baseline on top.
    }
    \label{tab:baseline}
    \tabspace
    \vspace{-1.5mm}
\end{table}
\begin{table}[t]
    \centering
    \subfloat[Warping\label{tab:ablation_warping}]{
        \begin{tabularx}{0.495\linewidth}{p{1.5cm} >{\centering\arraybackslash}X}
            \toprule
            Method & mPSNR$^\uparrow$(dB) \\
            \midrule
            Adaptive & 32.19 \\
            Layer & 32.25 \\
            AWL-SS & 32.23 \\
            AWL-MS & 32.24 \\
            \bottomrule
        \end{tabularx}    
    }
    \subfloat[Blending\label{tab:ablation_blending}]{
        \begin{tabularx}{0.495\linewidth}{p{1.5cm} >{\centering\arraybackslash}X}
            \toprule
            Method & mPSNR$^\uparrow$(dB)\\
            \midrule
            Average & 32.26 \\
            Concat. & 32.19 \\
            w/o $\mathbf{C}$ & 32.24 \\
            w/o $\mathbf{S}$ & 32.23 \\
            \bottomrule
        \end{tabularx}    
    }
    \\
    \tabcspace
    \caption{
        \textbf{Effects of warping and blending strategies in our SRWarp model.}
        %
        %
        We evaluate each method on the $\text{DIV2KW}_\text{Valid.}$ dataset.
        The proposed SRWarp achieves the mPSNR of 32.29dB under the same environment.
    }
    \label{tab:ablation}
    \tabspace
\end{table}

We also analyze two possible variants of our AWL.
AWL-SS in \tabref{tab:ablation_warping} shares the kernel estimator $\mathcal{K}$ across scales and channel dimensions, even with the multiscale SR backbone.
AWL-MS in \tabref{tab:ablation_warping} achives a minor performance gain of +0.01dB by utilizing scale-specific modules $\mathcal{K}_{\times s}$ as described in \secref{ssec:ms_warping}.
In the proposed SRWarp model, we further estimate the kernels in a depthwise manner, i.e., $C \times k \times k$ weights for each $\paren{ x', y' }$, and achieve an additional +0.05dB improvement in the mPSNR metric.

\begin{table}[t]
    \centering
    \begin{tabularx}{\linewidth}{p{0.3cm} l p{0.3cm} c}
        \toprule
        \multicolumn{3}{c}{Method} & mPSNR$^\uparrow$(dB) on $\text{DIV2KW}_\text{Test}$ \\
        \midrule
        \multicolumn{3}{c}{\cv{cv2} (Bicubic)~\cite{opencv_library}} & 27.85 (-2.41) \\
        \midrule
        \multirow{3}{*}{\makecell{$\times 2$}} & RDN~\cite{sr_rdn} & \multirow{3}{*}{\makecell{$+$\cv{cv2}}} & 30.22 (+0.00) \\
        & EDSR~\cite{sr_edsr} & & 30.42 (+0.20) \\
        & RCAN~\cite{sr_rcan} & & 30.45 (+0.23) \\
        \midrule
        \multirow{4}{*}{\makecell{$\times 4$}} & RDN~\cite{sr_rdn} & \multirow{4}{*}{\makecell{$+$\cv{cv2}}} & 30.50 (+0.28) \\
        & EDSR~\cite{sr_edsr} & & 30.66 (+0.44) \\
        & RCAN~\cite{sr_rcan} & & 30.71 (+0.49)\\
        & RRDB~\cite{sr_esrgan} & & \secondbest{30.76 (+0.54)} \\
        \midrule
        \multicolumn{3}{c}{\textbf{SRWarp~(MRDB)}} & \best{31.04 (+0.82)} \\
        \bottomrule
    \end{tabularx}
    \\
    \tabcspace
    \caption{
        \textbf{Comparison between our SRWarp and available warping methods.}
        $+$ \cv{cv2} denotes that we first apply a scale-specific SR model for supersampling and then transform the upscaled image with the traditional warping algorithm.
        Numbers in parenthesis denote performance gain over the $\times 2$ RDN $+$ \cv{cv2} method.
        The best and second-best performances are \best{bolded} and \secondbest{underlined}, respectively.
    }
    \label{tab:comp}
    \tabspace
\end{table}
\begin{figure*}[t]
    \centering
    \subfloat{\includegraphics[width=0.38\linewidth]{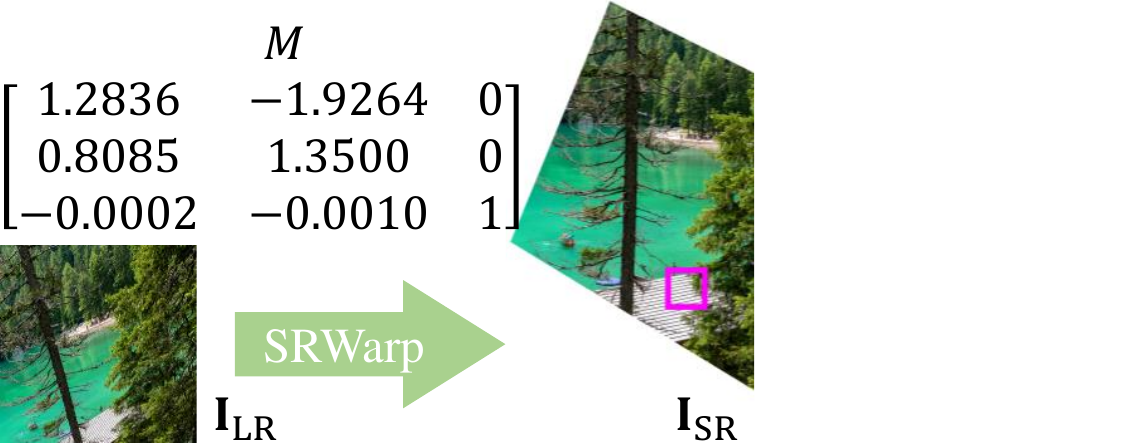}}
    \hfill
    \subfloat{\includegraphics[width=0.15\linewidth]{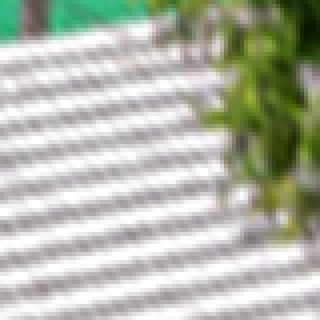}}
    \hfill
    \subfloat{\includegraphics[width=0.15\linewidth]{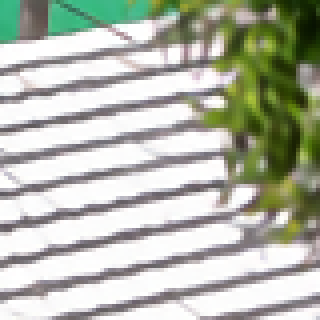}}
    \hfill
    \subfloat{\includegraphics[width=0.15\linewidth]{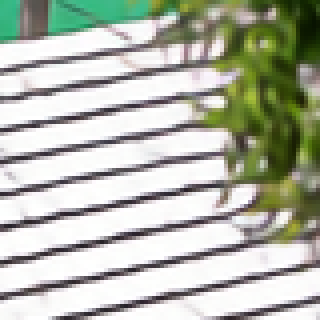}}
    \hfill
    \subfloat{\includegraphics[width=0.15\linewidth]{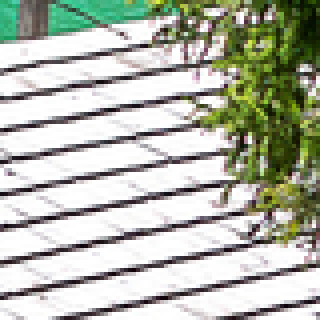}}
    \\
    \vspace{-3mm}
    \addtocounter{subfigure}{-5}
    \subfloat[$\img{LR}$, $\img{SR}$, and $M$]{\includegraphics[width=0.38\linewidth]{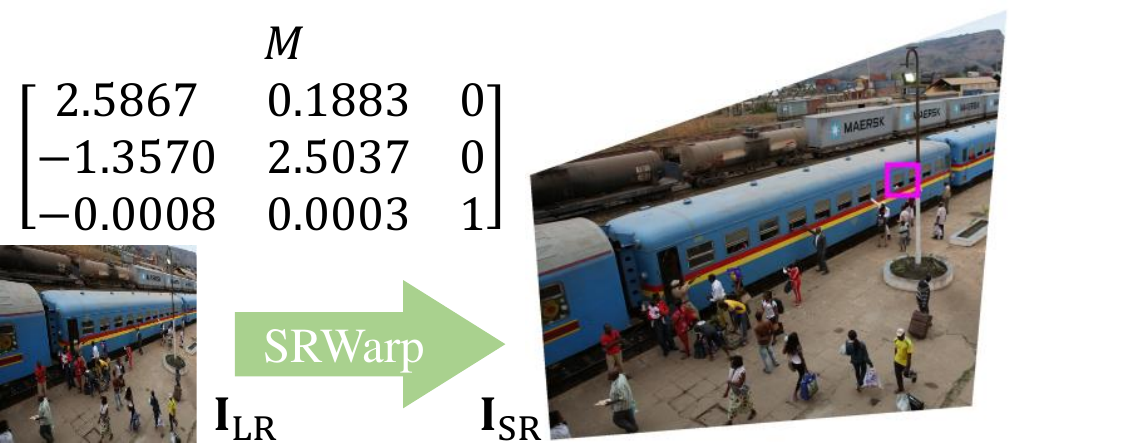}}
    \hfill
    \subfloat[\cv{cv2}]{\includegraphics[width=0.15\linewidth]{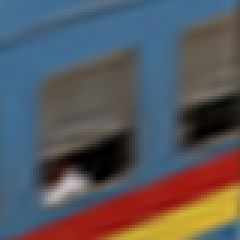}}
    \hfill
    \subfloat[RRDB $+$ \cv{cv2}]{\includegraphics[width=0.15\linewidth]{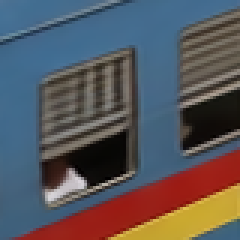}}
    \hfill
    \subfloat[\textbf{SRWarp}]{\includegraphics[width=0.15\linewidth]{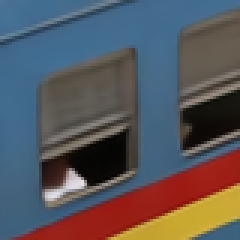}}
    \hfill
    \subfloat[GT]{\includegraphics[width=0.15\linewidth]{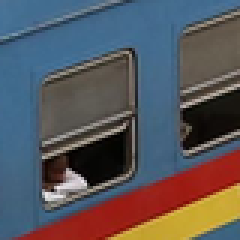}}
    \\
    \figcspace
    \caption{
        \textbf{Qualitative warping results on the DIV2KW$_\text{Test}$ dataset.}
        We provide input LR and output HR images with corresponding warping matrices $M_i$.
        Translation components are omitted for simplicity.
        Patches are cropped from the DIV2KW$_\text{Test}$ `\emph{0807.png}' and `\emph{0850.png}.'
        More visual comparisons are included in our supplementary material.
    }
    \label{fig:qualitative}
    \figspace
\end{figure*}

\Paragraph{Multiscale blending.}
\tabref{tab:baseline} shows that our multiscale approach (M) consistently improves the spatially-varying SR performance by a larger margin than the single-scale counterpart with the baseline EDSR~\cite{sr_edsr} backbone (A-R in \tabref{tab:baseline}).
We justify the design of our blending module in \tabref{tab:ablation_blending} by only changing the formulation to calculate the combination coefficients $w_{\times s}$ in \eqref{eq:blending}.
Interestingly, simply averaging (Average in \tabref{tab:ablation_blending}) the warped features $\mathbf{W}_{\times s}$ produces a better result than concatenating and blending them with a trainable $1 \times 1$ convolutional layer (Concat. in \tabref{tab:ablation_blending}).
Such performance decrease demonstrates that an appropriate design is required for the efficient blending module because concatenation is a more general formulation.
We also analyze how content and scale features support the blending module to combine multiscale representations effectively.
If the content information is ignored (w/o $\mathbf{C}$ in \tabref{tab:ablation_blending}), our SRWarp model suffers an mPSNR drop of 0.05dB.
Removing the scale features (w/o $\mathbf{S}$ in \tabref{tab:ablation_blending}) also brings a similar degree of performance degradation, justifying the design of the proposed multiscale blending.

\Paragraph{Reconstruction module and partial convolution.}
Since the reconstruction unit $\mathcal{R}$ can further refine output images, the module evidently brings additional performance gains for all combinations ($\ast$-R in \tabref{tab:baseline}).
We also examine the usefulness of the partial convolution~\cite{liu2018partialinpainting, liu2018partialpadding} in the content feature extractor and reconstruction module.
Compared to the previous SR methods, our SRWarp framework is more sensitive to boundary effects due to several reasons.
First, image boundaries, i.e., regions between valid and void areas, are not aligned with convolutional kernels and have irregular shapes.
Second, because we place irregular-shaped data on a regular 2D grid, numerous void pixels in the warped image negatively affect the following learnable layers.
SRWarp converges much slower without the partial convolution, and its final performance decreases by 0.06dB due to the severe boundary effects.

\Paragraph{Backbone architecture.}
The last two rows of \tabref{tab:baseline} show the effects of different backbone architectures on the performance of the SRWarp method.
Using the larger MRDB network with 17.1M parameters results in a significant PSNR gain of +0.27dB compared with the MDSR backbone with 1.7M parameters, indicating better fitting on the training data results in higher validation performance.

\begin{table*}[t]
    \centering
    \begin{tabularx}{\linewidth}{p{2.5cm} c r >{\centering\arraybackslash}X >{\centering\arraybackslash}X >{\centering\arraybackslash}X >{\centering\arraybackslash}X >{\centering\arraybackslash}X >{\centering\arraybackslash}X >{\centering\arraybackslash}X >{\centering\arraybackslash}X >{\centering\arraybackslash}X}
         \toprule
         & & & \multicolumn{9}{c}{PSNR$^\uparrow$(dB) on B100~\cite{data_bsd200} with arbitrary scale factors} \\
         Method & \# Params & Runtime & $\times 2.0$ & $\times 2.2$ & $\times 2.5$ & $\times 2.8$ & $\times 3.0$ & $\times 3.2$ & $\times 3.5$ & $\times 3.8$ & $\times 4.0$ \\
         \midrule
         SRCNN~\cite{sr_srcnn} & 0.06M & 2340ms & 27.11 & 27.85 & 28.62 & 28.71 & 28.37 & 27.89 & 27.17 & 26.59 & 26.27 \\
         VDSR~\cite{sr_vdsr} & 0.67M & 26ms & 31.82 & 30.36 & 29.54 & 28.84 & 28.77 & 28.15 & 27.82 & 27.46 & 27.27 \\
         Meta-EDSR~\cite{sr_meta} & 40.1M & 218ms & 32.26 & 31.31 & \secondbest{30.40} & 29.61 & 29.22 & 28.82 & 28.27 & \secondbest{27.86} & 27.67 \\
         Meta-RDN~\cite{sr_meta} & 22.4M & 253ms & \best{32.33} & \secondbest{31.45} & \best{30.46} & \secondbest{29.69} & \secondbest{29.26} & \secondbest{28.88} & \secondbest{28.41} & \best{28.01} & \secondbest{27.71} \\
         \midrule
         \textbf{SRWarp~(MRDB)} & 18.3M & 155ms & \secondbest{32.31} & \best{31.46} & \best{30.46} & \best{29.71} & \best{29.27} & \best{28.89} & \best{28.42} & \best{28.01} & \best{27.77} \\
         \bottomrule
    \end{tabularx}
    \\
    \tabcspace
    \caption{
        \textbf{Quantitative comparison of the arbitrary-scale SR task.}
        We use official implementations of each method and compare them in a unified environment.
        The average runtime is measured on the $\times 3.0$ SR task using 100 test images excluding initialization, I/O, and the other overheads.
        For the SRCNN~\cite{sr_srcnn} model, the $\times 3$ network (9-5-5) is evaluated across all scaling factors~\cite{sr_vdsr} on CPU.
        Our SRWarp consistently outperforms the other approaches even with fewer parameters.
    }
    \label{tab:fractional}
    \tabspace
\end{table*}

\subsection{Comparison with the Other Methods}
We compare the proposed SRWarp with existing methods.
We note that providing an exact comparison with other methods is difficult given that our approach is the \emph{first} attempt toward generalized image SR.
First, we adopt a conventional interpolation-based warping algorithm from the OpenCV~\cite{opencv_library}.
We use {\fontfamily{qcr}\selectfont cv2.WarpPerspective} function with a bicubic kernel to synthesize warped images.
For alternatives, we combine state-of-the-art SR models and the traditional warping operation.
Since the given LR images are supersampled before interpolation, the warping function can synthesize high-quality results directly.
We note that the transformation matrix $M$ is compensated to $M M_{s^{-1} s^{-1}}$ for $\times s$ SR model because the outputs from SR models are $\times s$ larger than the original input.

\tabref{tab:comp} provides quantitative comparison of various methods.
For fairness, we adopt the DIV2KW$_\text{Test}$ dataset rather than the validation split used in \secref{ssec:ablation}.
Compared with the traditional \cv{cv2} algorithm, using the SR methods provides significant improvements with the mPSNR gain of at least +2.41dB.
Higher-scale $\times 4$ models tend to perform better than their $\times 2$ counterparts, justifying the importance of the fine-grained supersampling.
Our SRWarp model further outperforms the other SR-based formulations by a large margin.
\figref{fig:qualitative} shows that our approach reconstructs much sharper images with less aliasing, demonstrating the effectiveness of AWL and multiscale blending.

\subsection{Arbitrary-scale SR}
Our SRWarp provides a generalization of the conventional SR models defined on scaling matrices in \eqref{eq:scale_matrix} only.
To justify that the proposed framework is compatible with existing formulations, we evaluate our method on the regular SR tasks of arbitrary scaling factors.
We train our SRWarp model with RRDB~\cite{sr_esrgan} backbone on fractional-scale DIV2K dataset~\cite{sr_meta} and evaluate it following Hu~\etal~\cite{sr_meta}.
The input transformation of the SRWarp is constrained to \eqref{eq:scale_matrix}, and the other configurations are fixed.
\tabref{tab:fractional} shows an average PSNR of the luminance (Y) channel between SR results and ground-truth images on B100~\cite{data_bsd200} dataset.
We note that SRCNN~\cite{sr_srcnn} and VDSR~\cite{sr_vdsr} first resize the input image to an arbitrary target resolution before take it into the network.
Compared with the meta-upscale module (Meta-EDSR and Meta-RDN), our adaptive warping layer and multiscale blending provide an efficient and generalized model for the arbitrary-scale SR task.

\begin{figure}[t]
    \centering
    \renewcommand{\wp}{0.28}
    \vspace{-3mm}
    \makecell{Sine}
    \subfloat{\includegraphics[align=c, width=\wp \linewidth]{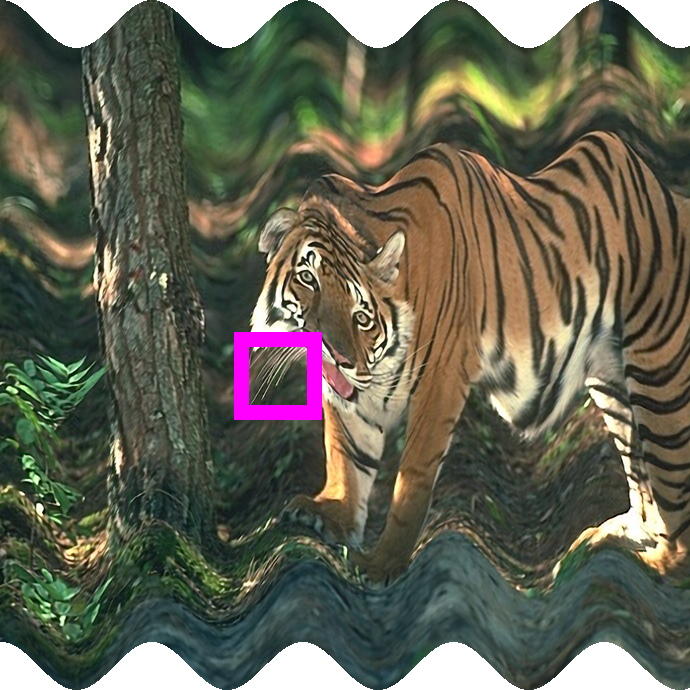}}
    \hspace{1mm}
    \subfloat{\includegraphics[align=c, width=\wp \linewidth]{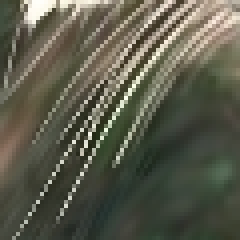}}
    \hspace{1mm}
    \subfloat{\includegraphics[align=c, width=\wp \linewidth]{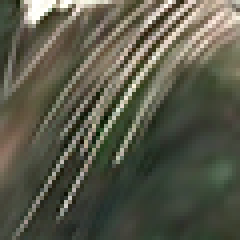}}
    \\
    \vspace{-3.5mm}
    \addtocounter{subfigure}{-3}
    \makecell{Barrel}
    \subfloat[$\img{SR}$]{\includegraphics[align=c, width=\wp \linewidth]{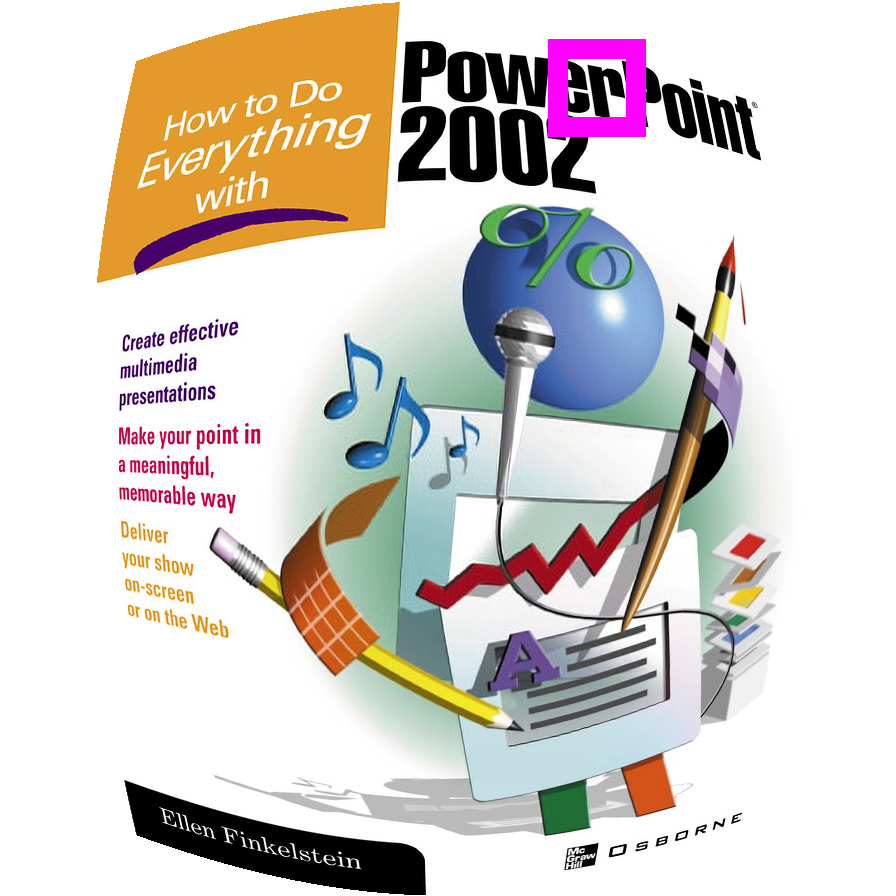}}
    \hspace{1mm}
    \subfloat[RRDB~\cite{sr_esrgan}]{\includegraphics[align=c, width=\wp \linewidth]{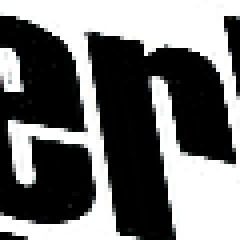}}
    \hspace{1mm}
    \subfloat[\textbf{SRWarp}]{\includegraphics[align=c, width=\wp \linewidth]{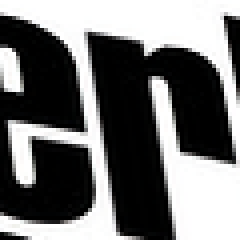}}
    \\
    \figcspace
    \caption{
        \textbf{General image warping with our SRWarp.}
        %
        We apply various functional transforms to samples from B100~\cite{data_bsd200} `\emph{108005.png}' and Set14~\cite{data_set14} '\emph{ppt3.png}.'
        (b) RRDB corresponds to RRDB + \cv{cv2} in \tabref{tab:comp}.
    }
    \label{fig:functional}
    \figspace
\end{figure}

\subsection{Over the Homographic Transformation}
Our SRWarp model is trained on homographic transformation only.
However, we can extend the method to an arbitrary backward mapping equation in a functional form $f^{-1}_{M} \paren{ x', y' } = \paren{ x, y }$ without \emph{any} modification.
Although we adopt homographic transformations only for the training, the adaptive warping layer and multiscale blending help the method be generalized well on unseen deformations.
\figref{fig:functional} compares our results on various functional transforms against the combination of RRDB~\cite{sr_esrgan} and traditional bicubic interpolation.
Our SRWarp model can provide more flexibility and diversity in general image editing tasks by reconstructing visually pleasing edge structures.
\section{Conclusion}
We propose a generalization of the conventional SR tasks under image transformation for the first time.
Our SRWarp framework deals with the spatially-varying upsampling task when arbitrary resolutions and shapes are required to the output image.
We also provide extensive ablation studies on the proposed method to validate the contributions of several novel components, e.g., adaptive warping layer and multiscale blending, in our design.
The visual comparison demonstrates why the SRWarp model is required for image warping, justifying the advantage of the proposed method.

\section*{Acknowledgement}
This work was supported by IITP grant funded by the Ministry of Science and ICT of Korea (No. 2017-0-01780) and AIRS Company in Hyundai Motor Company \& Kia Motors Corporation through HKMC-SNU AI Consortium Fund.

\newpage
\newtoggle{isarxiv}
\toggletrue{isarxiv}

\renewcommand{\thetable}{S\arabic{table}}
\renewcommand{\thefigure}{S\arabic{figure}}
\renewcommand{\theequation}{S\arabic{equation}}
\renewcommand{\thesection}{S\arabic{section}}
\setcounter{table}{0}
\setcounter{figure}{0}
\setcounter{equation}{0}
\setcounter{section}{0}

\section{Details about the DIV2KW Dataset}
We synthesize the proposed DIV2KW dataset with various random projective transformations.
Warping parameters are determined by combining sheering, rotation, scaling, and projection matrices, denoted as $H$, $R$, $S$, and $P$, respectively.
We construct $M_{i}^{-1}$ first and inverse the matrix to implement the actual transform $M_i$ to construct feasible transformations.
Detailed specifications of the transformations are described as follows:
\begin{equation}
    \begin{split}
        M_i^{-1} &= H R S P, \\
        H &= \begin{pmatrix}
            1 & h_x & 0 \\
            h_y & 1 & 0 \\
            0 & 0 & 1 \\
        \end{pmatrix}, \\
        R &= \begin{pmatrix}
            \cos{\theta} & \sin{\theta} & 0 \\
            -\sin{\theta} & \cos{\theta} & 0 \\
            0 & 0 & 1 \\
        \end{pmatrix}, \\
        S &= \begin{pmatrix}
            s_x & 0 & 0 \\
            0 & s_y & 0 \\
            0 & 0 & 1 \\
        \end{pmatrix}, \\
        P &= \begin{pmatrix}
            1 & 0 & t_x \\
            0 & 1 & t_y \\
            p_x & p_y & 1 \\
        \end{pmatrix},
    \end{split}
    \label{eq:transform}
\end{equation}
where the variables are randomly sampled from uniform $\paren{ \mathcal{U} }$ or normal $\paren{ \mathcal{N} }$ distributions.
\tabref{tab:transform_random} shows parameters of the random distributions we use.
We note that the projection matrix $P$ varies depending on the size of the HR sample $\img{HR}$ to normalize image shapes after warping.

\iftoggle{isarxiv}{}{
\begin{table}[t!]
    \centering
    \begin{tabularx}{\linewidth}{>{\centering\arraybackslash}X l l}
        \toprule
        Variable(s) & Sampling distribution & Note \\
        \midrule
        $h_x$, $h_y$ & $\mathcal{U} \paren{ -0.25, 0.25 }$ & Random sheering \\
        $\theta$ & $\mathcal{N} \paren{ 0, {15^{\circ}}^2 }$ & Random rotation \\
        $s_x$, $s_y$ & $\mathcal{U} \paren{0.35, 0.5}$ & Random scaling \\
        $t_x$ & $\mathcal{U} \paren{ -0.75w, 0.125w }$ & \multirow{4}{*}{Random projection} \\
        $t_x$ & $\mathcal{U} \paren{ -0.75h, 0.125h }$ & \\
        $p_x$ & $\mathcal{U} \paren{ -0.6w, 0.6w }$ & \\
        $p_y$ & $\mathcal{U} \paren{ -0.6h, 0.6h }$ & \\
        \bottomrule
    \end{tabularx}
    \\
    \tabcspace
    \caption{
        \textbf{Random variable specifications for the transformation matrix.}
        $h \times w$ denotes the resolution of an HR patch or image.
    }
    \label{tab:transform_random}
\end{table}
\begin{figure}[t!]
    \centering
    \subfloat[Training patches and transforms]{\includegraphics[width=\linewidth]{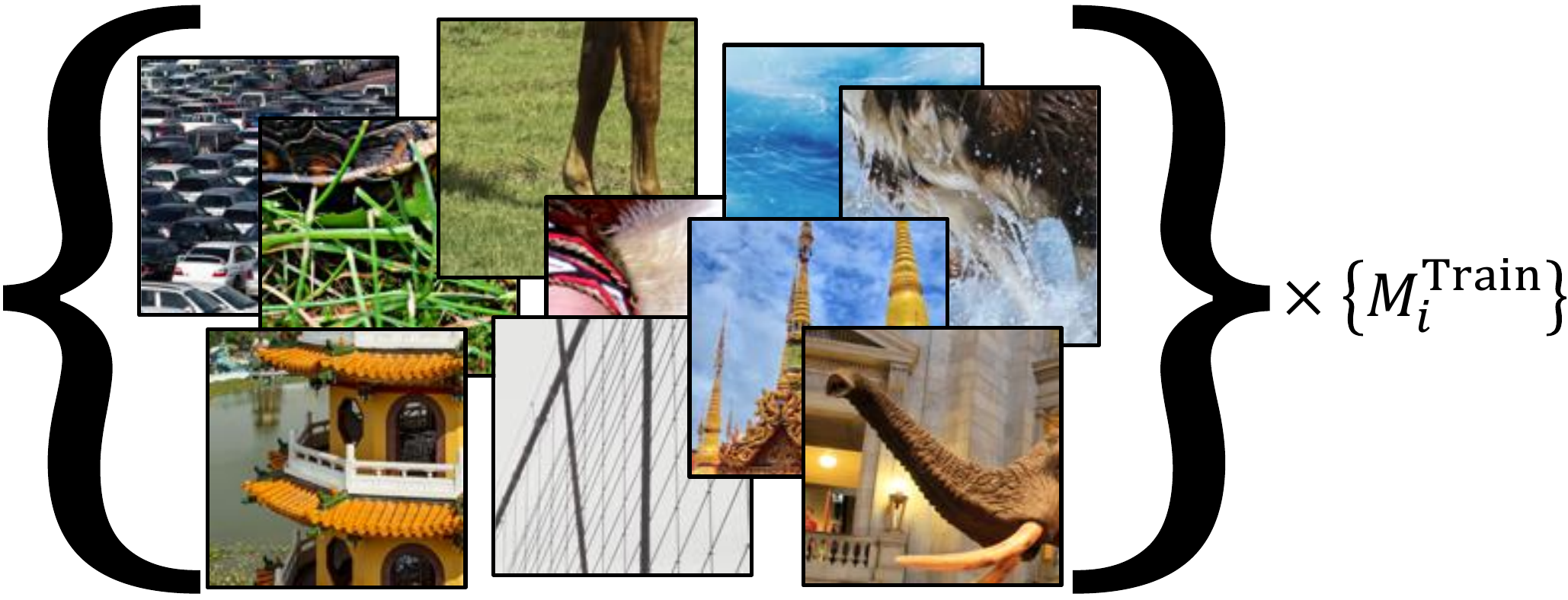}} \\
    \vspace{-2mm}
    \subfloat[Validation patches and transforms]{\includegraphics[width=\linewidth]{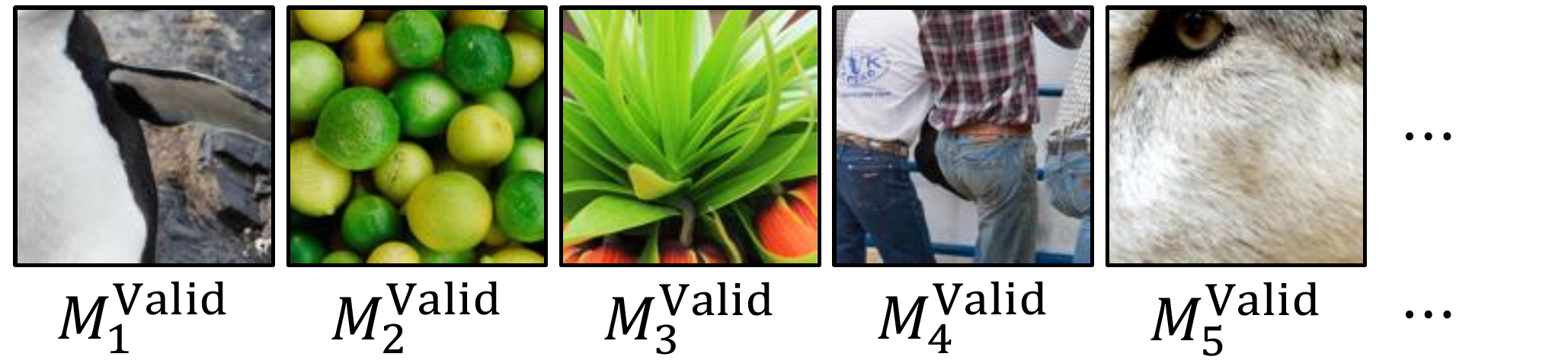}} \\
    \vspace{-2mm}
    \subfloat[Test images and transforms]{\includegraphics[width=\linewidth]{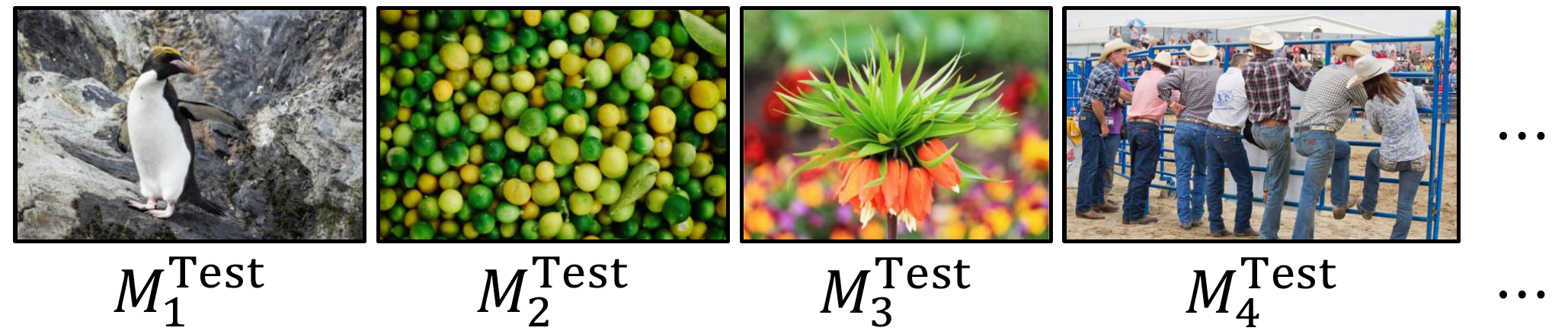}} \\
    \figcspace
    \caption{
        \textbf{Illustration of the DIV2KW data splits.}
        We use the same images for validation and test, but their sizes and corresponding matrices are different.
    }
    \label{fig:data_split}
    \figspace
\end{figure}
}

For training, we randomly crop $N$ many $384 \times 384$ patches $\img{HR}$ from 800 images in the DIV2K~\cite{data_div2k} dataset and apply arbitrary $M_i^{-1}$ to get corresponding LR samples $\img{LR}$.
We leverage a widely-used bicubic interpolation for the synthesis.
A cropped version of the LR image $\img{LR\text{-crop}}$ has a maximum size of $96 \times 96$, which is randomly acquired by ignoring \emph{void} pixels.
For evaluation, we follow the similar pipeline but center-crop one $384 \times 384$ patch from each DIV2K validation image, i.e., from `\emph{0801.png}' to `\emph{0900.png}.'
100 transformation matrices $M_i$ is assigned to each image, and the cropped region $\img{LR\text{-crop}}$ is also fixed to ensure reproducibility.

To fairly compare our SRWarp model against the conventional warping method, we use full-size images $\img{HR}$ as test inputs.
The smallest HR image in the DIV2K validation dataset has a resolution of $816 \times 2040$, and the largest one is $2040 \times 2040$.
Since the transformation matrix $M_i$ depends on the image resolution, each of the 100 test examples has assigned fixed warping parameters similar to the validation case.
We also note that the test transformations are totally different from training and evaluation due to the size difference.
\figref{fig:data_split} demonstrates a visual comparison between the training, validation, and test splits.

\iftoggle{isarxiv}{
\begin{table}[t!]
    \centering
    \begin{tabularx}{\linewidth}{>{\centering\arraybackslash}X l l}
        \toprule
        Variable(s) & Sampling distribution & Note \\
        \midrule
        $h_x$, $h_y$ & $\mathcal{U} \paren{ -0.25, 0.25 }$ & Random sheering \\
        $\theta$ & $\mathcal{N} \paren{ 0, {15^{\circ}}^2 }$ & Random rotation \\
        $s_x$, $s_y$ & $\mathcal{U} \paren{0.35, 0.5}$ & Random scaling \\
        $t_x$ & $\mathcal{U} \paren{ -0.75w, 0.125w }$ & \multirow{4}{*}{Random projection} \\
        $t_x$ & $\mathcal{U} \paren{ -0.75h, 0.125h }$ & \\
        $p_x$ & $\mathcal{U} \paren{ -0.6w, 0.6w }$ & \\
        $p_y$ & $\mathcal{U} \paren{ -0.6h, 0.6h }$ & \\
        \bottomrule
    \end{tabularx}
    \\
    \tabcspace
    \caption{
        \textbf{Random variable specifications for the transformation matrix.}
        $h \times w$ denotes the resolution of an HR patch or image.
    }
    \label{tab:transform_random}
\end{table}
\begin{figure}[t!]
    \centering
    \subfloat[Training patches and transforms]{\includegraphics[width=\linewidth]{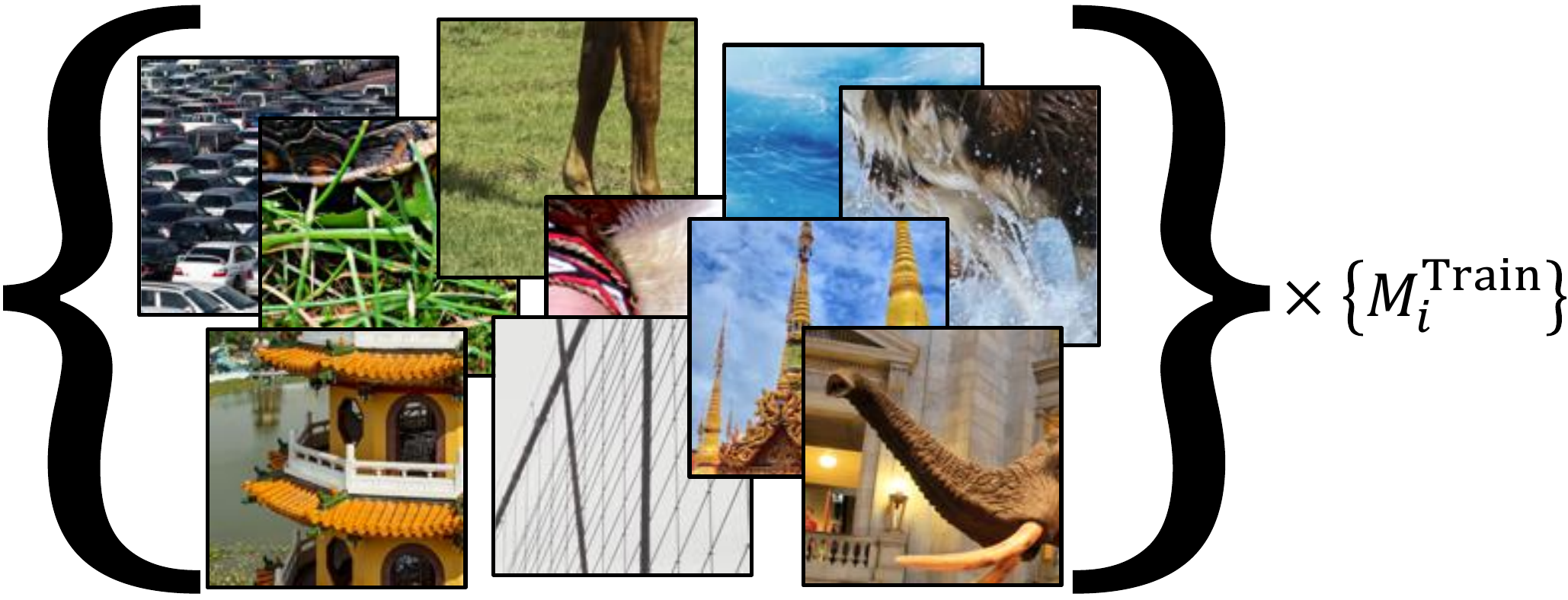}} \\
    \vspace{-2mm}
    \subfloat[Validation patches and transforms]{\includegraphics[width=\linewidth]{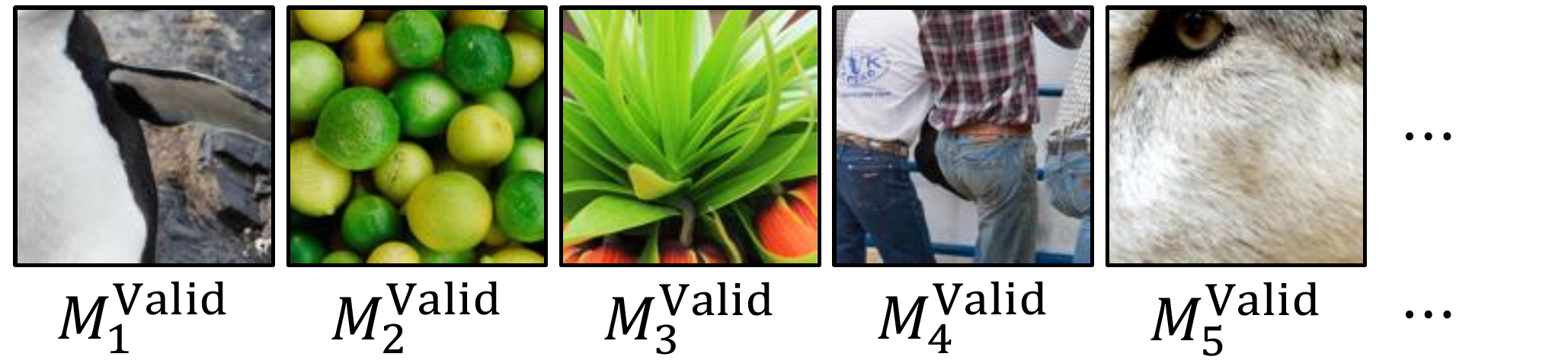}} \\
    \vspace{-2mm}
    \subfloat[Test images and transforms]{\includegraphics[width=\linewidth]{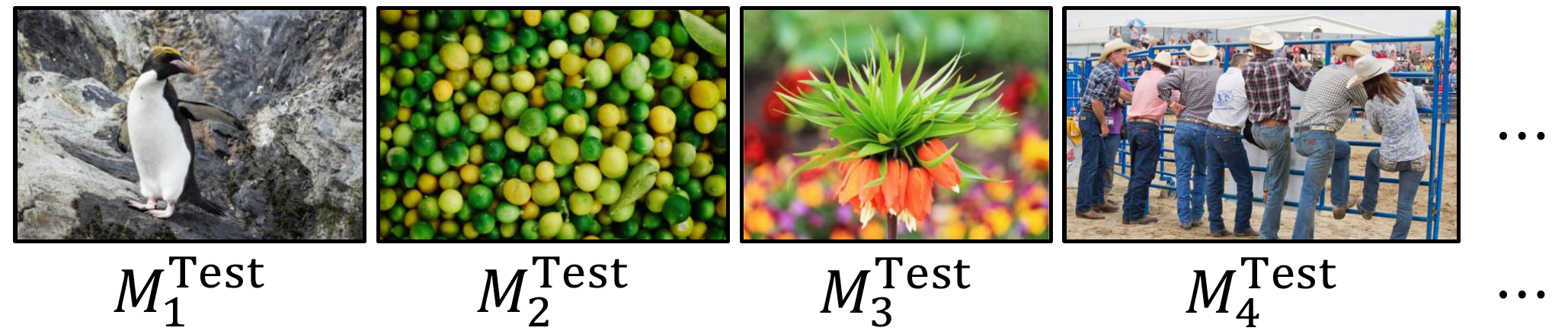}} \\
    \figcspace
    \caption{
        \textbf{Illustration of the DIV2KW data splits.}
        We use the same images for validation and test, but their sizes and corresponding matrices are different.
    }
    \label{fig:data_split}
    \figspace
\end{figure}
}{}

\section{Details about the AWL}
\begin{figure}[t]
    \centering
    \definecolor{mygreen}{RGB}{0, 176, 80}
    \definecolor{mynavy}{RGB}{0, 32, 96}
    \includegraphics[width=\linewidth]{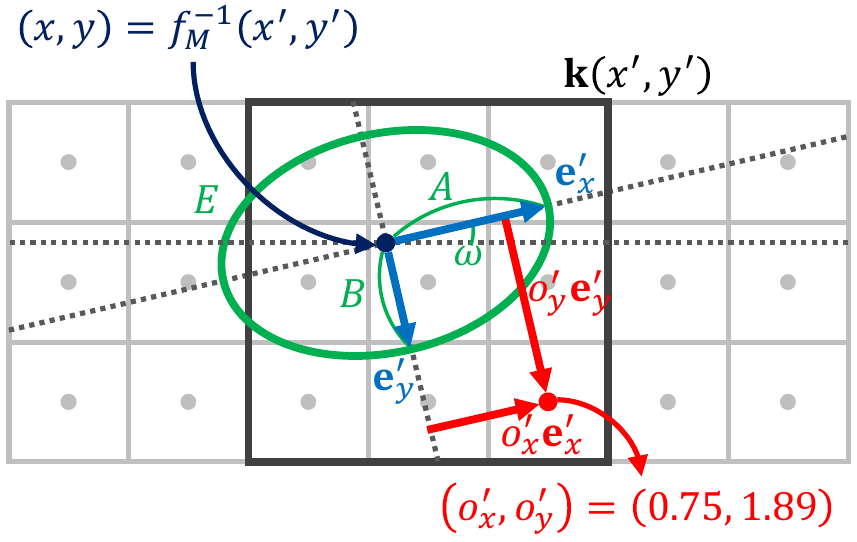}
    \\
    \figcspace
    \caption{
        \textbf{Detailed illustration of the adaptive resampling coordinate.}
        We find the \textcolor{mygreen}{\textbf{green}} ellipse $E$ for each position $\paren{ x', y' }$ on the target coordinate.
        Best viewed with Figure~\fakeref{3} in our main manuscript.
        $\paren{o'_x, o'_y}$ denotes a relative offset vector, i.e., $\mathbf{o}'_{11}$, of the example point in \textcolor{red}{\textbf{red}} with respect to the reference $\paren{ x, y }$ in \textcolor{mynavy}{\textbf{navy}} on the adaptive grid.
    }
    \label{fig:kernel_supple}
    \figspace
\end{figure}

We concretely describe how the adaptive sampling coordinate in Section~\fakeref{3.2} of our main manuscript is calculated.
A unit circle $x'^2 + y'^2 = 1$ on the target domain is mapped to a general ellipse $E$ on the source coordinate by the affine transform $J = \begin{pmatrix} \mathbf{u}\transpose & \mathbf{v}\transpose \end{pmatrix}$.
For simplicity, we will formulate the shape $E$ as a rotated version of an axis-aligned ellipse, which can be described as follows:
\begin{equation}
    \frac{ \paren{ x\cos{\omega} + y\sin{\omega} }^2 }{A^2} + \frac{ \paren{ y\cos{\omega} - x\sin{\omega} }^2 }{B^2} = 1,
    \label{eq:ellipse}
\end{equation}
where $A$ and $B$ are lengths of principal axes, and $\omega$ denotes the rotated angle, respectively, as shown in \figref{fig:kernel_supple}.
By assuming that $\mathbf{u} = \paren{u_x, u_y}$ and $\mathbf{v} = \paren{v_x, v_y}$, we get the following equation:
\begin{equation}
    \begin{pmatrix}
        x' \\ y'
    \end{pmatrix}
    = \frac{1}{D}
    \begin{pmatrix}
        v_y & -v_x \\
        -u_y & u_x
    \end{pmatrix}
    \begin{pmatrix}
        x \\ y
    \end{pmatrix},
    \label{eq:inverse}
\end{equation}
where $D = \det{J} = u_x v_y - v_x u_y$.
By substituting $x'$ and $y'$ in the equation of the unit circle with \eqref{eq:inverse}, we get the following:
\begin{equation}
    \paren{ v_y x - v_x y }^2 + \paren{ u_x y - u_y x }^2 = D^2.
    \label{eq:backward}
\end{equation}
Since \eqref{eq:ellipse} and \eqref{eq:backward} are \emph{equivalent}, the following three identites can be derived by developing the equations:
\begin{equation}
    \begin{split}
        \frac{\cos^2{\omega}}{A^2} + \frac{\sin^2{\omega}}{B^2} &= \frac{ u_y^2 + v_y^2 }{D^2}, \\
        \frac{\sin^2{\omega}}{A^2} + \frac{\cos^2{\omega}}{B^2} &= \frac{ u_x^2 + v_x^2 }{D^2}, \\
        \frac{2\cos{\omega}\sin{\omega}}{A^2} - \frac{2\cos{\omega}\sin{\omega}}{B^2} &= -\frac{2 u_x u_y + 2 v_x v_y}{D^2},
    \end{split}
    \label{eq:three_id}
\end{equation}
where right-hand-side terms are constant.
Using trigonometric identities, we can reduce \eqref{eq:three_id} as follows:
\begin{equation}
    \begin{split}
        \frac{1}{A^2} + \frac{1}{B^2} &= \frac{ \normfrob{J}^2 }{D^2}, \\
        \cos{2 \omega} \paren{ \frac{1}{A^2} - \frac{1}{B^2} } &= \frac{u_y^2 + v_y^2 - u_x^2 - v_x^2}{D^2}, \\
        \sin{2 \omega} \paren{ \frac{1}{A^2} - \frac{1}{B^2} } &= -\frac{2 u_x u_y + 2 v_x v_y}{D^2},
    \end{split}
    \label{eq:organized}
\end{equation}
where $\normfrob{J}^2 = u_x^2 + v_x^2 + u_y^2 + v_y^2$.
Using \eqref{eq:organized}, it is possible to represent three unknowns $A$, $B$, and $\omega$ with respect to $J$ in a straighforward fashion as follows:
\begin{equation}
    \begin{split}
        \omega &= \frac{1}{2} \arctan{ \frac{2 u_x u_y + 2 v_x v_y}{u_x^2 + v_x^2 - u_y^2 - v_y^2} }, \\
        A &= \sqrt{ \frac{ D^2 \cos{2\omega} }{ \cos^2{\omega} \normfrob{J}^2 - u_x^2 - v_x^2 } }, \\
        B &= \sqrt{ \frac{ D^2 \cos{2\omega} }{ \cos^2{\omega} \normfrob{J}^2 - u_y^2 - v_y^2 } }.
    \end{split}
\end{equation}
Then, the basis vectors $\mathbf{e}'_x$ and $\mathbf{e}'_y$ of the adaptive grid can be calculated as follows:
\begin{equation}
    \begin{split}
        \mathbf{e}'_x &= \paren{ A \cos{\omega}, A\sin{\omega} }, \\
        \mathbf{e}'_y &= \paren{ -B \sin{\omega}, B\cos{\omega} }.
    \end{split}
\end{equation}
\figref{fig:kernel_supple} shows a visualization of the ellipse $E$ with its principal axes on the source domain.

\begin{figure*}[t!]
    \centering
    \renewcommand{\wp}{0.333\linewidth}
    \subfloat[Correspondences between points in $\img{LR}$ and $\img{SR}$]{\includegraphics[width=\linewidth]{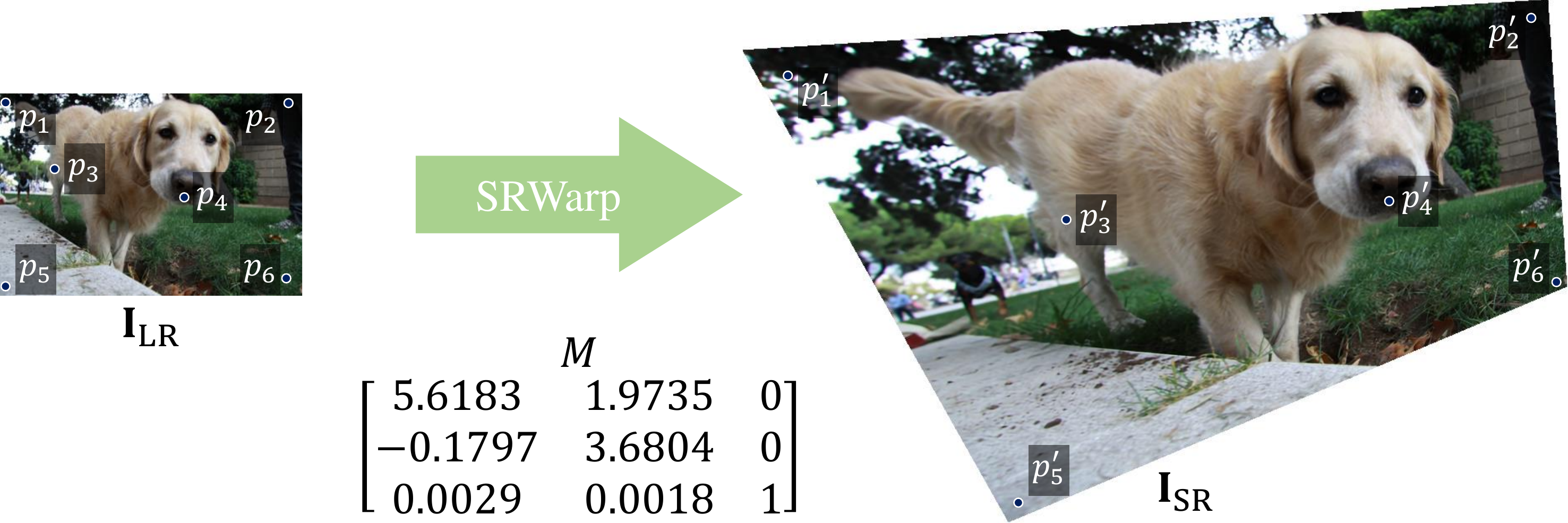}}
    \\
    \subfloat[Resampling coordinate at $p_1$]{\includegraphics[width=\wp]{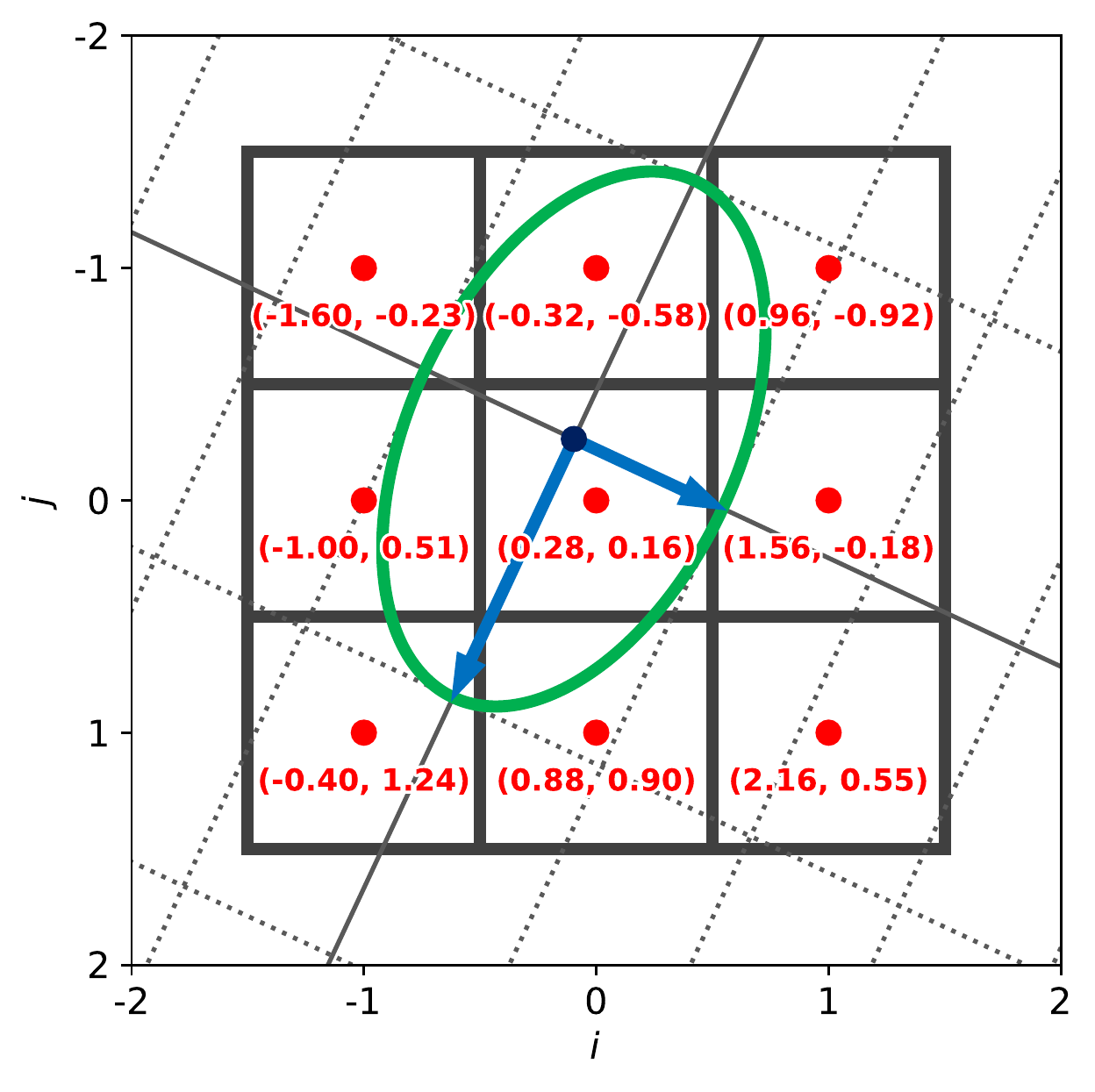}}
    \hfill
    \subfloat[Resampling coordinate at $p_2$]{\includegraphics[width=\wp]{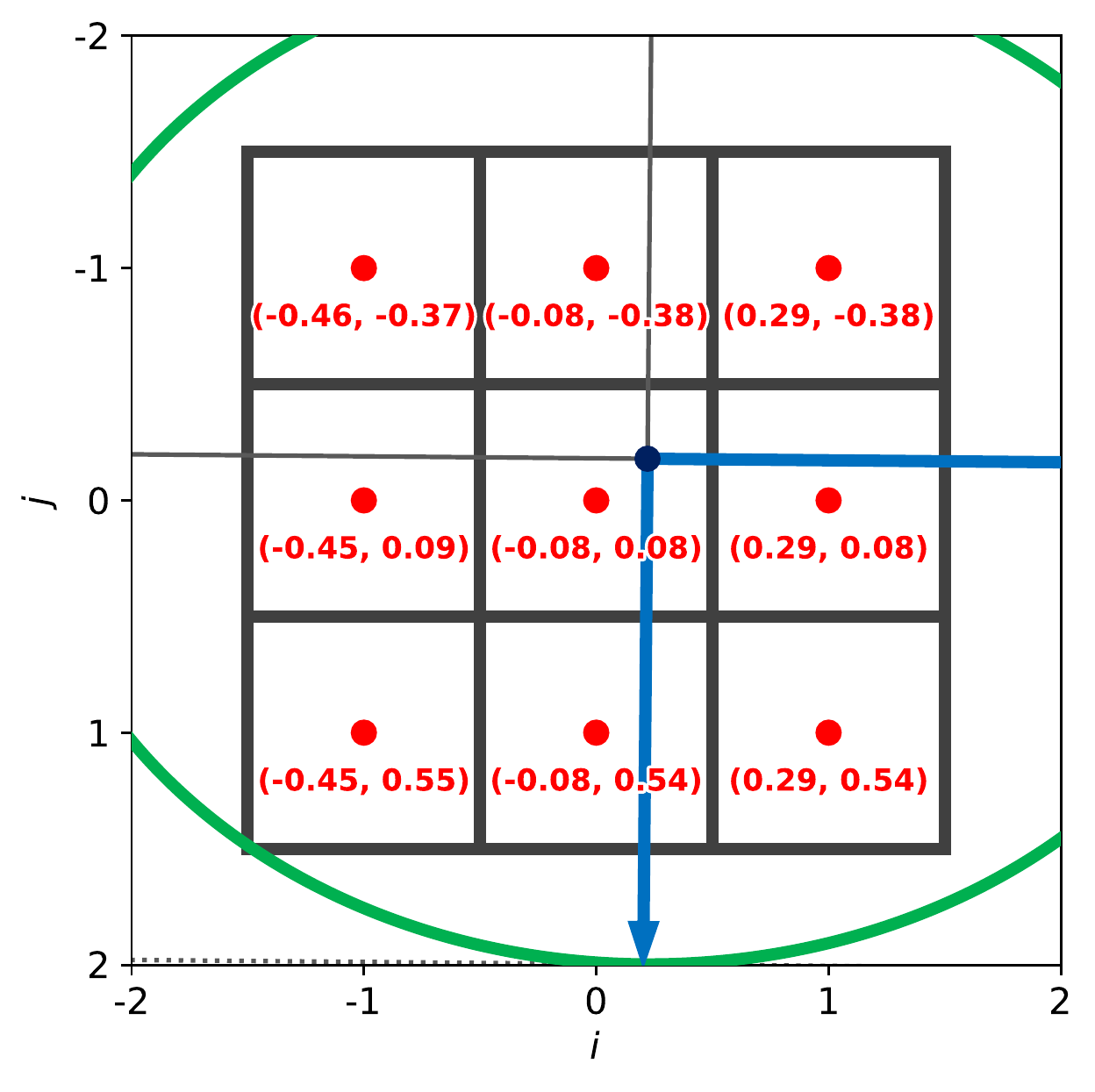}}
    \hfill
    \subfloat[Resampling coordinate at $p_3$]{\includegraphics[width=\wp]{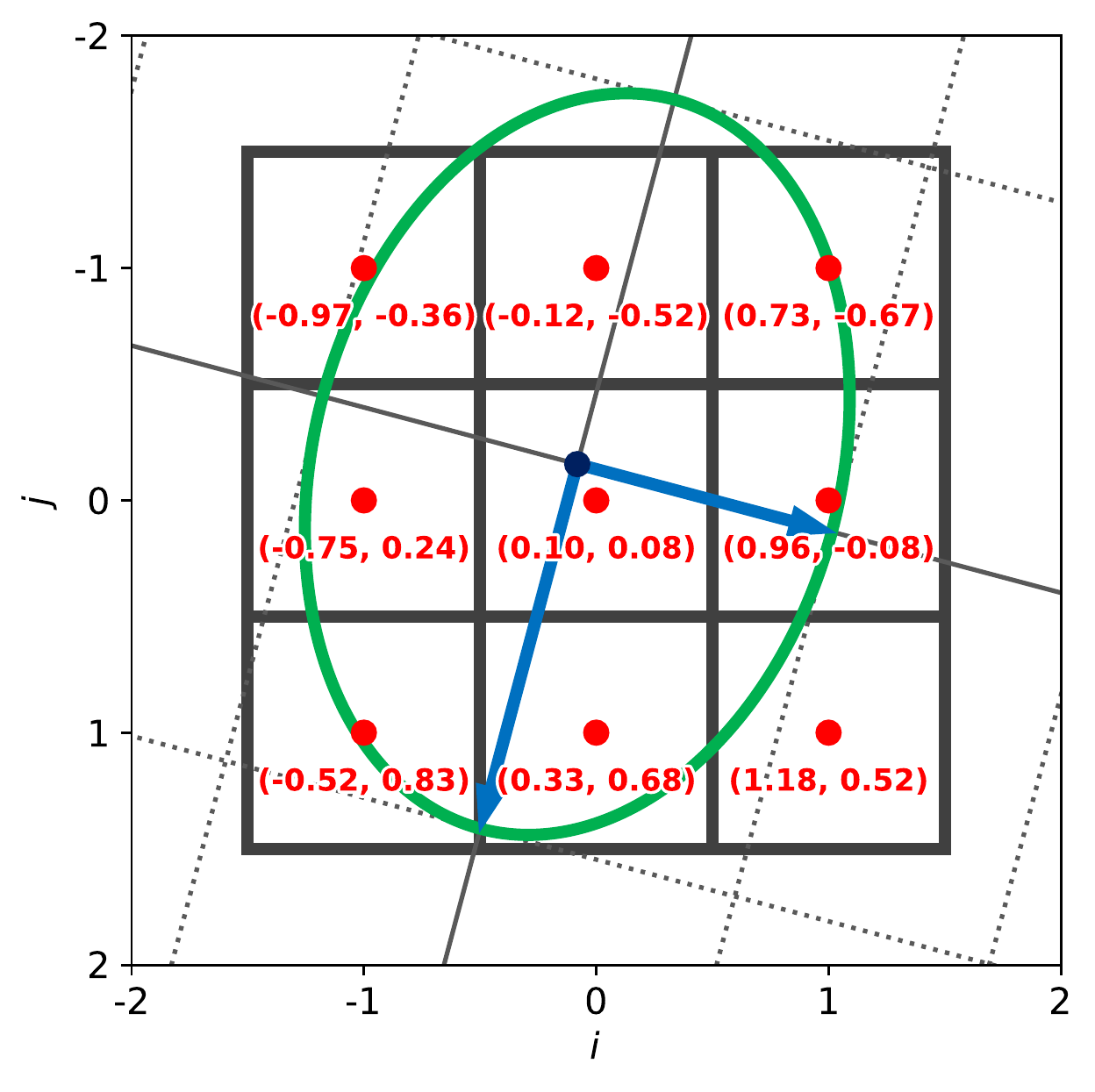}}
    \\
    \vspace{-3mm}
    \subfloat[Resampling coordinate at $p_4$]{\includegraphics[width=\wp]{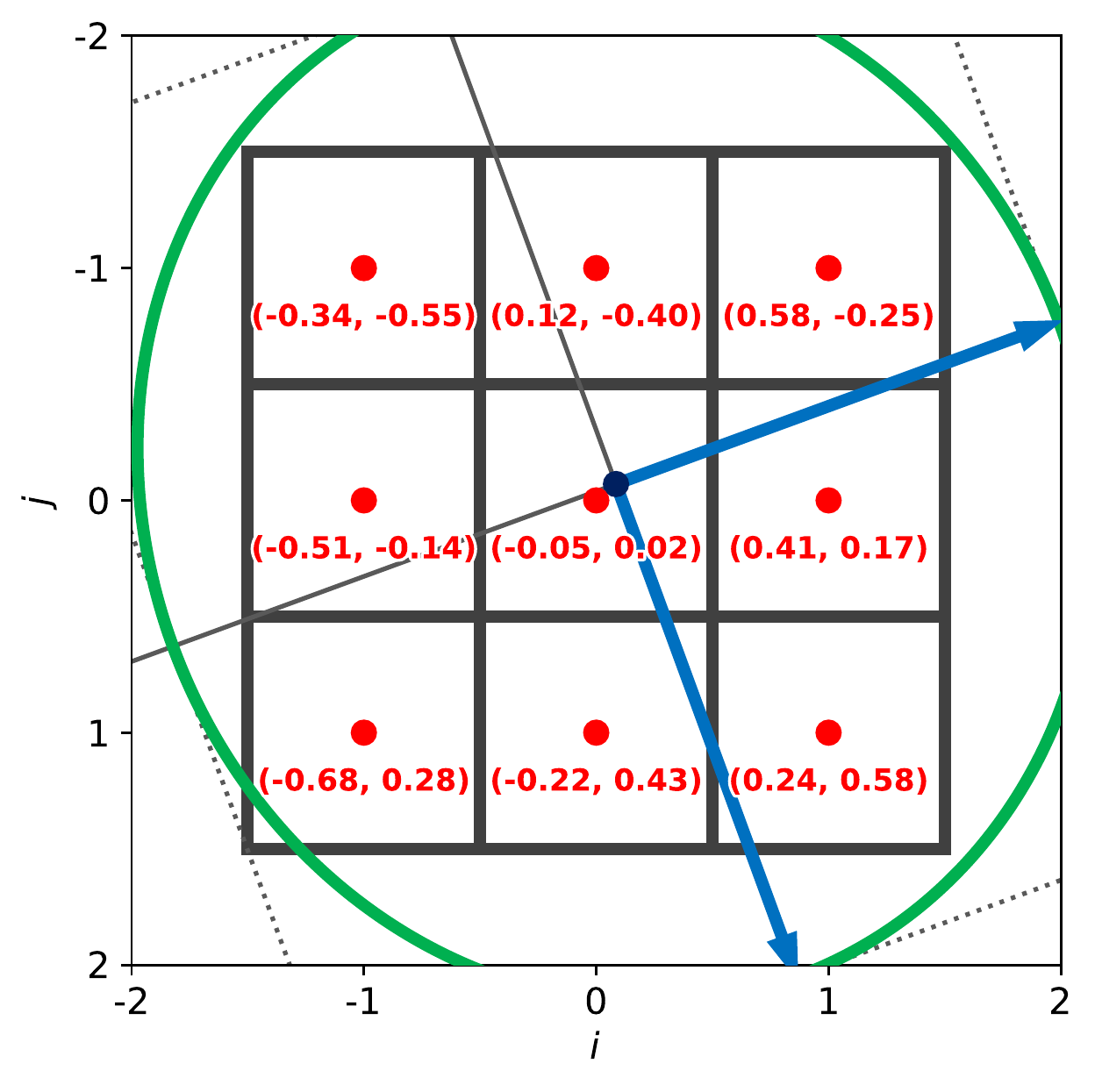}}
    \hfill
    \subfloat[Resampling coordinate at $p_5$]{\includegraphics[width=\wp]{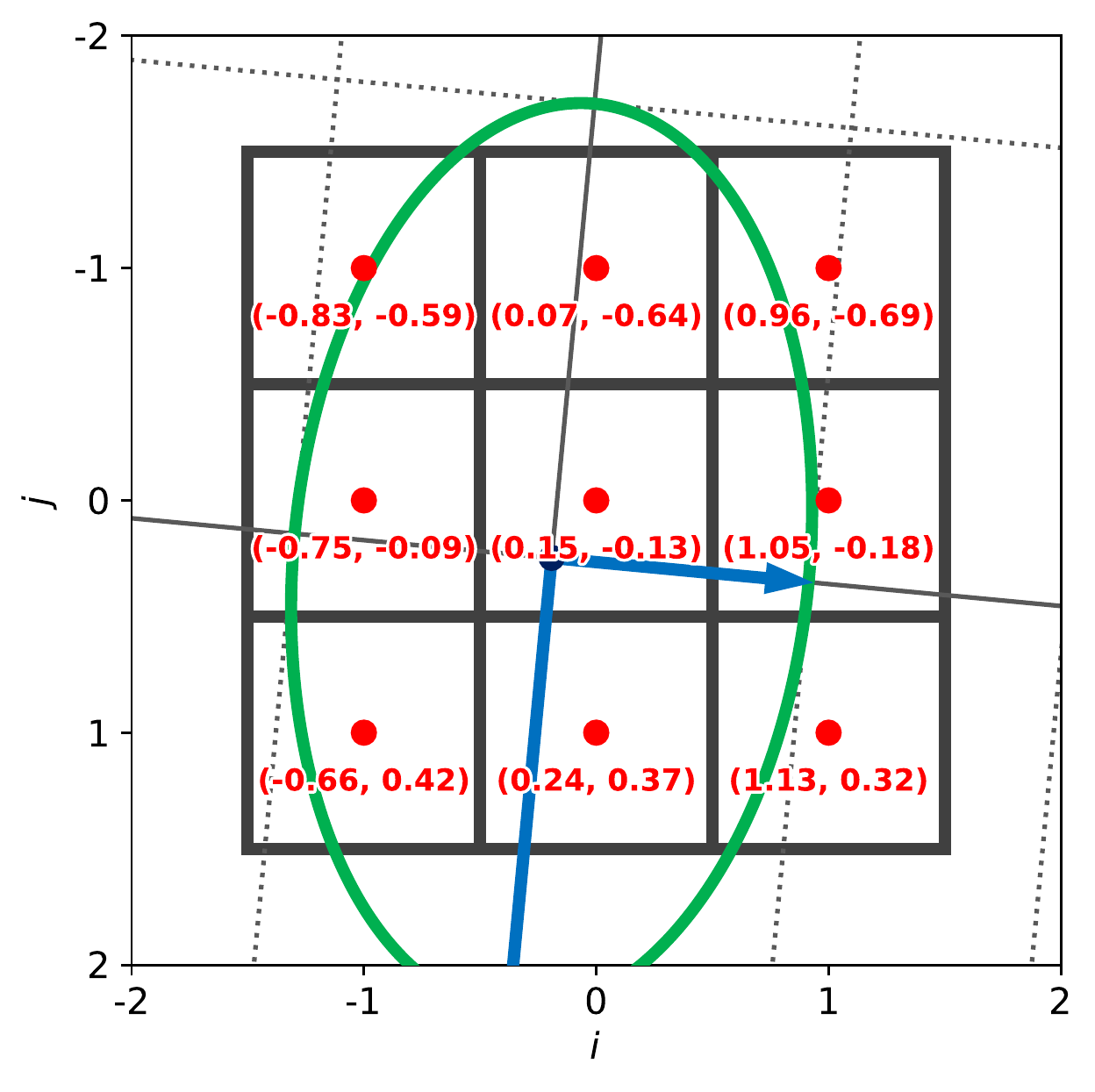}}
    \hfill
    \subfloat[Resampling coordinate at $p_6$]{\includegraphics[width=\wp]{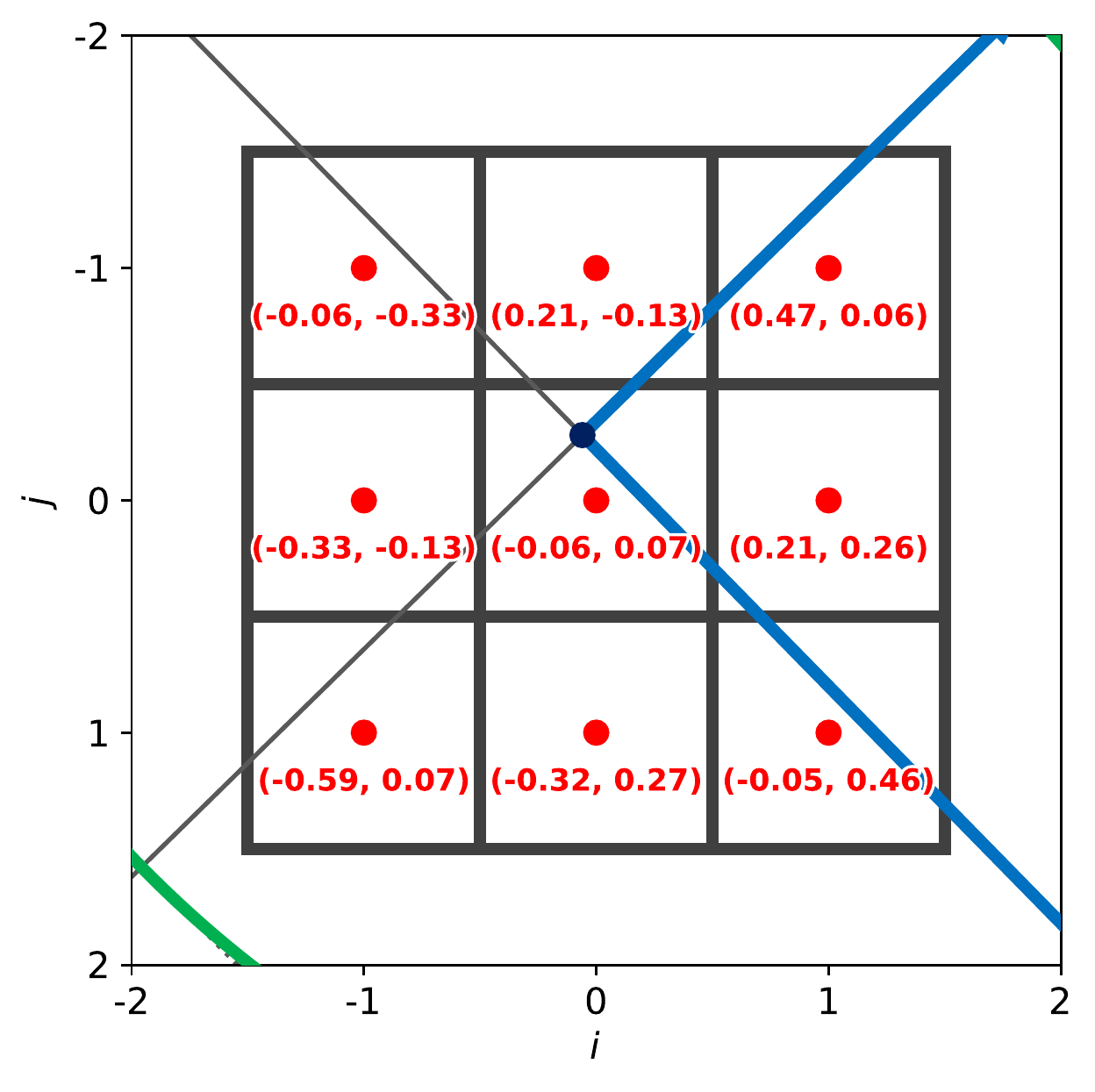}}
    \\
    \figcspace
    \caption{
        \textbf{Visualization of the spatially-varying adaptive grid.}
        (a) We note that the correspondence is calculated as $Mp_i = p'_i$.
        (b$-$g) We display locally-varying sampling grids for the proposed AWL.
        The $\times 4$ feature $\mathbf{F}_{\times 4}$ is used as a source domain for visualization.
        A bold $3 \times 3$ grid shows a resampling kernel $\mathbf{k} \paren{x', y'}$ on the source domain.
        We visualize the relative offsets in \textcolor{red}{\textbf{red}} with respect to the adaptive coordinate.
        Our kernel estimator $\mathcal{K}$ takes the $3 \times 3$ offset vectors and predicts appropriate resampling weights.
        Then, the output value $\mathbf{W}_{\times 4} \paren{ x', y' }$ is calculated by combining $3 \times 3$ points in \textcolor{red}{\textbf{red}} on the source domain with the corresponding kernel.
        The sampling grid tends to be shrunk if the feature is locally enlarged and vice versa.
    }
    \label{fig:sv_kernels}
    \figspace
\end{figure*}

A bold window in \figref{fig:kernel_supple} visualizes a $k \times k$ kernel support where the resampling weights $\mathbf{k} \paren{ x', y' }$ is evaluated on.
As described in (\fakeref{2}) of our main manuscript, we first locate $k \times k$ points in the window, i.e., $\mathcal{P} = \left\{ p_{ij} = \paren{ \round{x} + i, \round{y} + j} | a \leq i, j \leq b \right\}$.
For the case in \figref{fig:kernel_supple}, we note that $a = -1$ and $b = 1$ and therefore $\lvert \mathcal{P} \rvert = 9$.
Then, we calculate \emph{relative} offsets of each point with respect to the reference $\paren{ x, y }$, i.e., $\mathbf{o}_{ij} = p_{ij} - \paren{x, y}$, on a regular grid.
Finally, we find $k \times k$ many vectors $\mathbf{o}'_{ij}$, which is a representation of the offsets $\mathbf{o}_{ij}$ using the orthogonal basis $\left\{ \mathbf{e}'_x, \mathbf{e}'_y \right\}$ as follows:
\begin{equation}
    \mathbf{o}_{ij}^{'\mathsf{T}} = 
    \begin{pmatrix}
        A^{-1} & 0 \\
        0 & B^{-1}
    \end{pmatrix}
    \begin{pmatrix}
        \cos{\omega} & \sin{\omega} \\
        -\sin{\omega} & \cos{\omega}
    \end{pmatrix}
    \mathbf{o}_{ij}\transpose.
\end{equation}
Using the calculated offset vector $\mathbf{o}_{ij}^{'\mathsf{T}}$, we rescale the original offsets and evaluate the kernel function as described in our main manuscript.
We note that in traditional bicubic interpolation method, resampling basis is fixed to $\left\{ \paren{1, 0}, \paren{0, 1} \right\}$ across all output positions $\paren{x', y'}$.
For the Adaptive configuration in Table~\fakeref{2} of our main manuscript, we use the conventional cubic spline to calculate the weight for each offset vector $\mathbf{o}_{ij}^{'\mathsf{T}}$.
In the proposed AWL, we concatenate $k \times k$ many vectors and fed them to a learnable network $\mathcal{K}$.
By doing so, a $k \times k$ resampling kernel $\mathbf{k} \paren{ x', y' }$ can be calculated at once.
We provide more detailed visualization of AWL in \secref{sec:visualize} and implementation of the kernel estimation module $\mathcal{K}$ in \secref{sec:arch}.

\section{Visualizing Spatially-Varying Property}
\label{sec:visualize}
We analyze how the proposed SRWarp operates in a spatially-varying manner.
\figref{fig:sv_kernels} shows our adaptive resampling grids for different local regions.
Specifically, when a local region is shrunk to a certain direction, the green ellipse is stretched to the particular axis.
Then, points in such direction are pulled to the origin when calculating the kernel function and contribute more, i.e., have larger weights.
The same thing happens oppositely in the case of locally enlarging distortions.
Our SRWarp can reconstruct high-quality images without severe artifacts and aliasing by considering the spatially-varying coordinate system in the resampling process.
Combined with the learnable kernel estimator $\mathcal{K}$, the adaptive grid can improve our SRWarp model without requiring \emph{any} additional parameters.

\begin{figure*}[t!]
    \centering
    \hfill
    \includegraphics[width=0.7484\linewidth]{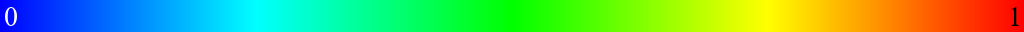}
    \\
    \vspace{-3mm}
    \subfloat[$\img{SR}$]{\includegraphics[width=0.245\linewidth]{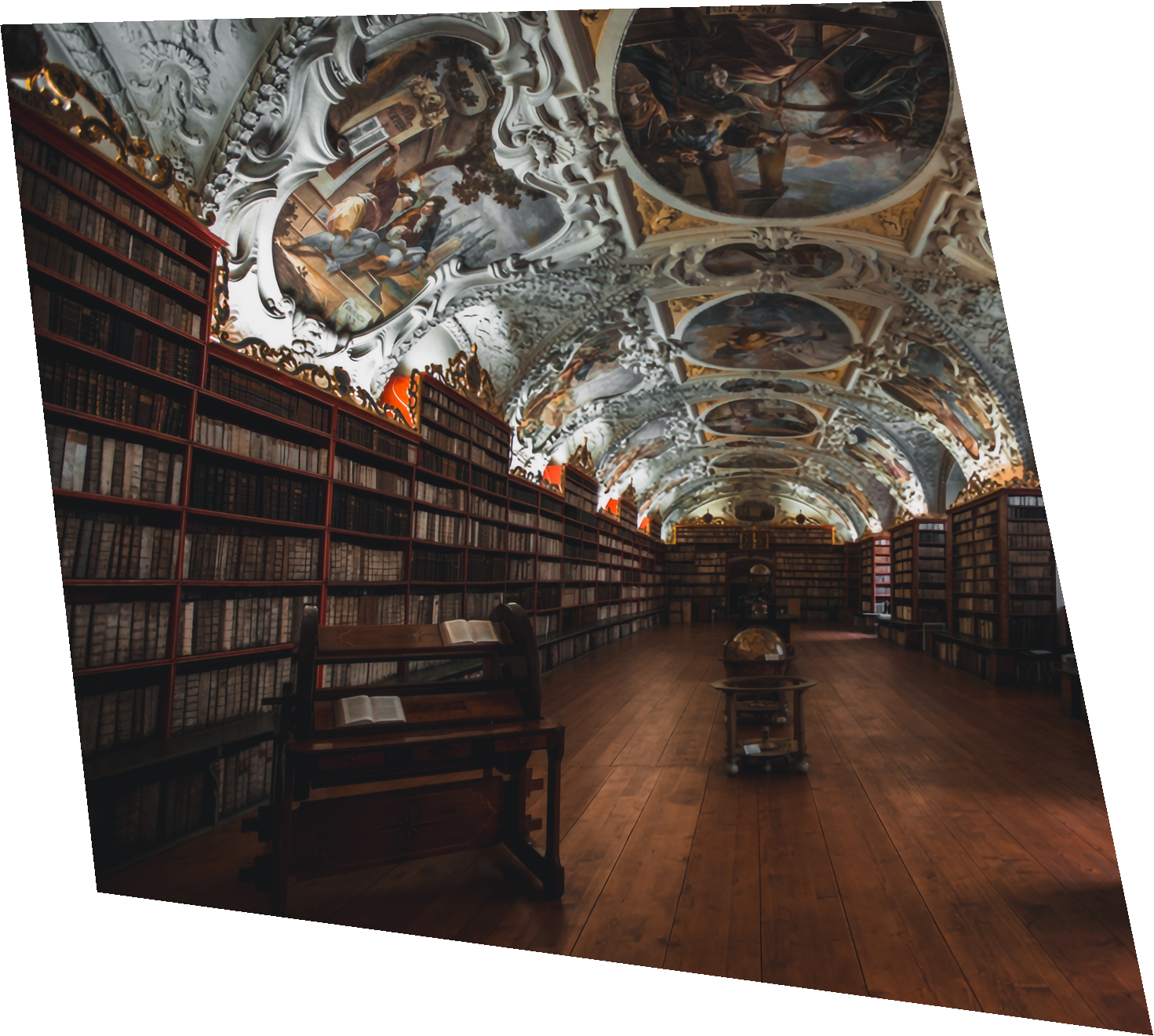}}
    \hfill
    \subfloat[$\lvert w_{\times 1} \rvert$ (with $\mathbf{C}$)\label{fig:ms_weight_w1}]{\includegraphics[width=0.245\linewidth]{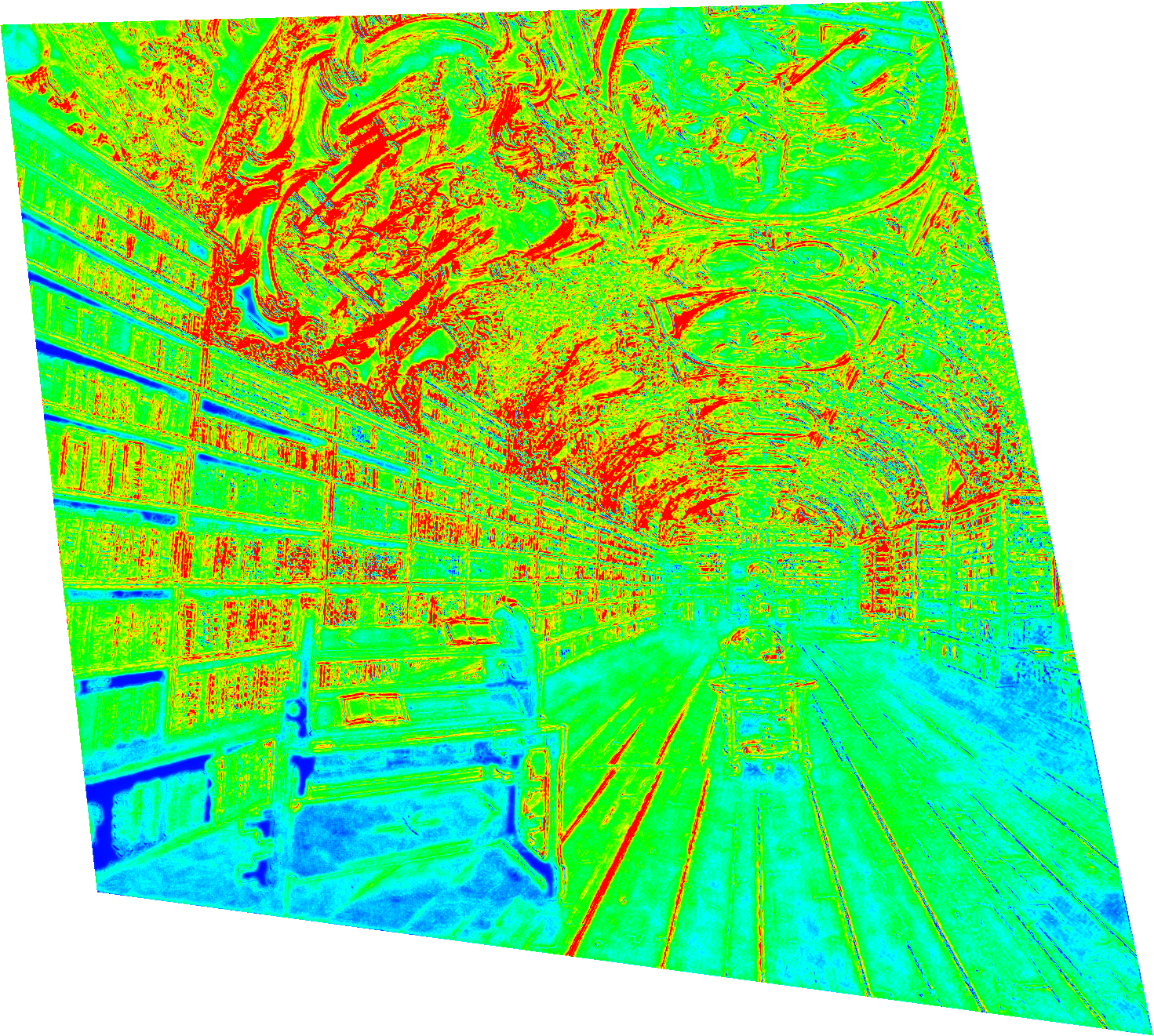}}
    \hfill
    \subfloat[$\lvert w_{\times 2} \rvert$ (with $\mathbf{C}$)\label{fig:ms_weight_w2}]{\includegraphics[width=0.245\linewidth]{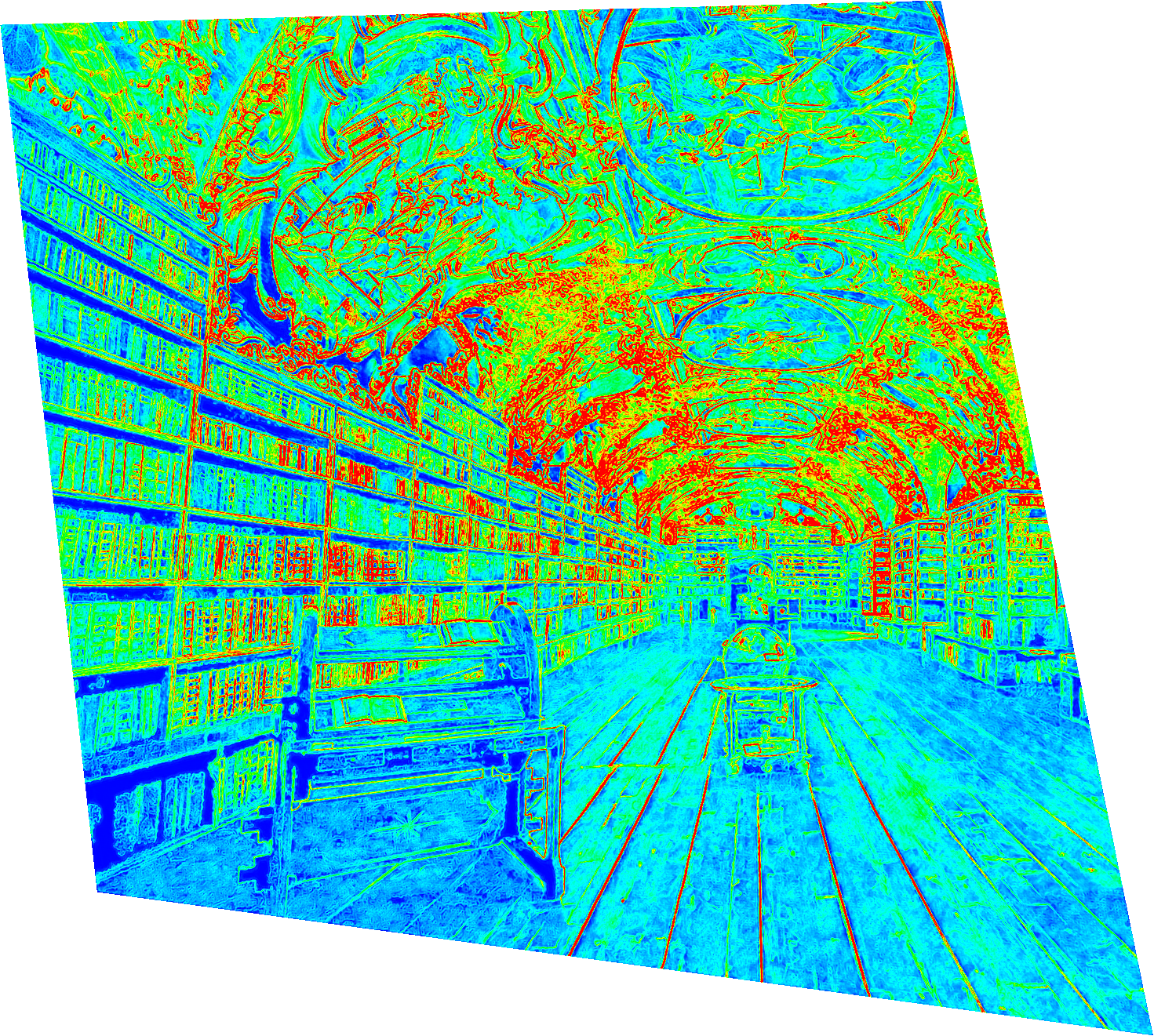}}
    \hfill
    \subfloat[$\lvert w_{\times 4} \rvert$ (with $\mathbf{C}$)\label{fig:ms_weight_w4}]{\includegraphics[width=0.245\linewidth]{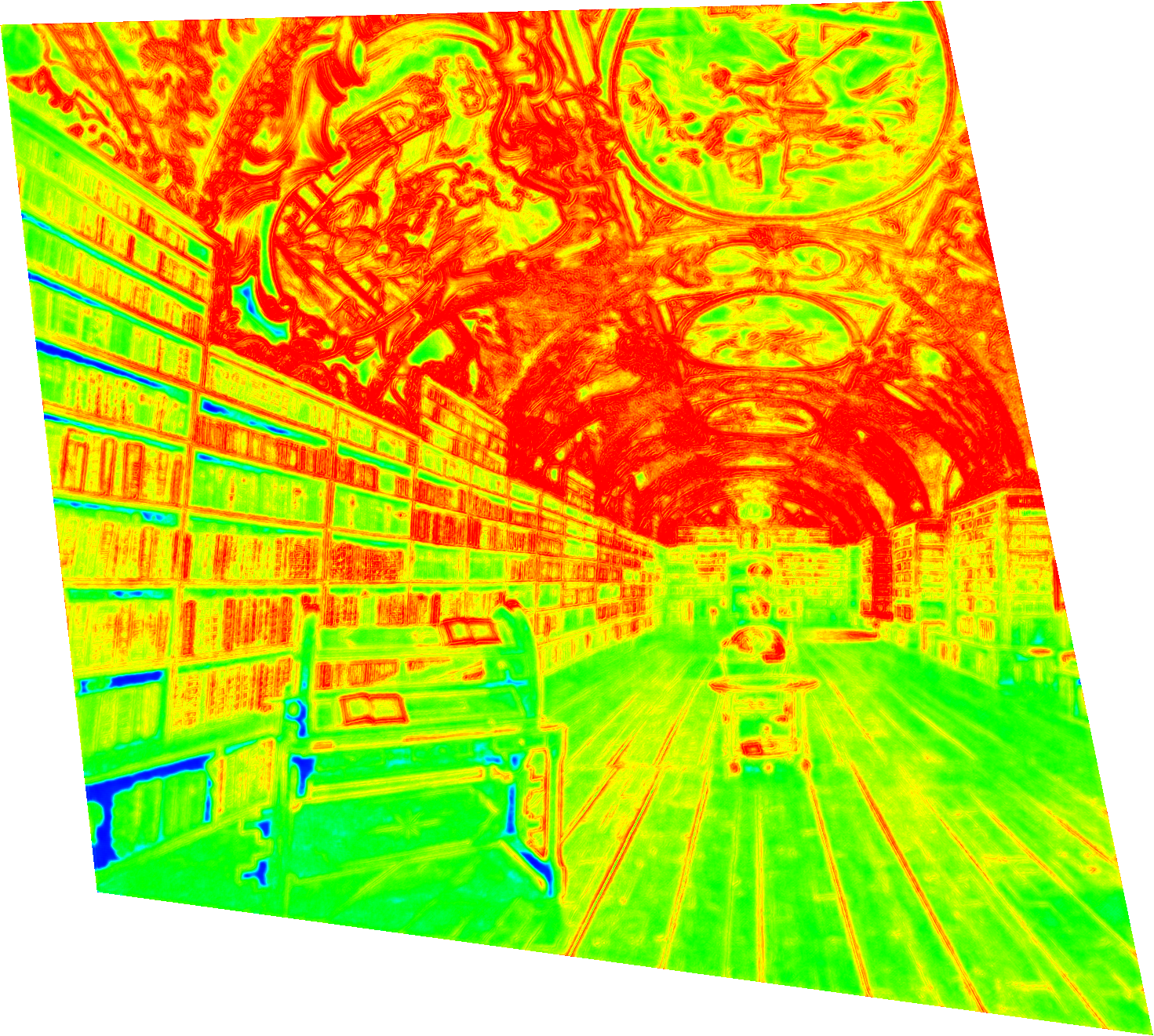}}
    \\
    \vspace{-3mm}
    \addtocounter{subfigure}{-1}
    \subfloat{\includegraphics[width=0.245\linewidth]{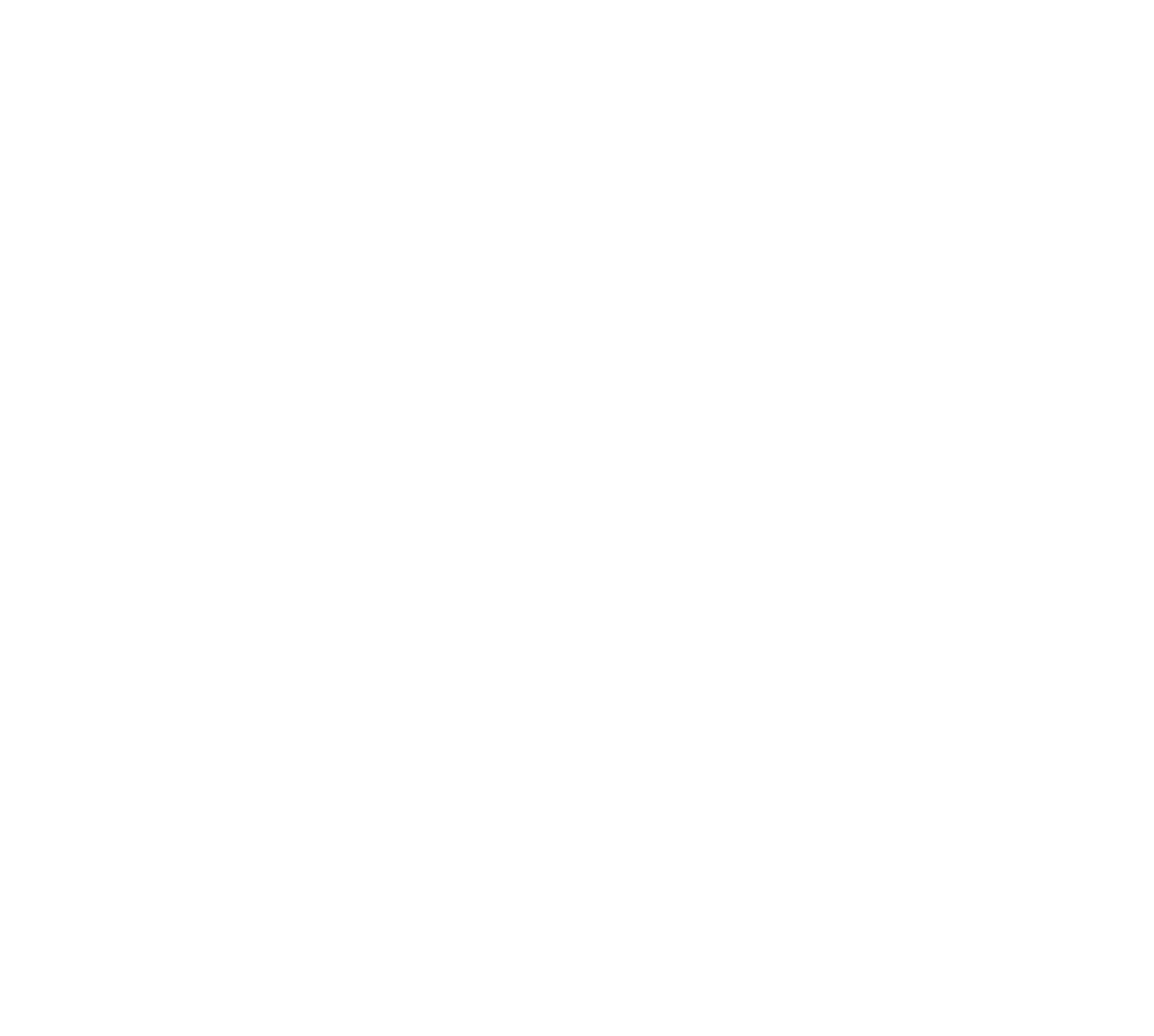}}
    \hfill
    \subfloat[$\lvert w_{\times 1} \rvert$ (w/o $\mathbf{C}$)\label{fig:ms_weight_wo1}]{\includegraphics[width=0.245\linewidth]{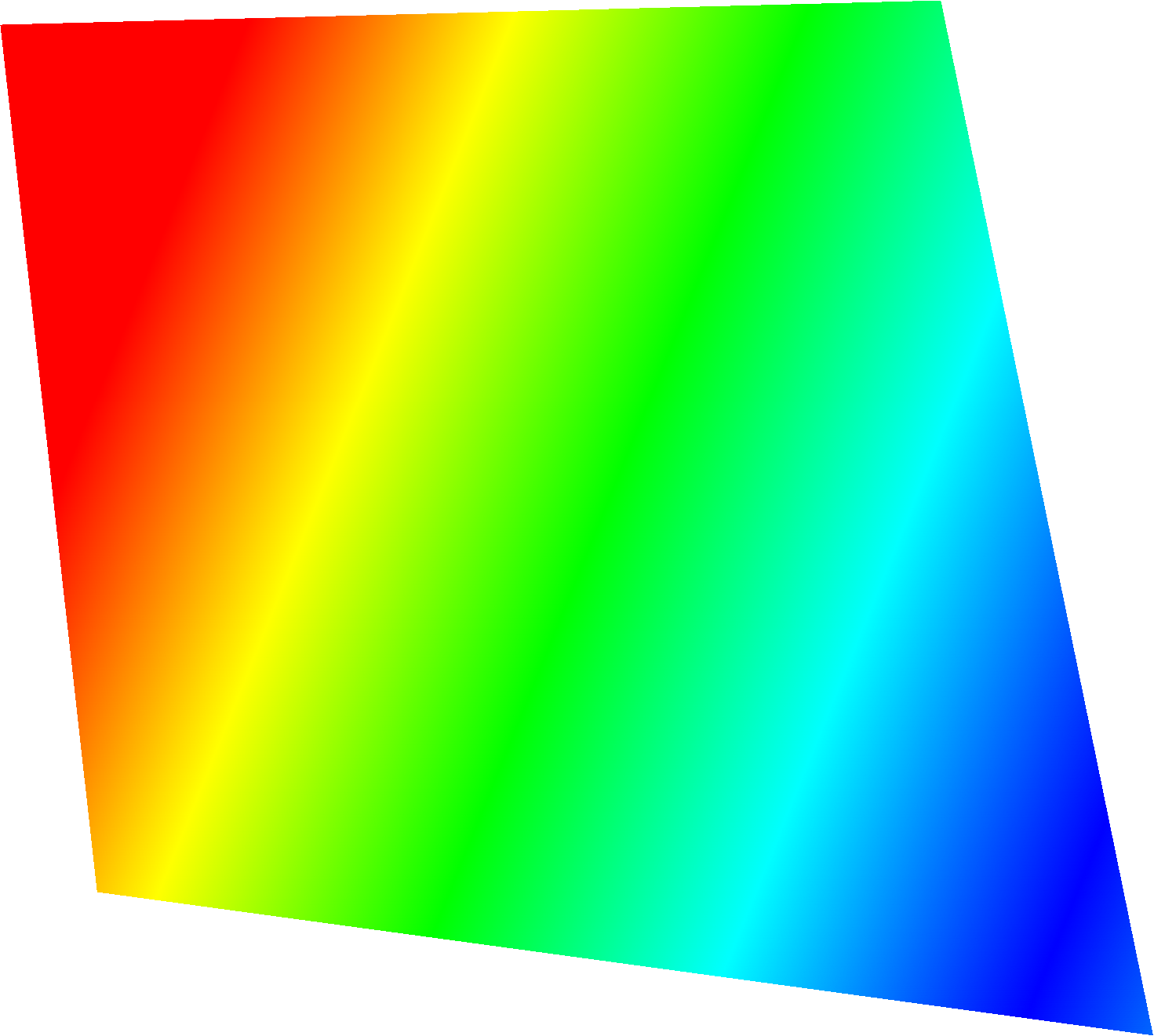}}
    \hfill
    \subfloat[$\lvert w_{\times 2} \rvert$ (w/o $\mathbf{C}$)\label{fig:ms_weight_wo2}]{\includegraphics[width=0.245\linewidth]{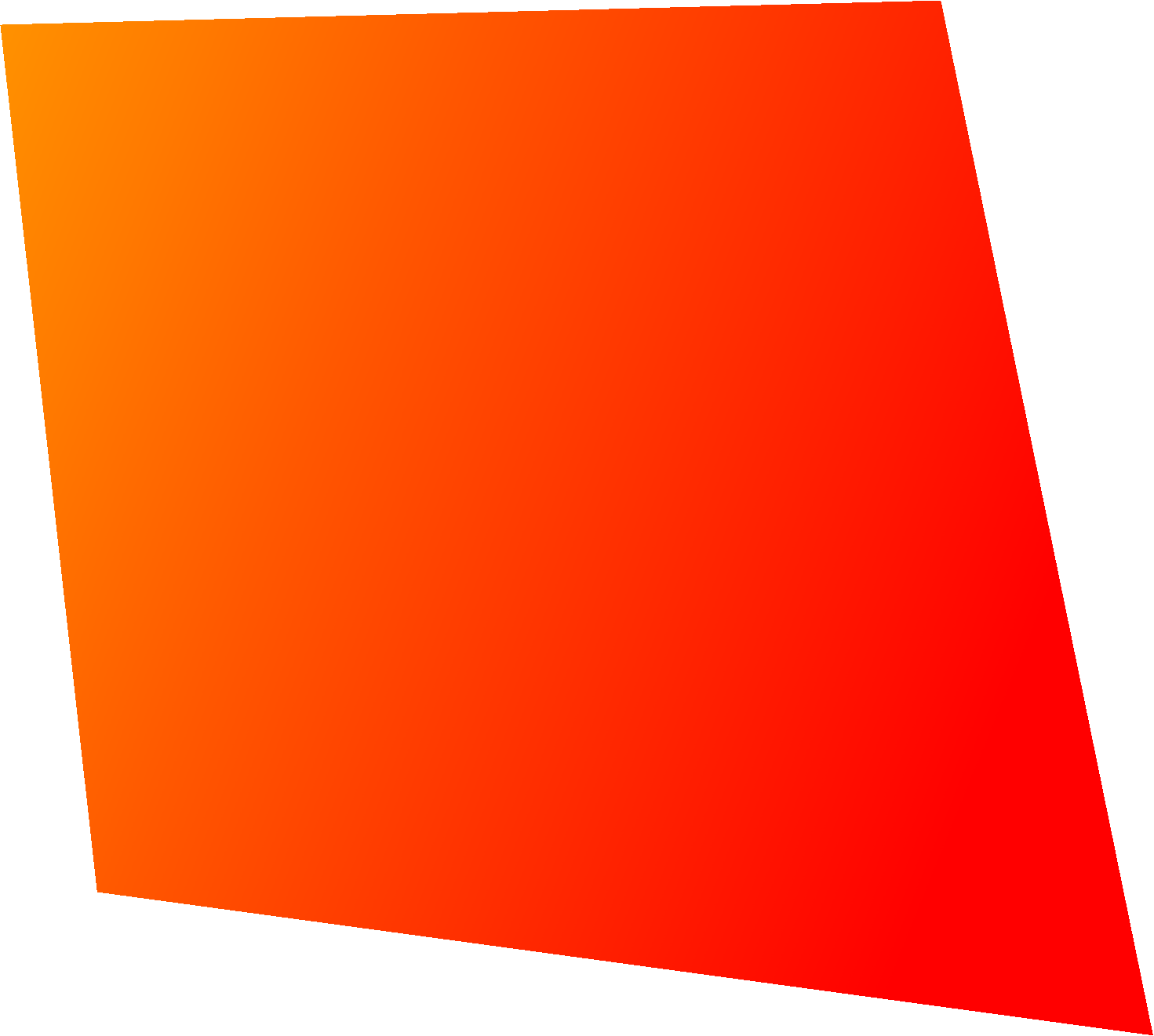}}
    \hfill
    \subfloat[$\lvert w_{\times 4} \rvert$ (w/o $\mathbf{C}$)\label{fig:ms_weight_wo4}]{\includegraphics[width=0.245\linewidth]{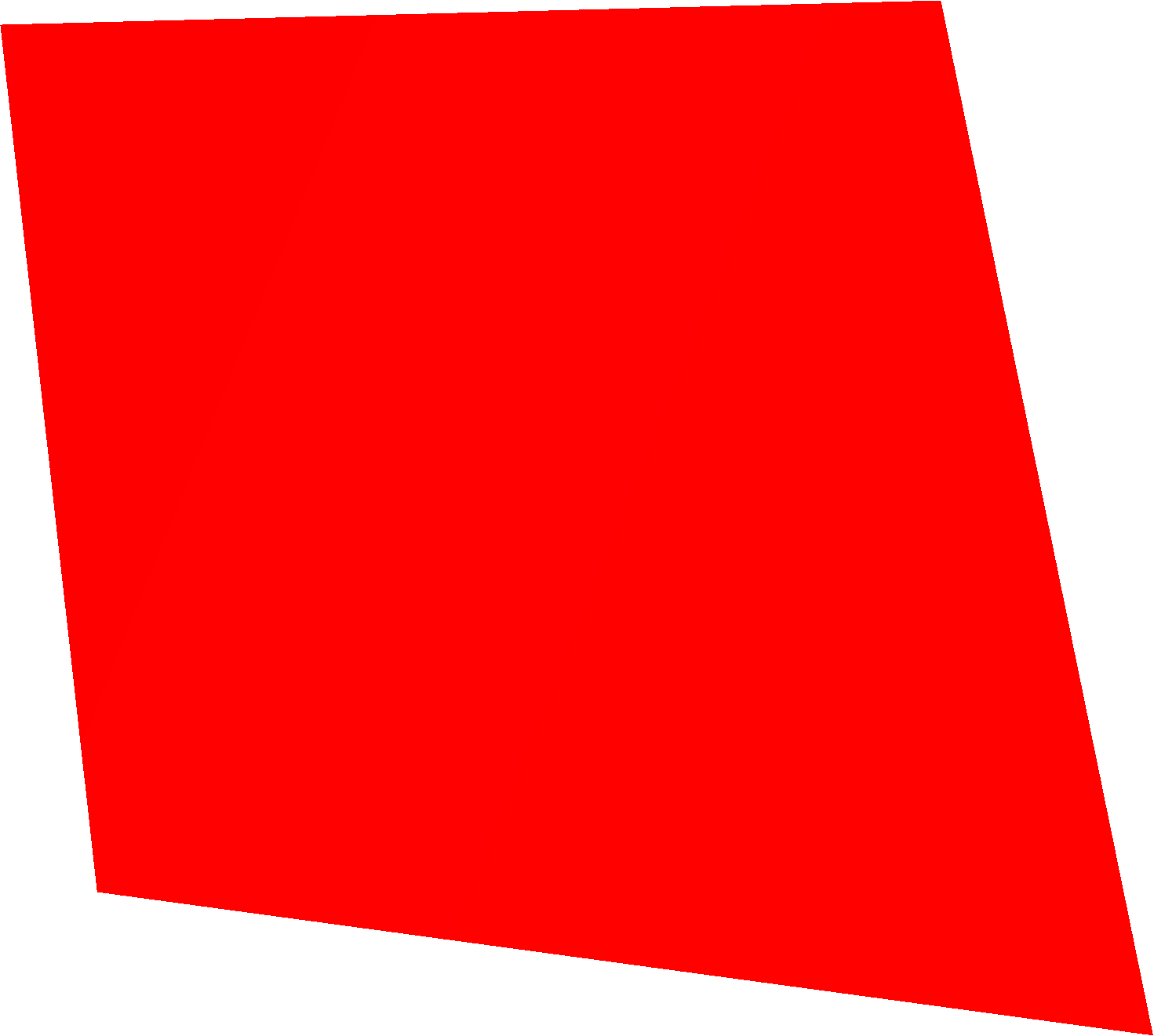}}
    \\
    \figcspace
    \caption{
        \textbf{Spatially-varying property of multi-scale blending weights.}
        We normalize the map $w_{\times s}$ \emph{within} each sample for better visualization.
        (e$-$g) We do not use the content feature $\mathbf{C}$ to estimate the weights $w_{\times s}$ and observe how the scale feature $\mathbf{S}$ affects the blending coefficients.
        The weights may not represent the relative importance of different scales $s$ as the features $\mathbf{W}_{\times s}$ are not normalized.
        Instead, the values indicate which regions are weighted more in each different scale.
    }
    \label{fig:ms_weight}
    \figspace
\end{figure*}

In \figref{fig:ms_weight}, we demonstrate how the multiscale blending coefficients vary depending on different contents and local distortions.
As shown in \figref{fig:ms_weight_w2} and \figref{fig:ms_weight_w4}, $\times 2$ and $\times 4$ features play an important role around edges.
On the other hand, the activation of the $\times 1$ feature in \figref{fig:ms_weight_w1} is widely distributed across the scene for low-frequency structures.
We note that our multiscale blending module places a high priority on fine-grained textures and details regardless of the feature scale $s$, showing that single-scale information may not be enough for the warping task.
We also validate the effectiveness of the scale feature $\mathbf{S}$ in (\fakeref{7}) of our main manuscript by ignoring $\mathbf{C}$ from the blending module.
Since the $\times 4$ feature plays a critical role in upsampling a given image, it is equally weighted across the target domain, as shown in \figref{fig:ms_weight_wo4}.
In contrast, relative contributions of $\times 1$ and $\times 2$ features show spatially-varying behavior, demonstrating the importance of considering local distortion.
Specifically, the $\times 1$ feature becomes less weighted with larger distortion (bottom right of \figref{fig:ms_weight_wo1}), while it can provide meaningful information to our SRWarp model if the deformation is not significant (top left of \figref{fig:ms_weight_wo1}).

\section{Detailed Model Architectures}
\label{sec:arch}
\begin{figure}[t]
    \centering
    \includegraphics[width=\linewidth]{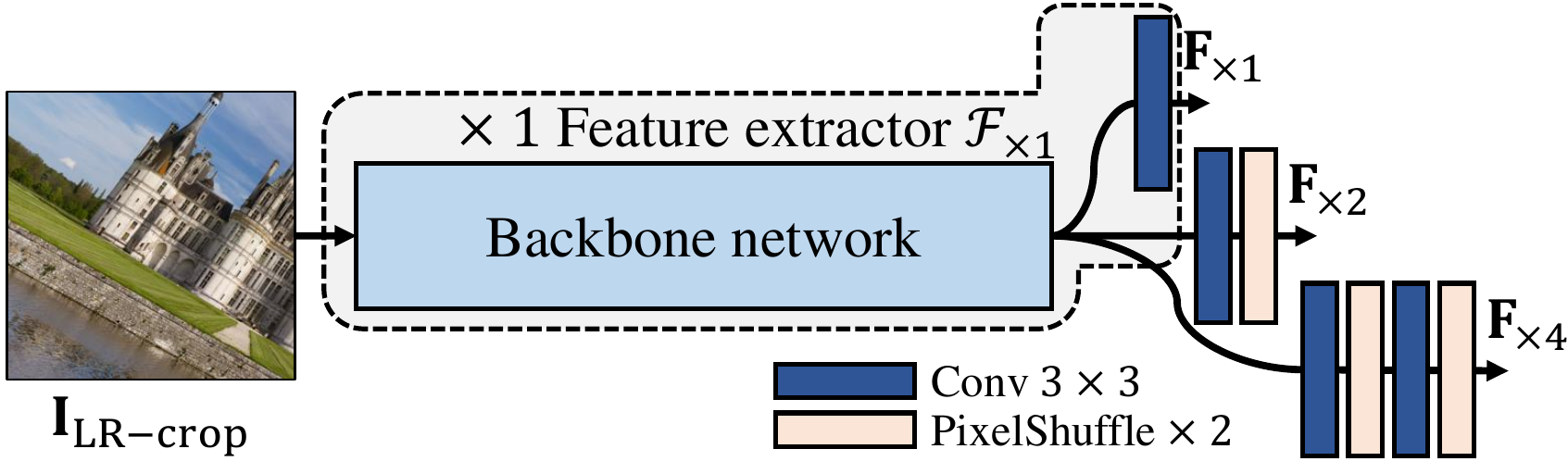} \\
    \figcspace
    \caption{
        \textbf{Multiscale feature extractor $\mathcal{F}_{\times s}$ in our SRWarp model.}
        For the modified MDSR structure, we adopt the MDSR~\cite{sr_edsr} model without scale-specific branches as the backbone network.
        In the MRDB architecture, the state-of-the-art RRDB~\cite{sr_esrgan} model without upsampling module is used for the backbone.
        We note that our feature extractors $\mathcal{F}_{\times s}$ share many parts in common across different scales $s$ for efficient computation~\cite{sr_edsr}.
        The $\times 1$ feature extractor $\mathcal{F}_{\times 1}$ is emphasized in the figure as an example.
        Each of the scale-specific features $\mathbf{F}_{\times s}$ has 64 channels.
    }
    \label{fig:backbone}
    \vspace{3cm}
\end{figure}

\Paragraph{Multiscale feature extractor.}
\figref{fig:backbone} demonstrates the modified MDSR~\cite{sr_edsr} and MRDB backbone architectures we introduce for efficient multiscale respresentation.
Since the proposed SRWarp model is designed to handle the SR task of arbitrary scales and shapes, we remove the scale-specific branch in front of the original MDSR structure.
We also replace $\times 2$, $\times 3$, and $\times 4$ upscaling modules to $\times 1$, $\times 2$, and $\times 4$ feature extractors.
Following the baseline model by Lim~\etal~\cite{sr_edsr}, we adopt 16 ResBlocks with 64 channels each for the main branch.
Our MRDB model is a multiscale version of the state-of-the-art RRDB~\cite{sr_esrgan} network.
We change the last upscaling layer to the proposed scale-specific feature extractors, similar to the modified MDSR model.

\begin{figure}[t]
    \centering
    \includegraphics[width=\linewidth]{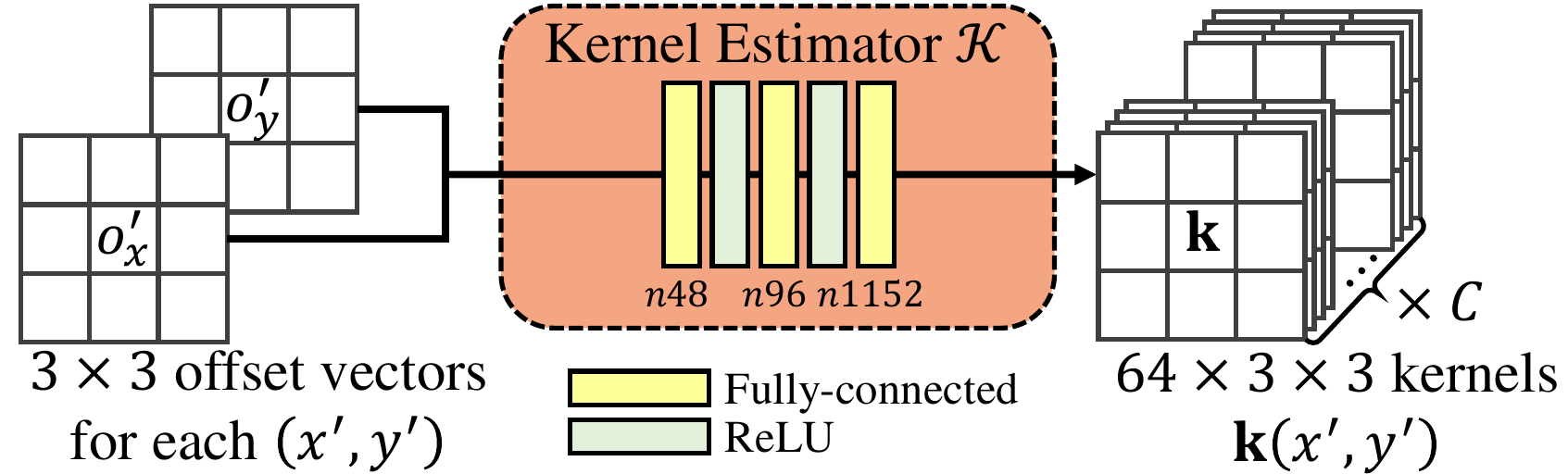} \\
    \figcspace
    \caption{
        \textbf{Structure of the kernel estimator $\mathcal{K}$ in our adaptive warping layer.}
        The module takes 18 input values for each point $\paren{ x', y' }$ in the target coordinate.
        $n$ describes the output dimension of fully-connected layers.
        We use $C = 64$ channels for dynamic kernels.
    }
    \label{fig:kernel_estimator}
    \figspace
\end{figure}

\begin{figure}[t]
    \centering
    \includegraphics[width=\linewidth]{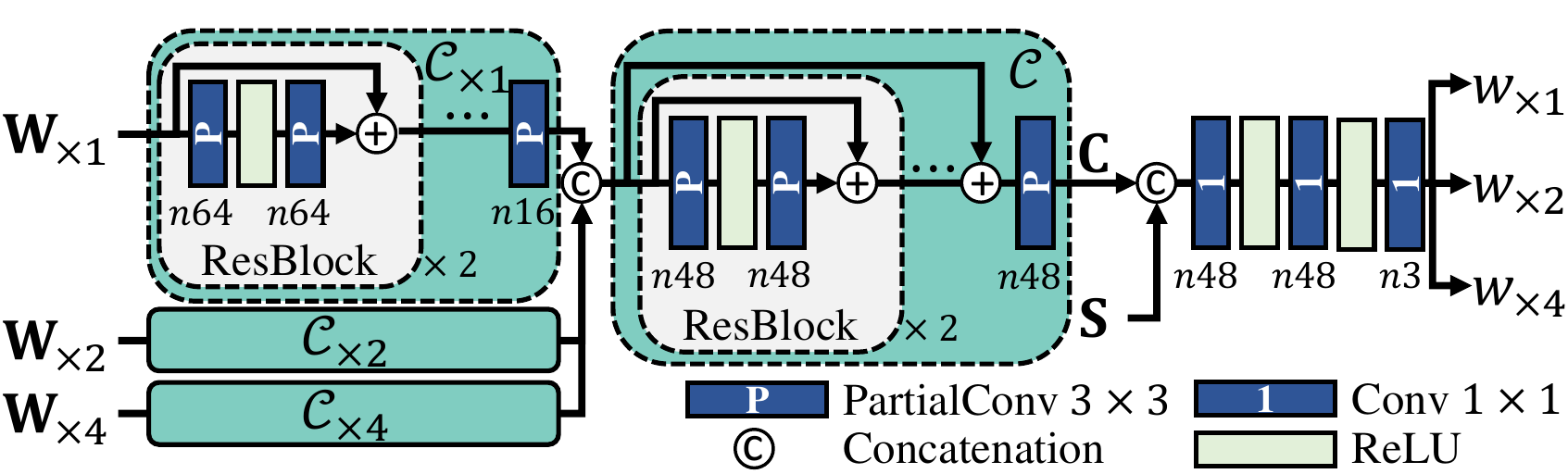} \\
    \figcspace
    \caption{
        \textbf{Weight prediction in our multiscale blending module.}
        We omit the input mask $\mathbf{m}$ for the partial convolutional layers for simplicity.
        The blended feature $\mathbf{W}_\text{blend}$ is calculated from the warped features $\mathbf{W}_{\times s}$ and their corresponding weights $w_{\times s}$ as described in (\fakeref{8}).
        Best viewed with Section~\fakeref{3.3} in our main manuscript.
    }
    \label{fig:ms_blending}
    \figspace
\end{figure}

\begin{figure}[t]
    \centering
    \includegraphics[width=\linewidth]{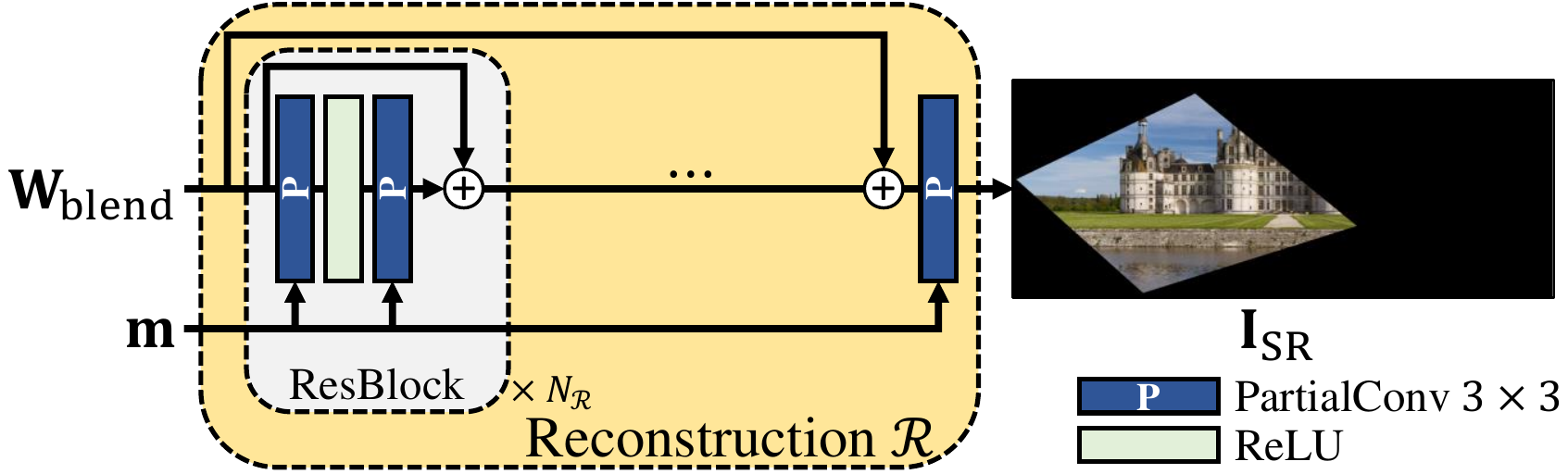} \\
    \figcspace
    \caption{
        \textbf{Structure of the reconstruction module $\mathcal{R}$.}
        We use $N_\mathcal{R} = 5$ many ResBlocks wich 64 output channels.
    }
    \label{fig:recon}
    \figspace
\end{figure}

\Paragraph{SRWarp architecture.}
Figure~\fakeref{4} in our main manuscript illustrates the overall pipeline of the proposed SRWarp method.
We construct the main modules with residual blocks~\cite{sr_srgan, sr_edsr} and skip connections~\cite{sr_vdsr} for stable and efficient training.
\figref{fig:kernel_estimator}, \figref{fig:ms_blending}, and \figref{fig:recon} show how the actual implementations of our kernel estimator, multiscale blending, and reconstruction modules are, respectively.

\begin{figure}
    \centering
    \subfloat[A-R configuration]{\includegraphics[width=\linewidth]{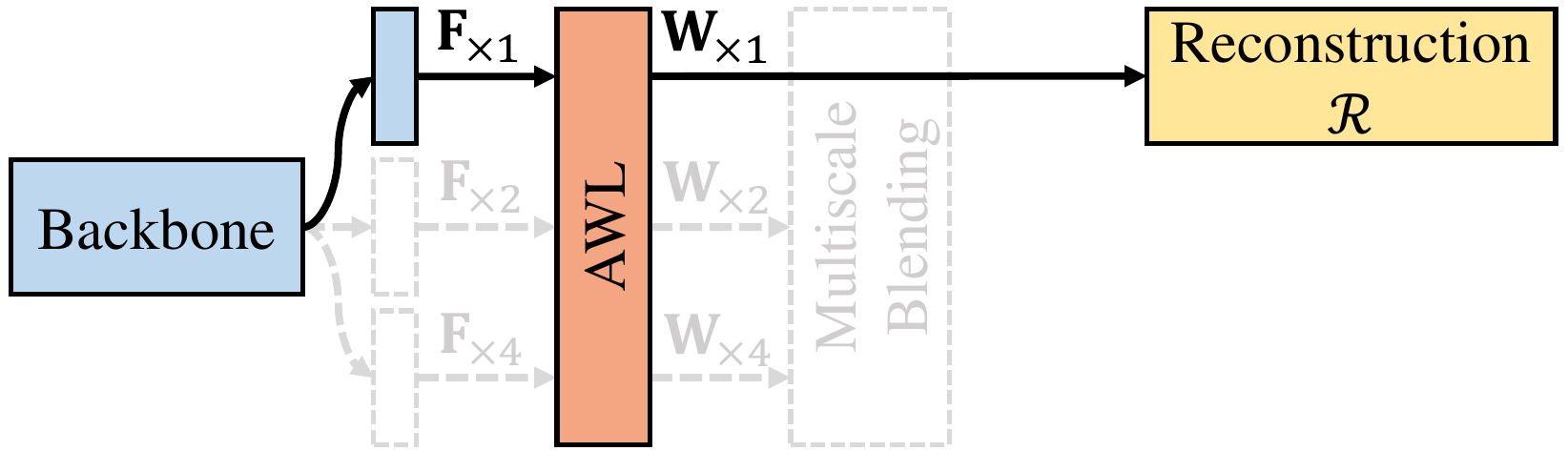}}
    \\
    \vspace{-3mm}
    \subfloat[A-M configuration]{\includegraphics[width=\linewidth]{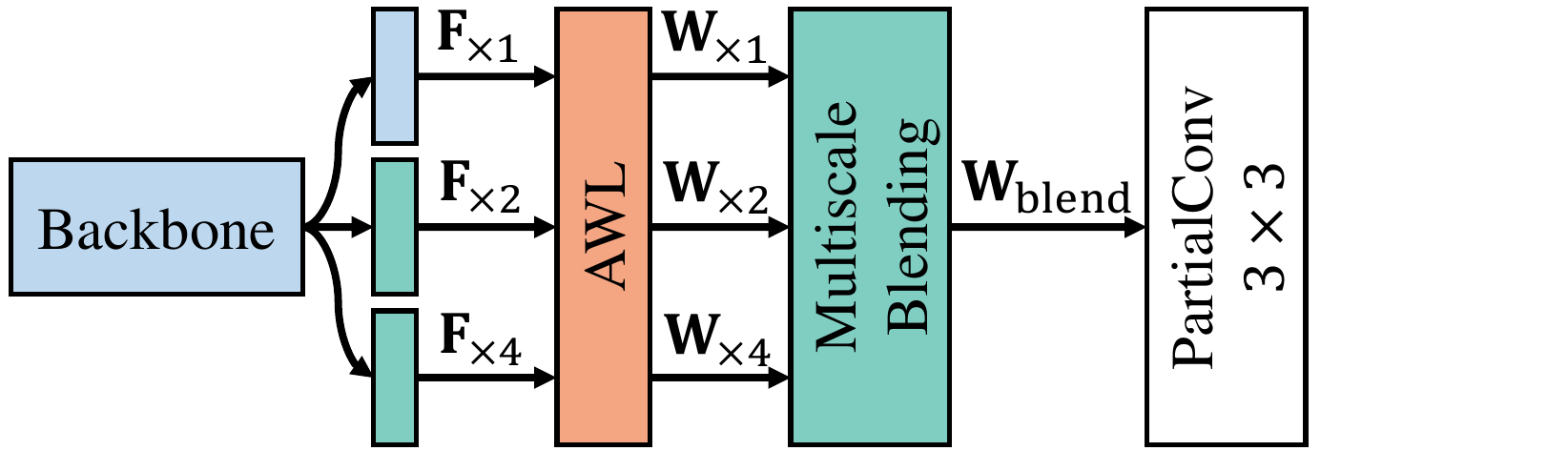}}
    \\ 
    \vspace{-3mm}
    \subfloat[M-R configuration]{\includegraphics[width=\linewidth]{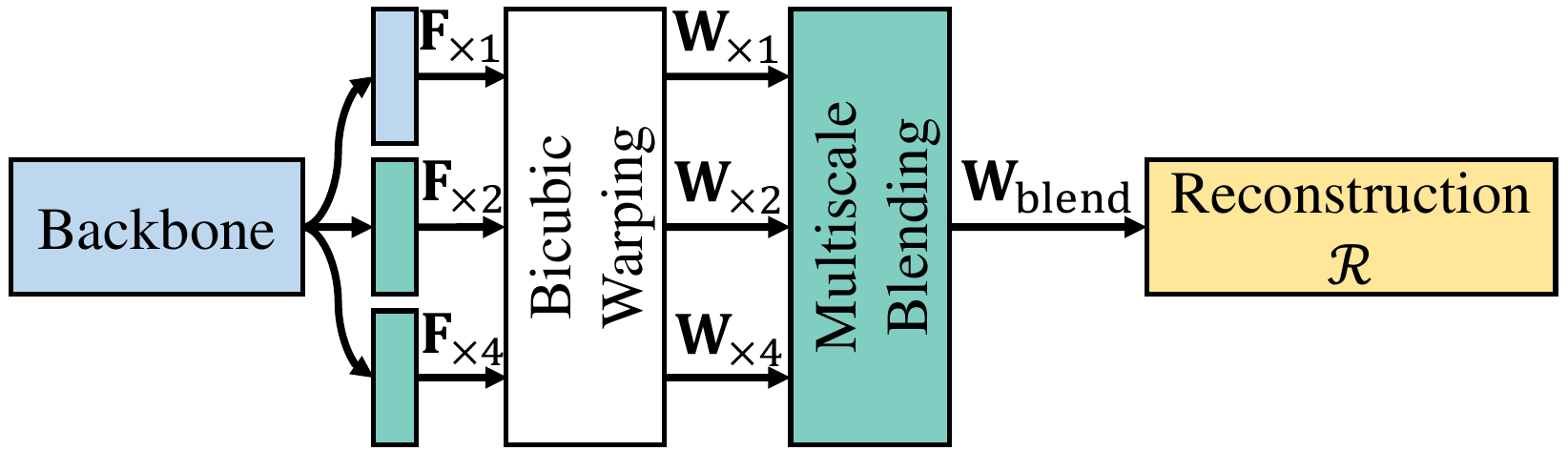}}
    \\
    \figcspace
    \caption{
        \textbf{Examples of the SRWarp architectures $\mathcal{S}$ for the ablation study.}
        Best viewed with Figure~\fakeref{4} and Table~\fakeref{1} in our main manuscript.
        (c) Bicubic warping denotes a differentiable warping layer with the conventional bicubic interpolation, without our adaptive resampling grid and kernel estimator $\mathcal{K}$.
    }
    \label{fig:arch_abl}
    \figspace
\end{figure}

\Paragraph{Architectures for the ablation study.}
\figref{fig:arch_abl} describes detailed model structures for Table~\fakeref{1} in the main manuscript.
We note that $\times 1$ feature $\mathbf{F}_{\times 1}$ is fed to the AWL since the module replaces conventional upsampling layers in SR networks.

\section{Detailed Training Configurations}
For stable convergence, we pre-train the SRWarp network on the fractional-scale DIV2K dataset with a mini-batch size of $N = 16$.
The model is updated for 600 epochs, where each epoch consists of 1,000 iterations.
We set the initial learning rate to $10^{-4}$ and halve it after 150, 300, and 450 epochs.
Then, we use a mini-batch of size $N = 4$ to fine-tune our SRWarp model on the DIV2KW dataset.
The network is updated for 200 epochs with an initial learning rate of $5 \times 10^{-5}$, which is halved after 100, 150, and 175 epochs.
For the small modified MDSR architecture, the training takes about 12 hours on a single RTX 2080 Ti GPU.
With the larger MRDB backbone, we optimize the model for about 60 hours on two RTX 2080 Ti GPUs.

\section{Efficiency of the Proposed Method}
\tabref{tab:efficiency} compares the actual runtime and peak memory usage of the proposed SRWarp against the meta-upscaling approaches~\cite{sr_meta} on the arbitrary-scale SR task.
We use a single RTX 2080 Ti GPU for all executions to fairly compare different methods in a unified environment.
The SRWarp model requires much lower memory than Meta-EDSR and Meta-RDN while achieving comparable performances, as shown in Table~\fakeref{4} in our main manuscript.
Notably, the proposed method is about 2.5 times faster than the Meta-RDN model while achieving better performance.

\begin{table}[t]
    \centering
    \begin{tabularx}{\linewidth}{l >{\centering\arraybackslash}X c}
        \toprule
        Method & Peak Mem. (GB) & Runtime (ms) \\
        \midrule
        Meta-EDSR~\cite{sr_meta} & 4.96 & 218 \\
        Meta-RDN~\cite{sr_meta} & 5.15 & 253 \\
        \midrule
        \textbf{SRWarp (MRDB)} & 3.08 & 155 \\
        \bottomrule
    \end{tabularx}
    \\
    \tabcspace
    \caption{
        \textbf{Efficiency comparison between the proposed method and meta-upscaling.}
        The evaluation is done on the fractional-scale B100~\cite{data_bsd200} dataset with a scale factor of $\times 3$.
        The runtime is measured over 100 test images in a unified environment and averaged \emph{excluding} initialization, I/O, and the other overheads.
        We also note that the peak memory usage does not correspond to the exact amount of memory each model consumes due to GPU-level optimization, caching, and many other practical reasons.
    }
    \label{tab:efficiency}
    \tabspace
\end{table}

\newcommand{\vgg}[1]{\mathcal{V}_\text{54} \paren{#1} }
\begin{figure*}[t]
    \centering
    \subfloat{\includegraphics[width=0.30\linewidth]{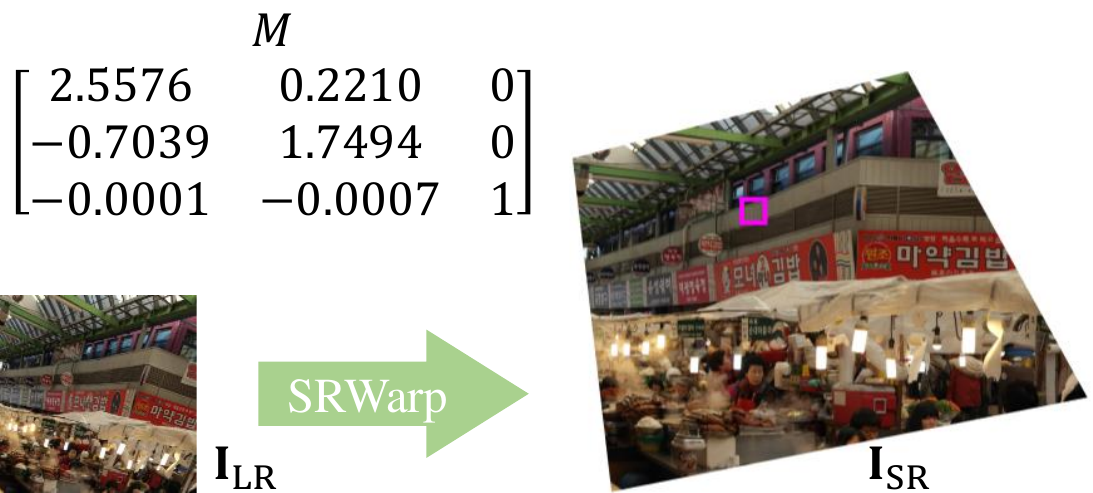}}
    \hfill
    \subfloat{\includegraphics[width=0.135\linewidth]{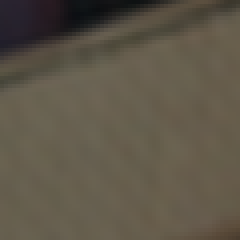}}
    \hfill
    \subfloat{\includegraphics[width=0.135\linewidth]{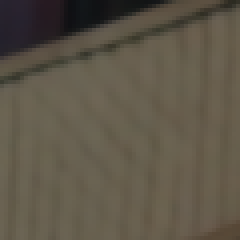}}
    \hfill
    \subfloat{\includegraphics[width=0.135\linewidth]{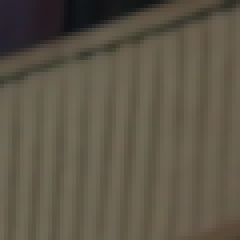}}
    \hfill
    \subfloat{\includegraphics[width=0.135\linewidth]{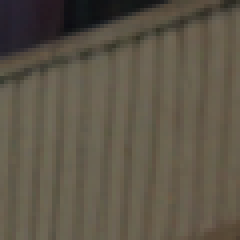}}
    \hfill
    \subfloat{\includegraphics[width=0.135\linewidth]{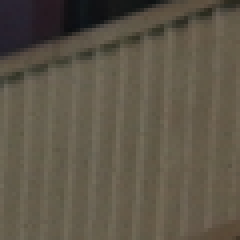}}
    \\
    \vspace{-3.5mm}
    \addtocounter{subfigure}{-6}
    \subfloat{\includegraphics[width=0.30\linewidth]{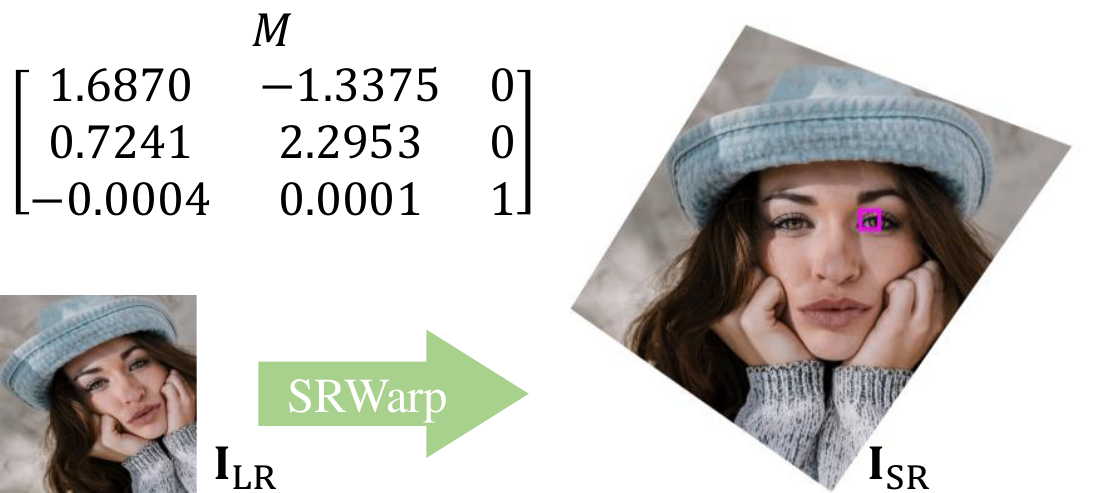}}
    \hfill
    \subfloat{\includegraphics[width=0.135\linewidth]{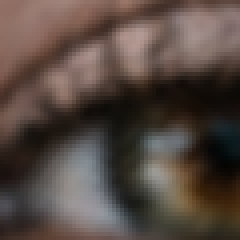}}
    \hfill
    \subfloat{\includegraphics[width=0.135\linewidth]{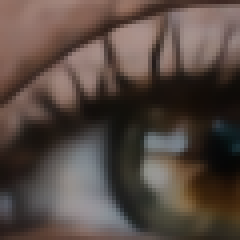}}
    \hfill
    \subfloat{\includegraphics[width=0.135\linewidth]{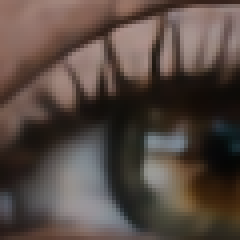}}
    \hfill
    \subfloat{\includegraphics[width=0.135\linewidth]{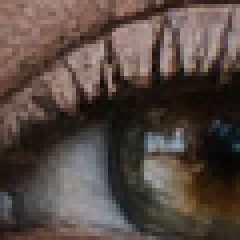}}
    \hfill
    \subfloat{\includegraphics[width=0.135\linewidth]{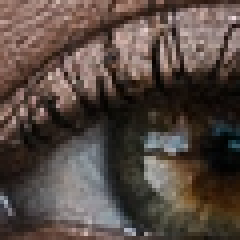}}
    \\
    \vspace{-3.5mm}
    \addtocounter{subfigure}{-6}
    \subfloat{\includegraphics[width=0.30\linewidth]{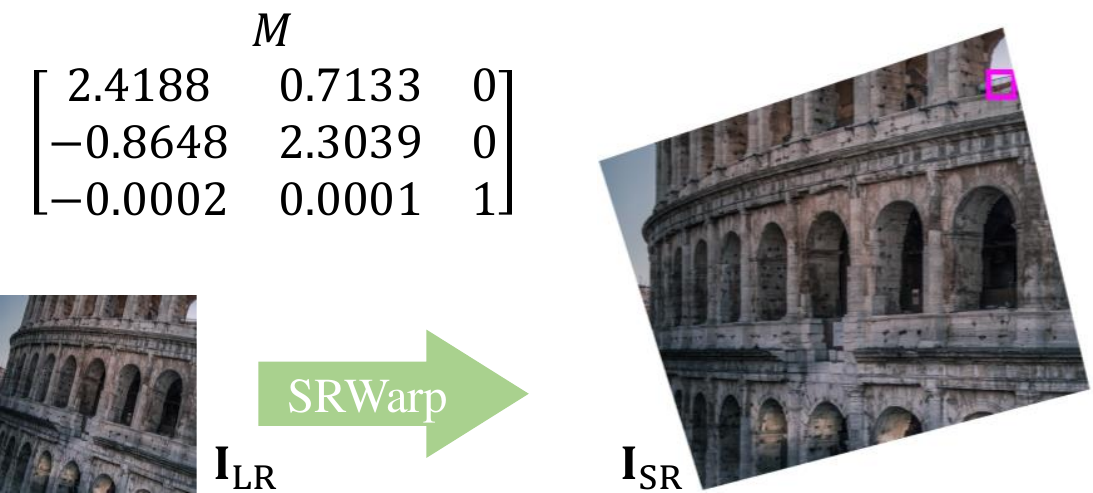}}
    \hfill
    \subfloat{\includegraphics[width=0.135\linewidth]{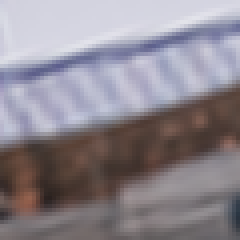}}
    \hfill
    \subfloat{\includegraphics[width=0.135\linewidth]{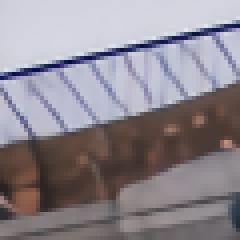}}
    \hfill
    \subfloat{\includegraphics[width=0.135\linewidth]{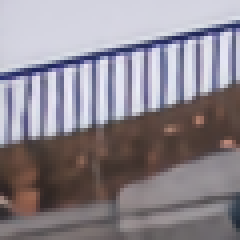}}
    \hfill
    \subfloat{\includegraphics[width=0.135\linewidth]{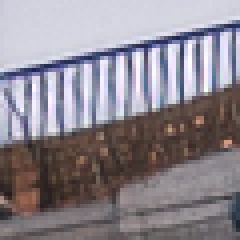}}
    \hfill
    \subfloat{\includegraphics[width=0.135\linewidth]{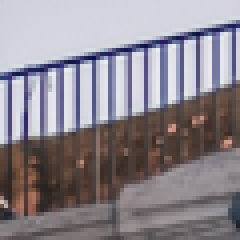}}
    \\
    \vspace{-3.5mm}
    \addtocounter{subfigure}{-6}
    \subfloat{\includegraphics[width=0.30\linewidth]{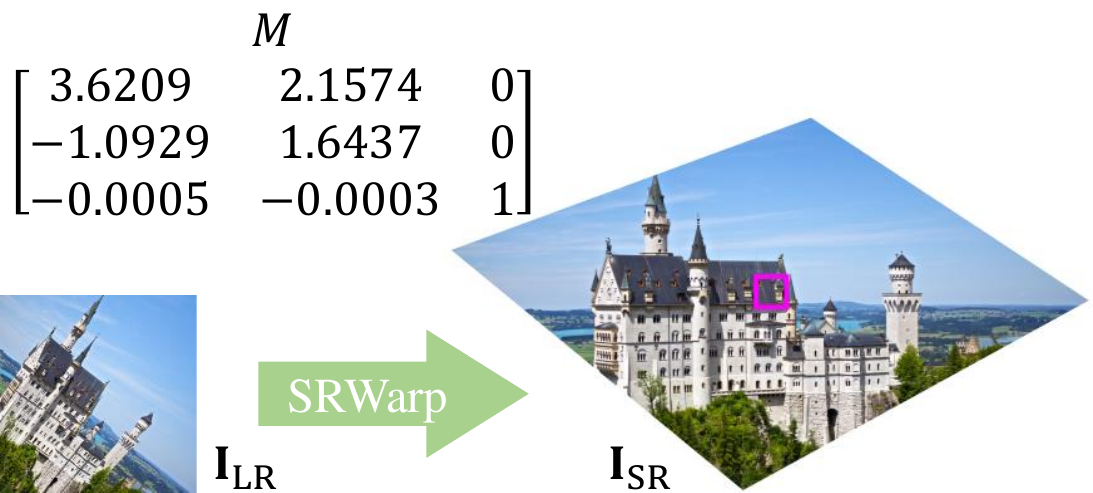}}
    \hfill
    \subfloat{\includegraphics[width=0.135\linewidth]{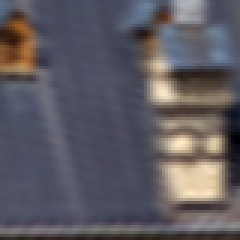}}
    \hfill
    \subfloat{\includegraphics[width=0.135\linewidth]{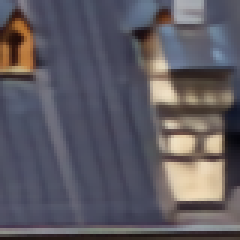}}
    \hfill
    \subfloat{\includegraphics[width=0.135\linewidth]{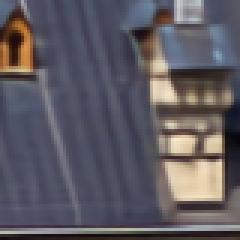}}
    \hfill
    \subfloat{\includegraphics[width=0.135\linewidth]{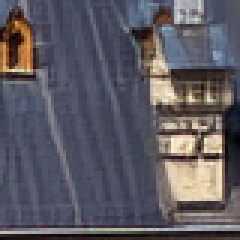}}
    \hfill
    \subfloat{\includegraphics[width=0.135\linewidth]{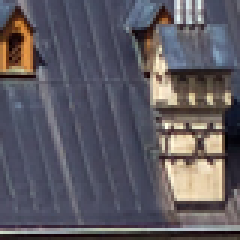}}
    \\
    \vspace{-3.5mm}
    \addtocounter{subfigure}{-6}
    \subfloat{\includegraphics[width=0.30\linewidth]{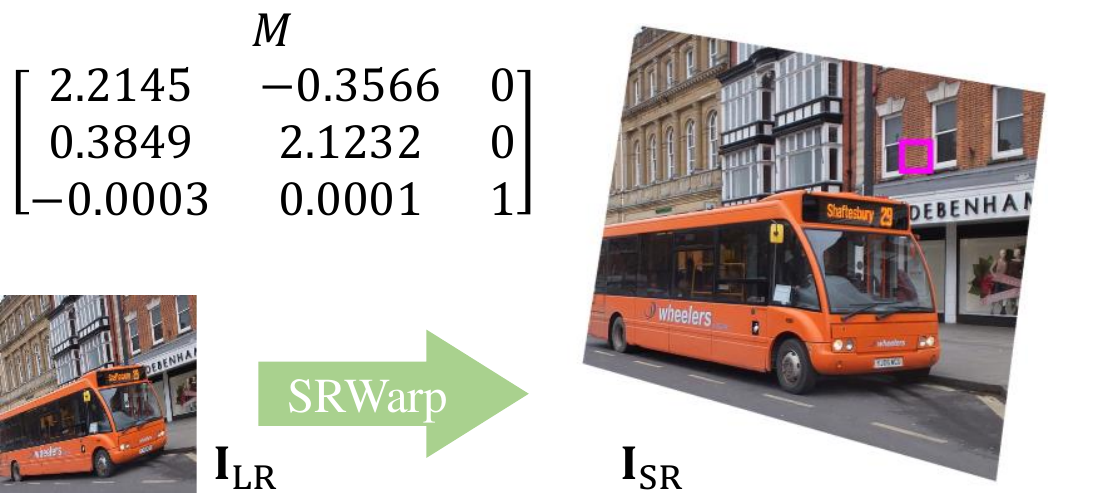}}
    \hfill
    \subfloat{\includegraphics[width=0.135\linewidth]{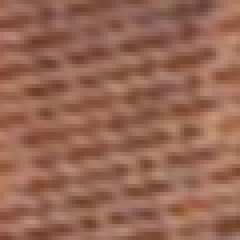}}
    \hfill
    \subfloat{\includegraphics[width=0.135\linewidth]{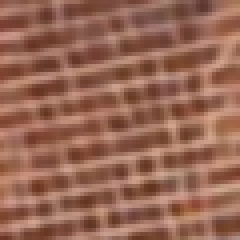}}
    \hfill
    \subfloat{\includegraphics[width=0.135\linewidth]{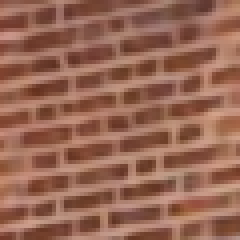}}
    \hfill
    \subfloat{\includegraphics[width=0.135\linewidth]{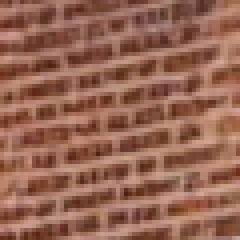}}
    \hfill
    \subfloat{\includegraphics[width=0.135\linewidth]{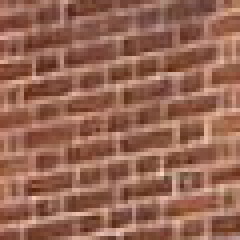}}
    \\
    \vspace{-3.5mm}
    \addtocounter{subfigure}{-6}
    \subfloat{\includegraphics[width=0.30\linewidth]{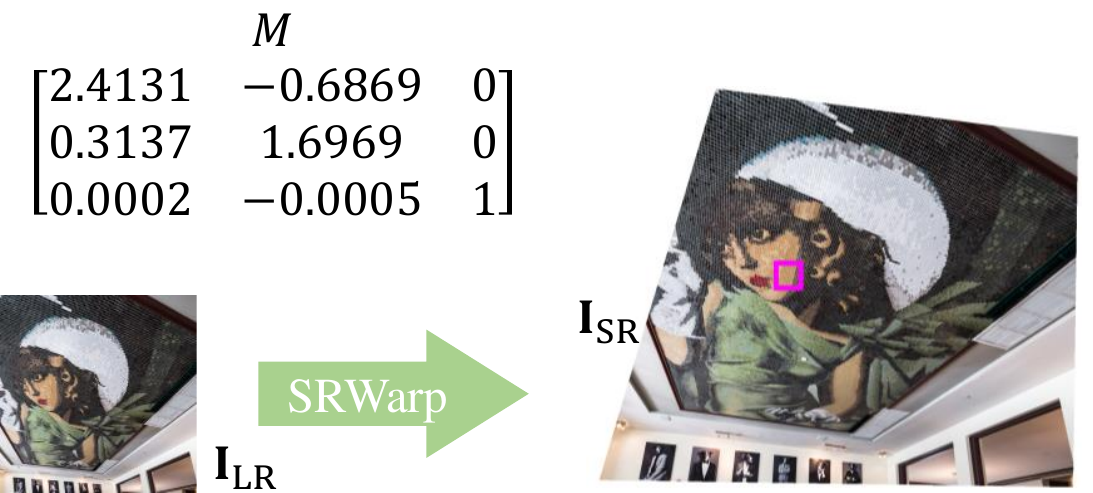}}
    \hfill
    \subfloat{\includegraphics[width=0.135\linewidth]{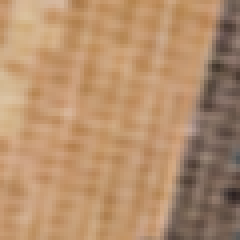}}
    \hfill
    \subfloat{\includegraphics[width=0.135\linewidth]{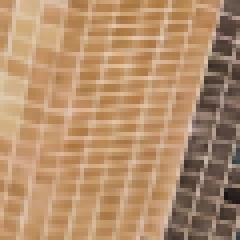}}
    \hfill
    \subfloat{\includegraphics[width=0.135\linewidth]{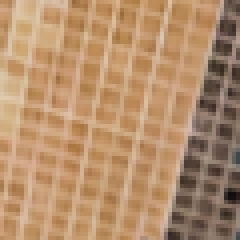}}
    \hfill
    \subfloat{\includegraphics[width=0.135\linewidth]{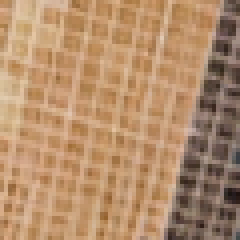}}
    \hfill
    \subfloat{\includegraphics[width=0.135\linewidth]{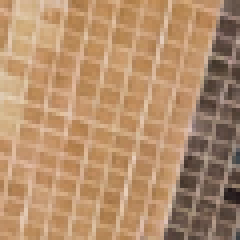}}
    \\
    \vspace{-3.5mm}
    \addtocounter{subfigure}{-6}
    \subfloat{\includegraphics[width=0.30\linewidth]{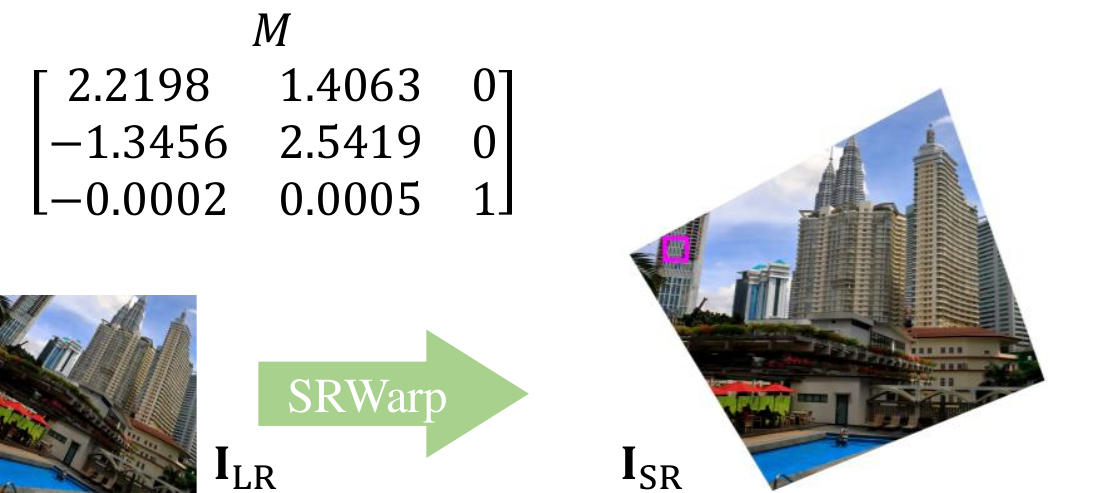}}
    \hfill
    \subfloat{\includegraphics[width=0.135\linewidth]{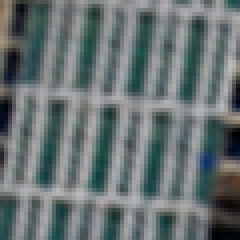}}
    \hfill
    \subfloat{\includegraphics[width=0.135\linewidth]{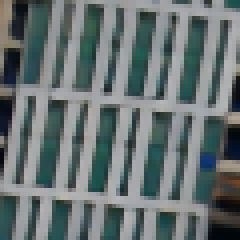}}
    \hfill
    \subfloat{\includegraphics[width=0.135\linewidth]{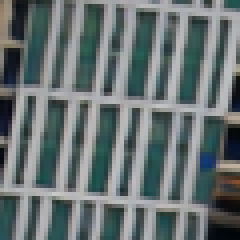}}
    \hfill
    \subfloat{\includegraphics[width=0.135\linewidth]{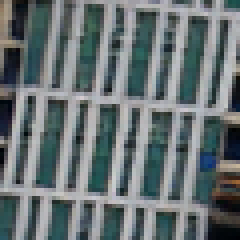}}
    \hfill
    \subfloat{\includegraphics[width=0.135\linewidth]{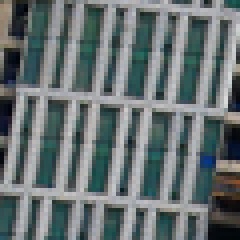}}
    \\
    \vspace{-3.5mm}
    \addtocounter{subfigure}{-6}
    \subfloat[$\img{LR}$, $\img{SR}$, and $M$]{\includegraphics[width=0.30\linewidth]{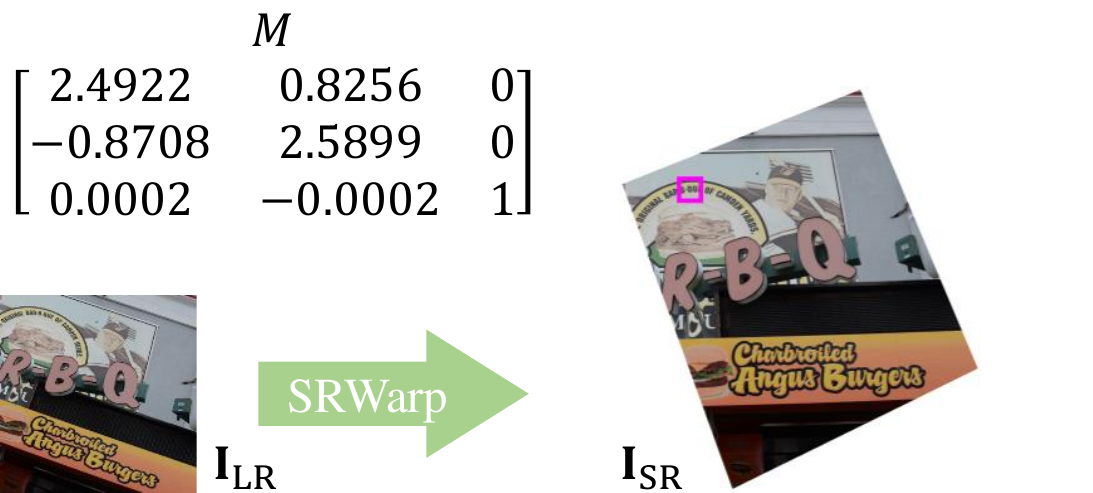}}
    \hfill
    \subfloat[\cv{cv2}]{\includegraphics[width=0.135\linewidth]{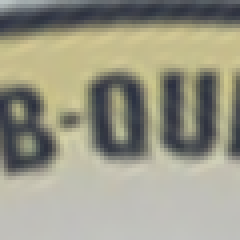}}
    \hfill
    \subfloat[RRDB]{\includegraphics[width=0.135\linewidth]{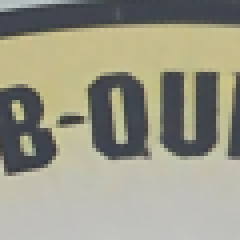}}
    \hfill
    \subfloat[\textbf{SRWarp}]{\includegraphics[width=0.135\linewidth]{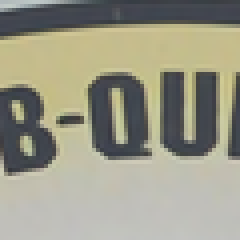}}
    \hfill
    \subfloat[\textbf{SRWarp$^+$}]{\includegraphics[width=0.135\linewidth]{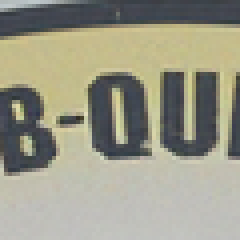}}
    \hfill
    \subfloat[GT]{\includegraphics[width=0.135\linewidth]{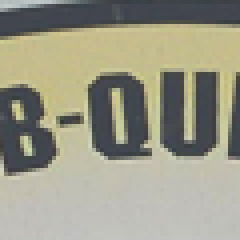}}
    \\
    \figcspace
    \caption{
        \textbf{More qualitative warping results on various images.}
        RRDB denotes the method which combines the state-of-the-art $\times 4$ RRDB~\cite{sr_esrgan} model with the bicubic warping algorithm from OpenCV~\cite{opencv_library}.
        \textbf{Top four rows:}
        To show the generalization ability of the proposed method, we choose random DIV2K images (`\emph{0831.png},' `\emph{0855.png},' `\emph{0864.png},' and `\emph{0887.png}') and transformation matrices to test the proposed SRWarp method.
        \textbf{Bottom four rows:}
        We construct the novel Flickr2KW$_\text{Test}$ dataset from the high-quality Flickr2K~\cite{sr_edsr} images, following the same pipeline of the DIV2KW$_\text{Test}$.
        Patches are cropped from the Flickr2KW$_\text{Test}$ `\emph{000004.png},' `\emph{000957.png},' `\emph{001219.png}', and `\emph{001290.png},' respectively. 
    }
    \label{fig:qualitative_more}
    \figspace
\end{figure*}

\section{Additional Results}
\figref{fig:qualitative_more} shows additional qualitative results from the proposed warping method.
Since we optimize the widely-used $L_1$ objective function to train our SRWarp, it is straightforward to introduce perceptual~\cite{others_perceptual} and adversarial~\cite{gans, gans_dc} loss terms to acquire photo-realistic~\cite{sr_srgan, sr_esrgan} warped results (SRWarp$^+$).
We first construct a discriminator network $\mathcal{D}$ following Ledig~\etal~\cite{sr_srgan}, while the fully-connected layers at the end of the original model are replaced to $1 \times 1$ convolutions~\cite{others_pix2pix}.
The spectral normalization~\cite{gans_sn} is also applied for stable training.
We update the discriminator network by the corresponding loss term $\mathcal{L}_\text{dis}$, which is defined as follows:
\begin{equation}
    \mathcal{L}_\text{dis} = -\log{ \mathcal{D} \paren{ \img{HR-crop} } } - \log{ \paren{1 - \mathcal{D} \paren{ \img{SR-crop} } } },
    \label{eq:dis}
\end{equation}
where we crop $96 \times 96$ patches $\img{HR-crop}$ and $\img{SR-crop}$ from the original images $\img{HR}$ and $\img{SR}$ to avoid void regions to be fed to the following discriminator model.
We omit averaging over the batch dimension for simplicity.
Then, we alternately optimize the perceptual objective function $\mathcal{L}_\text{per}$ for the SRWarp model which is defined as follows:
\begin{equation}
    \begin{split}
        \mathcal{L}_\text{per} 
        &= \alpha \frac{1}{ \normzero{\mathbf{m}} } \normone{\mathbf{m} \odot \paren{ \img{SR} - \img{HR} }} \\
        &+ \normone{ \vgg{ \img{SR-crop} } - \vgg{ \img{HR-crop} } } \\
        &- \beta \log{ \mathcal{D} \paren{\img{SR-crop}} },
    \end{split}
    \label{eq:per}
\end{equation}
where $\mathcal{V}_\text{54}$ is a pre-trained VGG-19~\cite{net_vgg, others_perceptual} network up to \cv{conv5\_4} layer, $\alpha = 0.002$ and $\beta =0.01$ are hyperparameters, respectively.
We note that the cropped images $\img{SR-crop}$ and $\img{HR-crop}$ are aligned so that the VGG-loss term can be used to improve the perceptual quality of the SR results.

\newpage
{\small
\bibliographystyle{ieee_fullname}
\bibliography{egbib}

\begin{thebibliography}{10}\itemsep=-1pt

\bibitem{data_div2k}
Eirikur Agustsson and Radu Timofte.
\newblock {NTIRE} 2017 challenge on single image super-resolution: Dataset and
  study.
\newblock In {\em CVPR Workshops}, 2017.

\bibitem{sr_explorable}
Yuval Bahat and Tomer Michaeli.
\newblock Explorable super resolution.
\newblock In {\em CVPR}, 2020.

\bibitem{opencv_library}
G. Bradski.
\newblock {The OpenCV Library}.
\newblock {\em Dr. Dobb's Journal of Software Tools}, 2000.

\bibitem{sr_realworld}
Jianrui Cai, Hui Zeng, Hongwei Yong, Zisheng Cao, and Lei Zhang.
\newblock Toward real-world single image super-resolution: A new benchmark and
  a new model.
\newblock In {\em ICCV}, 2019.

\bibitem{sr_camera}
Chang Chen, Zhiwei Xiong, Xinmei Tian, Zheng-Jun Zha, and Feng Wu.
\newblock Camera lens super-resolution.
\newblock In {\em CVPR}, 2019.

\bibitem{dai2017deformable}
Jifeng Dai, Haozhi Qi, Yuwen Xiong, Yi Li, Guodong Zhang, Han Hu, and Yichen
  Wei.
\newblock Deformable convolutional networks.
\newblock In {\em ICCV}, 2017.

\bibitem{sr_san}
Tao Dai, Jianrui Cai, Yongbing Zhang, Shu-Tao Xia, and Lei Zhang.
\newblock Second-order attention network for single image super-resolution.
\newblock In {\em CVPR}, 2019.

\bibitem{sr_srcnn}
Chao Dong, Chen~Change Loy, Kaiming He, and Xiaoou Tang.
\newblock Image super-resolution using deep convolutional networks.
\newblock {\em TPAMI}, 2016.

\bibitem{Fu_2019_CVPR}
Ying Fu, Tao Zhang, Yinqiang Zheng, Debing Zhang, and Hua Huang.
\newblock Hyperspectral image super-resolution with optimized {RGB} guidance.
\newblock In {\em CVPR}, 2019.

\bibitem{gao2019deformable}
Hang Gao, Xizhou Zhu, Steve Lin, and Jifeng Dai.
\newblock Deformable kernels: Adapting effective receptive fields for object
  deformation.
\newblock In {\em ICLR}, 2020.

\bibitem{gans}
Ian Goodfellow, Jean Pouget-Abadie, Mehdi Mirza, Bing Xu, David Warde-Farley,
  Sherjil Ozair, Aaron Courville, and Yoshua Bengio.
\newblock Generative adversarial nets.
\newblock In {\em NIPS}, 2014.

\bibitem{ewa}
N. {Greene} and P.~S. {Heckbert}.
\newblock Creating raster omnimax images from multiple perspective views using
  the elliptical weighted average filter.
\newblock {\em CG \& A}, 6(6):21--27, 1986.

\bibitem{sr_dbpn}
Muhammad Haris, Gregory Shakhnarovich, and Norimichi Ukita.
\newblock Deep back-projection networks for super-resolution.
\newblock In {\em CVPR}, 2018.

\bibitem{vsr_starnet}
Muhammad Haris, Greg Shakhnarovich, and Norimichi Ukita.
\newblock Space-time-aware multi-resolution video enhancement.
\newblock In {\em CVPR}, 2020.

\bibitem{net_residual}
Kaiming He, Xiangyu Zhang, Shaoqing Ren, and Jian Sun.
\newblock Deep residual learning for image recognition.
\newblock In {\em CVPR}, 2016.

\bibitem{sr_meta}
Xuecai Hu, Haoyuan Mu, Xiangyu Zhang, Zilei Wang, Tieniu Tan, and Jian Sun.
\newblock Meta-{SR}: A magnification-arbitrary network for super-resolution.
\newblock In {\em CVPR}, 2019.

\bibitem{others_pix2pix}
Phillip Isola, Jun-Yan Zhu, Tinghui Zhou, and Alexei~A. Efros.
\newblock Image-{T}o-{I}mage translation with conditional adversarial networks.
\newblock In {\em CVPR}, 2017.

\bibitem{jaderberg2015spatial}
Max Jaderberg, Karen Simonyan, Andrew Zisserman, and Koray Kavukcuoglu.
\newblock Spatial transformer networks.
\newblock In {\em NIPS}, 2015.

\bibitem{jeon2017active}
Yunho Jeon and Junmo Kim.
\newblock Active convolution: Learning the shape of convolution for image
  classification.
\newblock In {\em CVPR}, 2017.

\bibitem{jia2016dynamic}
Xu Jia, Bert De~Brabandere, Tinne Tuytelaars, and Luc~V Gool.
\newblock Dynamic filter networks.
\newblock In {\em NIPS}, 2016.

\bibitem{vsr_duf}
Younghyun Jo, Seoung Wug~Oh, Jaeyeon Kang, and Seon Joo~Kim.
\newblock Deep video super-resolution network using dynamic upsampling filters
  without explicit motion compensation.
\newblock In {\em CVPR}, 2018.

\bibitem{others_perceptual}
Justin Johnson, Alexandre Alahi, and Li Fei-Fei.
\newblock Perceptual losses for real-time style transfer and super-resolution.
\newblock In {\em ECCV}, 2016.

\bibitem{sr_vdsr}
Jiwon Kim, Jung~Kwon Lee, and Kyoung~Mu Lee.
\newblock Accurate image super-resolution using very deep convolutional
  networks.
\newblock In {\em CVPR}, 2016.

\bibitem{others_adam}
Diederik~P Kingma and Jimmy Ba.
\newblock Adam: A method for stochastic optimization.
\newblock In {\em ICLR}, 2015.

\bibitem{sr_lapsrn}
Wei-Sheng Lai, Jia-Bin Huang, Narendra Ahuja, and Ming-Hsuan Yang.
\newblock Deep laplacian pyramid networks for fast and accurate
  super-resolution.
\newblock In {\em CVPR}, 2017.

\bibitem{lai2018fast}
Wei-Sheng Lai, Jia-Bin Huang, Narendra Ahuja, and Ming-Hsuan Yang.
\newblock Fast and accurate image super-resolution with deep laplacian pyramid
  networks.
\newblock {\em TPAMI}, 41(11):2599--2613, 2018.

\bibitem{le2020deep}
Hoang Le, Feng Liu, Shu Zhang, and Aseem Agarwala.
\newblock Deep homography estimation for dynamic scenes.
\newblock In {\em CVPR}, 2020.

\bibitem{sr_srgan}
Christian Ledig, Lucas Theis, Ferenc Huszar, Jose Caballero, Andrew Cunningham,
  Alejandro Acosta, Andrew Aitken, Alykhan Tejani, Johannes Totz, Zehan Wang,
  and Wenzhe Shi.
\newblock Photo-realistic single image super-resolution using a generative
  adversarial network.
\newblock In {\em CVPR}, 2017.

\bibitem{vsr_mucan}
Wenbo Li, Xin Tao, Taian Guo, Lu Qi, Jiangbo Lu, and Jiaya Jia.
\newblock {M}u{CAN}: Multi-correspondence aggregation network for video
  super-resolution.
\newblock In {\em ECCV}, 2020.

\bibitem{sr_edsr}
Bee Lim, Sanghyun Son, Heewon Kim, Seungjun Nah, and Kyoung~Mu Lee.
\newblock Enhanced deep residual networks for single image super-resolution.
\newblock In {\em CVPR Workshops}, 2017.

\bibitem{liu2018partialinpainting}
Guilin Liu, Fitsum~A. Reda, Kevin~J. Shih, Ting-Chun Wang, Andrew Tao, and
  Bryan Catanzaro.
\newblock Image inpainting for irregular holes using partial convolutions.
\newblock In {\em ECCV}, 2018.

\bibitem{liu2018partialpadding}
Guilin Liu, Kevin~J. Shih, Ting-Chun Wang, Fitsum~A. Reda, Karan Sapra, Zhiding
  Yu, Andrew Tao, and Bryan Catanzaro.
\newblock Partial convolution based padding.
\newblock In {\em arXiv}, 2018.

\bibitem{sr_srflow}
Andreas Lugmayr, Martin Danelljan, Luc Van~Gool, and Radu Timofte.
\newblock {SRF}low: Learning the super-resolution space with normalizing flow.
\newblock In {\em ECCV}, 2020.

\bibitem{data_bsd200}
David Martin, Charless Fowlkes, Doron Tal, and Jitendra Malik.
\newblock A database of human segmented natural images and its application to
  evaluating segmentation algorithms and measuring ecological statistics.
\newblock In {\em ICCV}, 2001.

\bibitem{gans_sn}
Takeru Miyato, Toshiki Kataoka, Masanori Koyama, and Yuichi Yoshida.
\newblock Spectral normalization for generative adversarial networks.
\newblock In {\em ICLR}, 2018.

\bibitem{sr_han}
Ben Niu, Weilei Wen, Wenqi Ren, Xiangde Zhang, Lianping Yang, Shuzhen Wang,
  Kaihao Zhang, Xiaochun Cao, and Haifeng Shen.
\newblock Single image super-resolution via a holistic attention network.
\newblock In {\em ECCV}, 2020.

\bibitem{gans_dc}
Alec Radford, Luke Metz, and Soumith Chintala.
\newblock Unsupervised representation learning with deep convolutional
  generative adversarial networks.
\newblock {\em arXiv}, 2015.

\bibitem{sr_espcn}
Wenzhe Shi, Jose Caballero, Ferenc Huszar, Johannes Totz, Andrew~P. Aitken, Rob
  Bishop, Daniel Rueckert, and Zehan Wang.
\newblock Real-time single image and video super-resolution using an efficient
  sub-pixel convolutional neural network.
\newblock In {\em CVPR}, 2016.

\bibitem{net_vgg}
Karen Simonyan and Andrew Zisserman.
\newblock Very deep convolutional networks for large-scale image recognition.
\newblock In {\em ICLR}, 2015.

\bibitem{swaminathan2000nonmetric}
Rahul Swaminathan and Shree~K Nayar.
\newblock Nonmetric calibration of wide-angle lenses and polycameras.
\newblock {\em TPAMI}, 22(10):1172--1178, 2000.

\bibitem{sr_crossnetpp}
Yang Tan, Haitian Zheng, Yinheng Zhu, Xiaoyun Yuan, Xing Lin, David Brady, and
  Lu Fang.
\newblock Cross{N}et++: Cross-scale large-parallax warping for reference-based
  super-resolution.
\newblock {\em TPAMI}, 2020.

\bibitem{vsr_tdan}
Yapeng Tian, Yulun Zhang, Yun Fu, and Chenliang Xu.
\newblock {TDAN}: Temporally-deformable alignment network for video
  super-resolution.
\newblock In {\em CVPR}, 2020.

\bibitem{sr_arb}
Longguang Wang, Yingqian Wang, Zaiping~Lin Lin, Jungang Yang, Wei An, and Yulan
  Guo.
\newblock Learning for scale-arbitrary super-resolution from scale-specific
  networks.
\newblock {\em arXiv}, 2020.

\bibitem{vsr_edvr}
Xintao Wang, Kelvin~CK Chan, Ke Yu, Chao Dong, and Chen Change~Loy.
\newblock {EDVR}: Video restoration with enhanced deformable convolutional
  networks.
\newblock In {\em CVPRW}, 2019.

\bibitem{sr_esrgan}
Xintao Wang, Ke Yu, Shixiang Wu, Jinjin Gu, Yihao Liu, Chao Dong, Yu Qiao, and
  Chen Change~Loy.
\newblock {ESRGAN:} enhanced super-resolution generative adversarial networks.
\newblock In {\em ECCV Workshops}, 2018.

\bibitem{wang2020spatial}
Yingqian Wang, Longguang Wang, Jungang Yang, Wei An, Jingyi Yu, and Yulan Guo.
\newblock Spatial-angular interaction for light field image super-resolution.
\newblock In {\em ECCV}, 2020.

\bibitem{sr_cdc}
Pengxu Wei, Ziwei Xie, Hannan Lu, ZongYuan Zhan, Qixiang Ye, Wangmeng Zuo, and
  Liang Lin.
\newblock Component divide-and-conquer for real-world image super-resolution.
\newblock In {\em ECCV}, 2020.

\bibitem{wronski2019handheld}
Bartlomiej Wronski, Ignacio Garcia-Dorado, Manfred Ernst, Damien Kelly, Michael
  Krainin, Chia-Kai Liang, Marc Levoy, and Peyman Milanfar.
\newblock Handheld multi-frame super-resolution.
\newblock {\em ACM TOG}, 38(4):1--18, 2019.

\bibitem{vsr_zooming_slowmo}
Xiaoyu Xiang, Yapeng Tian, Yulun Zhang, Yun Fu, Jan~P. Allebach, and Chenliang
  Xu.
\newblock Zooming {S}low-{M}o: Fast and accurate one-stage space-time video
  super-resolution.
\newblock In {\em CVPR}, 2020.

\bibitem{xue2019learning}
Zhucun Xue, Nan Xue, Gui-Song Xia, and Weiming Shen.
\newblock Learning to calibrate straight lines for fisheye image rectification.
\newblock In {\em CVPR}, 2019.

\bibitem{yao2020cross}
Jing Yao, Danfeng Hong, Jocelyn Chanussot, Deyu Meng, Xiaoxiang Zhu, and
  Zongben Xu.
\newblock Cross-attention in coupled unmixing nets for unsupervised
  hyperspectral super-resolution.
\newblock In {\em ECCV}, 2020.

\bibitem{data_set14}
Roman Zeyde, Michael Elad, and Matan Protter.
\newblock On single image scale-up using sparse-representations.
\newblock In {\em Curves and Surfaces}, 2010.

\bibitem{zhang2020content}
Jirong Zhang, Chuan Wang, Shuaicheng Liu, Lanpeng Jia, Nianjin Ye, Jue Wang, Ji
  Zhou, and Jian Sun.
\newblock Content-aware unsupervised deep homography estimation.
\newblock In {\em ECCV}, 2020.

\bibitem{Zhang_2020_CVPR}
Lei Zhang, Jiangtao Nie, Wei Wei, Yanning Zhang, Shengcai Liao, and Ling Shao.
\newblock Unsupervised adaptation learning for hyperspectral imagery
  super-resolution.
\newblock In {\em CVPR}, 2020.

\bibitem{Zhang_2019_CVPR}
Shuo Zhang, Youfang Lin, and Hao Sheng.
\newblock Residual networks for light field image super-resolution.
\newblock In {\em CVPR}, 2019.

\bibitem{sr_zllz}
Xuaner Zhang, Qifeng Chen, Ren Ng, and Vladlen Koltun.
\newblock Zoom to learn, learn to zoom.
\newblock In {\em CVPR}, 2019.

\bibitem{sr_rcan}
Yulun Zhang, Kunpeng Li, Kai Li, Lichen Wang, Bineng Zhong, and Yun Fu.
\newblock Image super-resolution using very deep residual channel attention
  networks.
\newblock In {\em ECCV}, 2018.

\bibitem{sr_rdn}
Yulun Zhang, Yapeng Tian, Yu Kong, Bineng Zhong, and Yun Fu.
\newblock Residual dense network for image super-resolution.
\newblock In {\em CVPR}, 2018.

\bibitem{ir_rdn}
Yulun Zhang, Yapeng Tian, Yu Kong, Bineng Zhong, and Yun Fu.
\newblock Residual dense network for image restoration.
\newblock {\em TPAMI}, 2020.

\bibitem{zhu2019deformable}
Xizhou Zhu, Han Hu, Stephen Lin, and Jifeng Dai.
\newblock Deformable {C}onv{N}ets v2: More deformable, better results.
\newblock In {\em CVPR}, 2019.

\end{thebibliography}
}

\end{document}